\documentclass[aip,pof,amsmath,amssymb,reprint]{revtex4-1} 

\usepackage{xcolor}

\newcommand{\chgenv}{\color{blue}}
\newcommand{\chg}[1]{\textcolor{blue}{#1}}
\definecolor{BLUE}{rgb}{0,0,1}  

\renewcommand{\chgenv}{}
\renewcommand{\chg}[1]{#1}

\makeatletter
\@ifclasswith{revtex4-1}{linenumbers}{
  \usepackage{etoolbox}
  \newcommand*{\linenomathpatch}[1]{
     \cspreto{#1}{\linenomath}
     \cspreto{#1*}{\linenomath}
     \csappto{end#1}{\endlinenomath}
     \csappto{end#1*}{\endlinenomath}}
  \AtBeginDocument{
    \linenomathpatch{equation}
    \linenomathpatch{gather}
    \linenomathpatch{align}}}{}
\makeatother

\usepackage{mathtools}

\usepackage{multirow}
\usepackage{graphicx}
\usepackage[caption=false]{subfig}
\graphicspath{{Figures/}}

\usepackage[plainruled]{algorithm2e}

\makeatletter
\def\@algocf@capt@plainruled{above}
\renewcommand{\algocf@caption@plainruled}{
  \vskip\AlCapSkip
  \box\algocf@capbox
  \vskip 4\algoheightrule}
\makeatother

\AtBeginDocument{\SetInd{0.1em}{1em}}

\SetAlgoCaptionSeparator{\chg{.}}

\usepackage{hyperref}
\hypersetup{
  colorlinks=false,
  final
}

\usepackage{cleveref}

\crefname{figure}{\chg{Fig.}}{\chg{Figs.}}
\Crefname{figure}{\chg{Figure}}{\chg{Figures}}
\crefname{table}{\chg{Table}}{\chg{Tables}}
\Crefname{table}{\chg{Table}}{\chg{Tables}}
\crefname{algorithm}{\chg{Alg.}}{\chg{Algs.}}
\Crefname{algorithm}{Algorithm}{Algorithms}

\makeatletter
\newcommand*{\oset}[2]{%
  {\mathop{#2}\limits^{\vbox to -1.0\ex@{\kern-\tw@\ex@
   \hbox{\scriptsize #1}\vss}}}}
\makeatother

\DeclareMathOperator*{\argmax}{arg\,max}

\newcommand*{\diff}{\ensuremath{\mathop{}\!\mathrm{d}}}
\newcommand*{\jac}{\ensuremath{\mathcal{J}}}
\newcommand*{\nablaxi}{\ensuremath{\nabla_{\svec{\xi}}}}
\newcommand*{\smat}[1]{\ensuremath{\mathsf{#1}}}
\newcommand*{\stmat}[1]{\ensuremath{\underline{{#1}}}}
\newcommand*{\bmat}[1]{\ensuremath{\mathcal{#1}}}
\newcommand*{\svec}[1]{\ensuremath{\vec{#1}}}
\newcommand*{\csvec}[1]{\ensuremath{\svec{\tilde{#1}}}}
\newcommand*{\stvec}[1]{\ensuremath{\mathbf #1}}
\newcommand*{\stvecg}[1]{\ensuremath{\boldsymbol #1}}

\newcommand*{\bvec}[1]{\ensuremath{\oset{\text{\tiny$\leftrightarrow$}}{\mbox{\stvec{#1}}}}}
\newcommand*{\bvecg}[1]{\ensuremath{\oset{\text{\tiny$\leftrightarrow$}}{\stvecg{#1}}}}
\newcommand*{\cbvec}[1]{\ensuremath{\oset{\text{\tiny$\leftrightarrow$}}{\tilde{\stvec{#1}}}}}
\newcommand*{\cbvecg}[1]{\ensuremath{\oset{\text{\tiny$\leftrightarrow$}}{\tilde{\stvecg{#1}}}}}

\begin{document}

\title{\chg{An unsupervised machine-learning-based shock sensor for high-order supersonic flow solvers}}

\author{Andr\'es Mateo-Gab\'in}
\email{andres.mgabin@upm.es.}
\author{Kenza Tlales}
\affiliation{ETSIAE-UPM-School of Aeronautics, Universidad Polit\'ecnica de Madrid, Madrid-Spain}

\author{Eusebio Valero}
\author{Esteban Ferrer}
\author{Gonzalo Rubio}
\affiliation{ETSIAE-UPM-School of Aeronautics, Universidad Polit\'ecnica de Madrid, Madrid-Spain}
\affiliation{Center for Computational Simulation, Universidad Polit\'ecnica de Madrid, Madrid-Spain}

\date{\today}

\begin{abstract}
We present a novel unsupervised \chg{machine-learning sock sensor} based on Gaussian Mixture Models (GMMs).
The proposed GMM sensor demonstrates remarkable accuracy in detecting shocks and is robust across diverse test cases \chg{with significantly less parameter tuning than other options}.
We compare the GMM-based sensor with state-of-the-art alternatives.
All methods are integrated into a high-order compressible discontinuous Galerkin solver, where \chg{two stabilization approaches are coupled to the sensor to provide examples of possible applications.
The Sedov blast and double Mach reflection cases demonstrate that our proposed sensor can enhance hybrid sub-cell flux-differencing formulations by providing accurate information of the nodes that require low-order blending.
Besides, supersonic test cases including high Reynolds numbers} showcase the \chg{sensor} performance \chg{when used to introduce entropy-stable artificial viscosity to capture shocks}, demonstrating the same effectiveness as fine-tuned state-of-the-art sensors.
The adaptive nature and ability to function without extensive training datasets make this GMM-based sensor suitable for complex geometries and varied flow configurations.
Our study reveals the potential of unsupervised \chg{machine-learning} methods, exemplified by \chg{this} GMM sensor, to improve the robustness and efficiency of advanced CFD codes.
\end{abstract}

% ...


\maketitle

\section{Introduction}
\label{sec:intro}


Shock waves are complex and significant fluid phenomena in engineering, observed, for example, in high-speed transport or in combustion and detonation \chg{processes.~\cite{anderson1990modern}} High-speed flows exhibit a combination of smooth regions and thin regions with abrupt changes in flow properties.
\chg{To} effectively handle the various scales present in these flows, it is necessary to employ robust and computationally efficient numerical schemes that maintain a high level of \chg{precision.~\cite{pirozzoli2011numerical}}
Standard discretizations for smooth flows may exhibit oscillations when shocks are present, and require special techniques for shock regularization within designated \chg{regions.~\cite{pirozzoli2011numerical}}

\chg{The} shock-fitting approach, which explicitly tracks and fits shock waves, is one method used to handle \chg{shocks.~\cite{paciorri2009shock,Zahr2019,shi2022implicit}} However, the utilization of shock fitting is limited, primarily due to the difficulties it presents when applied to unstructured grids. 
An alternative and more commonly employed approach is the use of shock-capturing methods. The choice of the baseline discretization scheme determines the availability of various shock-capturing approaches. For finite volume \chg{discretizations,~\cite{eymard2000finite}} typical options include TVD limiting strategies~\cite{yee1989class} or essentially non-oscillatory (W)ENO \chg{reconstructions.~\cite{harten1986some,shu1988efficient,liu1994weighted,jiang1996efficient,leveque2002finite,shu2003high,stevens2020enhancement}} In the case of flux reconstruction (FR)~\cite{Huynh2007,vincent2011new} and discontinuous Galerkin (DG) \cite{cockburn2012discontinuous} schemes, the methods generally fall into two categories.
The first category involves the local switching of the discretization operator to a more robust one, achieved through h-refinement and/or p-coarsening. By employing appropriate limiting techniques, this operator ensures both accuracy and solution \chg{boundedness.~\cite{cockburn1990runge,burbeau2001problem,qiu2005hermite,qiu2005comparison,balsara2007sub,yang2009parameter,zhong2013simple,dumbser2014posteriori,ntoukas2020freeenergy,ntoukas2021entropy,mossier2022padaptive}} A similar approach involves performing a hybrid blending with a low-order \chg{sub-cell} variant of the \chg{scheme.~\cite{HENNEMANN2021109935,RUEDARAMIREZ2022105627,MATEOGABIN2023112298,lin2023high}}
In the second category, known as the artificial viscosity shock-capturing method, a local diffusion operator with a predetermined strength is introduced to regularize the solution once a shock is \chg{detected.~\cite{vonneumann1950method,tadmor1989convergence,guo2001spectral,hartmann2002adaptive,persson2006sub,feistauer2010discontinuous,barter2010shock,lv2016entropy,fraysse2016upwind,redondo2017artificial,mateogabin2022entropy}}

Regardless of the specific method used, accurately determining the precise location of shock waves is of paramount importance. This becomes particularly critical since shock regularization methods tend to introduce excessive dissipation in smooth regions, potentially leading to loss of accuracy. This task itself is generally accomplished using parameter-dependent indicator functions derived from physical \chg{considerations,~\cite{jameson1981numerical}} modal smoothness \chg{estimates,~\cite{persson2006sub,klockner2011viscous,huerta2012simple}} or image detection \chg{concepts.~\cite{rusanov1973processing,vorozhtsov1987shock,liou1995image,wu2013review,sheshadri2014shock}
In the context of general Galerkin methods, stabilized approaches such as the Streamline-Upwind/Petrov-–Galerkin (SUPG) formulation~\cite{Tezduyar2004,Tezduyar2006a,Tezduyar2006b,Tezduyar2007a,Tezduyar2007b} provide also examples of sensors specifically developed for multi-element discretizations.
Nevertheless, an} inherent limitation of all sensors is the requirement for manual adjustment of numerical parameters. An improper configuration of these parameters can potentially lead to simulation failures or crashes.

\chg{In} recent years, machine learning (ML) techniques have become increasingly integrated into natural sciences and \chg{engineering.~\cite{brunton_kutz_2019,GARNIER2021104973,vinuesa2022enhancing}} This trend has also extended to the field of computational fluid dynamics (CFD), where numerous successful applications of ML have emerged. The current state and future prospects of ML in fluid mechanics have been extensively discussed in \chg{various references,~\cite{brenner2019perspective,brunton2020machine,fukami2020assessment,vinuesa2022enhancing} including applications in engineering.~\cite{clainche2022improving}}
Shock capturing is no exception, and one popular approach with supervised ML is to use a multilayer perceptron to detect troubled \chg{cells~\cite{veiga2018general,ray2018artificial,ray2019detecting,morgan2020machine} where additional stabilization is applied by using TVD~\cite{BECK2020109824} and TVB~\cite{Xinyue2021} limiters, WENO schemes,~\cite{KOSSACZKA2021100201,Xue2022} or artificial viscosity.~\cite{ZEIFANG2021110475}}
An alternative to supervised ML is unsupervised ML, which involves algorithms that analyze data to identify patterns without relying on explicit labels or prior knowledge. These methods excel at uncovering hidden structures and relationships in complex datasets, making them valuable for tasks such as exploratory data analysis and anomaly detection. Examples of unsupervised ML techniques include k-means, \chg{GMM}, and DBSCAN. These methods have demonstrated effectiveness in various applications, enabling researchers to gain insights and discover valuable information in the data without the need for labeled training examples.
\chg{Particularly in CFD}, unsupervised ML models have been applied to tasks such as identifying flow regions within the flow field~\cite{otmaniRobustDetectionViscous2022,saettaIdentificationFlowFielda} and performing mesh \chg{adaptation.~\cite{tlales2022machine}}
However, the potential of unsupervised ML techniques to develop numerical schemes with shock-capturing \chg{capabilities~\cite{Zhu2021}} remains largely unexplored. 

This study focuses on the development of a novel shock-capturing sensor using unsupervised machine learning techniques. The sensor devised in this research demonstrates high accuracy by effectively identifying shocks, while other features such as turbulence or unresolved regions remain undetected. The shock \chg{sensor} is coupled with \chg{two different shock-regularization strategies} and the DGSEM \chg{method,~\cite{Black1999,kopriva2009implementing,Ferrer2023}} resulting in \chg{precise and resilient discretizations}. A notable feature of this sensor is its \chg{almost} parameter-free nature, \chg{having only one parameter with minimal influence on the final result}. To assess its performance, a comprehensive comparison is conducted with state-of-the-art alternatives in the field.

This work is organized as follows: \cref{sec:ns} presents the \chg{Navier--Stokes} equations, including the additional stabilization terms \chg{that implement the artificial viscosity method}. \Cref{sec:gmm} presents the methodology, explaining the new shock detection technique, and \cref{sec:sensors} describes the traditional sensors used as a reference. \Cref{sec:horses} summarizes the numerical approximation in space and time\chg{, including a node-wise hybrid formulation combining high- and low-order spatial discretizations}. \Cref{sec:results} displays the results, comparing the new methodology with state-of-the-art alternatives. Finally, \cref{sec:conclusions} concludes the study, outlining the findings and their implications.

\section{\chg{Navier--Stokes} equations}
\label{sec:ns}

The \chg{Navier--Stokes} equations are a set of advection-diffusion equations on a spatial domain~$\Omega$ for the conservative variables,~$\stvec{q} = (\rho, \rho\svec{v}, \rho e)^T$.  These equations can be expressed using the notation \chg{as follows\cite{Gassner2018}}:
\begin{equation}
\label{eq:ns:advdiff}
\begin{gathered}
    \stvec{q}_t + \nabla \cdot \bvec{f}_e = \nabla \cdot \bvec{f}_v, \\
    \bvec{f} = (\stvec{f}, \stvec{g}, \stvec{h})^T, \quad
    \nabla \cdot \bvec{f} = \frac{\partial \stvec{f}}{\partial x} + \frac{\partial \stvec{g}}{\partial y} + \frac{\partial \stvec{h}}{\partial z},
\end{gathered}
\end{equation}
where~$\rho$ denotes the density,~$\svec{v}$ represents the velocity vector, and~$e$ corresponds to the specific total energy.
Additional information on the advective and viscous fluxes,~$\bvec{f}_e$ and~$\bvec{f}_v$, can be found in~\cref{sub:ans}.

In this study, we present a shock-capturing sensor and \chg{assess its performance by combining it with two common stabilization strategies.
One of such approaches is the addition of artificial viscosity directly into the equations, with the primary objective of addressing the limited numerical dissipation of high-order schemes.
To achieve this, we incorporate an additional viscous term into \cref{eq:ns:advdiff}, such that ${\bvec{f}_v \rightarrow \bvec{f}_v + \bvec{f}_a}$. This augmentation allows us to compensate for the insufficient dissipation in the original formulation.
In the following we use the framework introduced by Guermond and Popov~\cite{Guermond2014} to define~$\bvec{f}_a$, as it takes into consideration entropy principles that lead to an entropy-stable semi-discretization~\cite{mateogabin2022entropy}:}
\begin{equation}
\label{eq:ns:gpflux}
\begin{gathered}
    \bvec{f}_a = \alpha_a \left(\begin{array}{c}
        \nabla\rho \\
        \nabla\rho\otimes\svec{v} \\
        \nabla\left(\rho e_i\right) + \frac{1}{2}\chg{\lVert\svec{v}\rVert}^2\nabla\rho
    \end{array}\right) + \mu_a \left(\begin{array}{c}
        0 \\
        \rho \nabla^s\svec{v} \\
        \rho \svec{v}\cdot\nabla^s\svec{v}
    \end{array}\right), \\
    \nabla^s\svec{v} = \frac{1}{2} \left(\nabla\svec{v} + \left(\nabla\svec{v}\right)^T\right).
\end{gathered}
\end{equation}
The values of $\alpha_a$ and $\mu_a$ in \cref{eq:ns:gpflux} govern the level of artificial dissipation, and we calculate them based on the value of a sensor $s \in [0,1]$. 

The specific relationship between the sensor, $s$, and the parameters $\alpha_a$ and $\mu_a$ is based on the spatial discretization employed. Typically, it requires scaling $\alpha_a$ and $\mu_a$ with an estimate of the sub-cell resolution to remove the dependence of viscosity on the element size. In this study, we adopt a standard scaling that assumes a spatial tessellation of the domain into elements, as expressed \chg{by~\cite{Persson2006}}
\begin{equation}\chgenv
\label{eq:ns:gpvisc}
    \alpha_a = \mu_a = \mu_0 h s, \quad h = \frac{V^{1/d}}{P + 1}.
\end{equation}
Here, $V$ denotes the volume of the element, $d$ is the \chg{number of spatial dimensions}, and $P$ is the polynomial order used to represent the solution within the element.
\chg{This scaling assumes an equispaced distribution of~$P+1$ nodes inside an element with no directionality, i.e. a segment, square or cube, depending on the number of spatial dimensions of the domain.
This is a common strategy in high-order CFD codes, typically used in the calculation of the time-step size, but it is not accurate when the elements are not completely isotropic.~\cite{Joshi2023}
Since the resolution of the mesh depends on the distance between solution nodes, a more accurate evaluation of~$h$ would consider the inter-nodal distance at the position of the shock and in the direction across it.
Conversely, using the maximum inter-nodal distance leads to a more conservative estimate that would overestimate the required viscosity in most cases.
The method of \cref{eq:ns:gpvisc} lays in between, being a simple approximation that yields good results in all the test cases of \cref{sec:results}.}

\chg{More} details on the numerical approach used to approximate the solution of \cref{eq:ns:advdiff} are elaborated in \cref{sec:horses,sec:aDisc}
\chg{In} the following section, we propose a novel method to compute the sensor value $s$\chg{, and} other standard approaches are detailed in \cref{sec:sensors}.

\section{Unsupervised machine learning based shock sensor}
\label{sec:gmm}

Clustering algorithms aim to identify patterns in the given feature space by grouping points based on shared properties. In the context of this study, our primary focus is \chg{detecting regions of the flow field} that contain shocks. Shocks are identified by high gradients in the flow variables. However, these discontinuities typically occupy a relatively small portion of the fluid domain, resulting in only a small percentage of nodes being part of the identified clusters.
Consequently, the majority of points in the feature space will tend to be concentrated around low values, while the clusters associated with discontinuities should be small and situated further into the higher-gradient region.

Among the various unsupervised machine learning algorithms available, we opt for the \chg{GMM} because of its advantages in our context. One of the key benefits is that it only requires one parameter, the number of clusters, and our implementation can handle it automatically, reducing the need for manual tuning.
Furthermore, the GMM exhibits excellent efficiency in terms of implementation. The operations involved in iteratively updating the clusters can be expressed as reductions over threads and processes, leading to minimized communication between them. This efficient parallel performance ensures that the cost of detecting shocks does not significantly impact the scalability of the software.
Given our ultimate goal of integrating this sensor into a stabilization approach for a \chg{CFD} solver, the favorable parallel performance of the GMM is crucial in maintaining the scalability of the software and overall performance.

\subsection{\chg{Gaussian Mixture Model}}

The GMM assumes that the points in the feature space have been randomly drawn from a combination of Gaussians, each with a certain probability, denoted $\tau$, of being selected. The Probability Density Function (PDF) of the GMM is given by:
\begin{equation}\chgenv
\label{eq:gmm:gm}
\begin{gathered}
    f\left(\svec{x}\right) = \sum_{j=0}^K f_j\left(\svec{x}\right), \quad
    f_j\left(\svec{x}\right) = \tau_j \mathcal{N}_v\left(\svec{x}; \svec{\mu}_j, \smat{S}_j\right), \\
    \mathcal{N}_v\left(\svec{x}; \svec{\mu}, \smat{S}\right) = \frac{1}{\sqrt{(2\pi)^v\lvert\smat{S}\rvert}} \exp \left[\frac{1}{2}\left(\svec{\mu}-\svec{x}\right)^T \smat{S}^{-1} \left(\svec{x}-\svec{\mu}\right)\right], \\
    \sum_{j=0}^K \tau_j = 1,
\end{gathered}
\end{equation}
where~$v$ is the dimension of the feature space (${\svec{x}_i \in \mathbb{R}^v}$),~$K$ is the number of components \chg{(or clusters)}, and with expected values~\chg{$\{\svec{\mu}_j\}_{j=0}^K$} and covariance matrices~\chg{$\{\smat{S}_j\}_{j=0}^K$}.
\chg{These matrices have size~$v \times v$ and are symmetric positive definite, a property that can be leveraged to optimize the computation of the inverse required in \cref{eq:gmm:gm}.}
The most common approach for fitting the mixture \chg{to a set of~$N+1$ given points} is the \emph{Expectation-Maximization} (EM) algorithm. This iterative technique is used to find the optimal values of the set of parameters \chg{$(\tau_j, \svec{\mu}_j, \smat{S}_j)$} that maximizes the log-likelihood of the data,
\begin{equation}\chgenv
\label{eq:gmm:log_likelihood}
    \log{L} = \sum_{i=0}^N \log \sum_{j=0}^K \tau_j \mathcal{N}_v\left(\svec{x}_i; \svec{\mu}_j, \smat{S}_j\right).
\end{equation}
The GMM fitting process \chg{is described in \cref{alg:gmm:gmm}, and starts with an initial Gaussian mixture.
It can be initialized in multiple ways, and the ones used in this work are discussed later in this section.}
\begin{algorithm}
    \caption{EM method applied to the \chg{GMM} (see also \cref{alg:gmm:estep,alg:gmm:mstep}).}
    \label{alg:gmm:gmm}
    \DontPrintSemicolon
    \KwIn{\chg{$\svec{x}$}, \chg{$\tau$}, $\svec{\mu}$, \chg{$\smat{S}$}, \chg{$max\_iters$}, $\epsilon$, $\delta$}
    \KwOut{\chg{$\tau$}, $\svec{\mu}$, \chg{$\smat{S}$}}
    \BlankLine

    Initialize \chg{$\tau$}, $\svec{\mu}$, \chg{$\smat{S}$}\;
    \BlankLine

    $prevlogL \gets \infty$\;
    \For{$iter$ in 1, \dots, max\_iters}{
        $logL, prob \gets Estep(\chg{\svec{x}}, \chg{\tau}, \svec{\mu}, \chg{\smat{S}})$\;
        $\chg{\tau}, \svec{\mu}, \chg{\smat{S}} \gets Mstep(\chg{\svec{x}}, \epsilon, prob, \chg{\tau}, \svec{\mu}, \chg{\smat{S}})$\;
        Adaptation step: delete overlapping clusters\;
        \BlankLine

        \eIf{$(logL - prevlogL) / logL < \delta$}{
            Leave the loop. The algorithm has converged\;
        }{
            $prevlogL \gets logL$\;
        }
    }
\end{algorithm}
\chg{This initial guess is iteratively refined in two steps (E and M) that are executed until the descent rate of the log-likelihood is below~$\delta \to 0$.
During the E step (\cref{alg:gmm:estep}) the mixture is not modified, and only the log-likelihood and the probability matrix,~$R_{ij}$, are computed:}
\begin{equation*}\chgenv
    R_{ij} = \log p(k=j|\svec{x}_i) = \log\left(\frac{f_j(\svec{x}_i)}{\sum_{k=0}^K f_k(\svec{x}_i)}\right).
\end{equation*}
\begin{algorithm}
    \caption{E step of the \chg{GMM}.}
    \label{alg:gmm:estep}
    \DontPrintSemicolon
    \KwIn{\chg{$\svec{x}$}, \chg{$\tau$}, $\svec{\mu}$, \chg{$\smat{S}$}}
    \KwOut{$logL$, $prob$}
    \BlankLine

    \chg{\For{$j$ in 0, \dots, K}{
        \For{$i$ in 0, \dots, N}{
            $R_{ij} \gets \log \left[\tau_j \mathcal{N}_v(\svec{x}_i; \svec{\mu}_j, \smat{S}_j)\right]$\;
        }
    }}
    \BlankLine

    $logL \gets 0$\;
    \chg{\For{$i$ in 0, \dots, N}{
        $s \gets \log\left[\sum_j \exp(R_{ij})\right]$\;
        $logL \gets logL + s$\;
        \For{$j$ in 0, \dots, K}{
            $R_{ij} \gets \exp(R_{ij} - s)$\;
        }
    }}
\end{algorithm}
\chg{In the M step (\cref{alg:gmm:mstep}), these probabilities are utilized to improve the mixture, reducing the value of the log-likelihood for the given set of points:}
\begin{equation*}\chgenv
\begin{gathered}
    \tau_j = \frac{N_j}{N+1}, \quad N_j = \sum_{i=0}^N R_{ij}, \\
    \svec{\mu}_j = \sum_{i=0}^N \frac{\svec{x}_i R_{ij}}{N_j}, \\
    \smat{S}_j = \frac{\sum_{i=0}^N R_{ij} (\svec{x}_i - \svec{\mu}_j)\otimes(\svec{x}_i - \svec{\mu}_j)}{N_j} + \epsilon \smat{I}_v,
\end{gathered}
\end{equation*}
\chg{with $\svec{a}\otimes\svec{b}$ the tensor product of both vectors.
In \cref{alg:gmm:mstep} we add~$\epsilon\smat{I}_v$, where~$\smat{I}_v$ is the identity matrix of size~$v\times v$ and~$\epsilon \to 0$, to the covariance matrix to avoid division-by-zero errors in \cref{eq:gmm:gm}.}
\begin{algorithm}
    \caption{M step of the \chg{GMM}.}
    \label{alg:gmm:mstep}
    \DontPrintSemicolon
    \KwIn{\chg{$\svec{x}$}, $\epsilon$, \chg{$R$}, \chg{$\tau$}, $\svec{\mu}$, \chg{$\smat{S}$}}
    \KwOut{\chg{$\tau$}, $\svec{\mu}$, \chg{$\smat{S}$}}
    \BlankLine

    \chg{\For{$j$ in 0, \dots, K}{
        $N_j \gets \sum_i R_{ij}$\;
        $\tau_j \gets N_j / (N+1)$\;
        $\mu_j \gets \sum_i x_i \cdot R_{ij} / N_j$\;
        $\smat{S}_j \gets \sum_i R_{ij} \cdot (x_i - \mu_j) \otimes (x_i - \mu_j) / N_j$\;
        $\smat{S}_{j,ii} \gets \smat{S}_{j,ii} + \epsilon\smat{I}_v$\;
    }}
\end{algorithm}

In the main loop \chg{of \cref{alg:gmm:gmm}} we have introduced an additional step to ensure that no two clusters overlap.
In our implementation, clusters are considered to overlap when all components of the vector connecting their centroids have an absolute value lower than a certain tolerance ($2\times 10^{-5}$ is used in this work).
When such overlap occurs, we remove the second cluster and then readjust the parameters of the Gaussian mixture accordingly. 
This additional step proves to be particularly valuable during initial iterations, especially if the initial conditions are incompatible with the boundary conditions.
In such cases, the presence of strong shocks and oscillations can potentially ``confuse'' the algorithm, leading to erroneous results or even causing the numerical scheme to diverge.
The removal of overlapping clusters helps stabilize the algorithm in such scenarios, preventing unwanted issues in the early stages of the computation.

Additionally, within the main loop of our CFD solver, where \cref{alg:gmm:gmm} is used, we employ different initialization methods based on the state of the previous time step:
\begin{itemize}
    \item If no previous time step information is available, we use the k-means initialization method explained in \cref{sec:aKM}.
    \item If previous information is available for a cluster, we use it as a ``warm start'' for the initialization process.
    \item If the cluster was deleted in the previous time step, we initialize it with a random centroid and a spherical covariance matrix.
\end{itemize}
Although the simulations consistently converged to very similar results, initializing from the k-means method helped us achieve more reproducible results and draw more robust conclusions. This approach ensured greater stability in the initialization process and contributed to better consistency in the results obtained during the simulations.

\subsection{Feature space}

As mentioned above, shock waves are characterized by large gradients and oscillations. However, these features can also appear in other regions, such as turbulent and under-resolved areas. For this reason, it is essential to carefully consider the physics of the problem when choosing the gradients to define the feature space. By taking into account the specific characteristics of shock waves and distinguishing them from other phenomena, we can ensure a more accurate and reliable identification of shocks, while avoiding potential misclassifications in turbulent or under-resolved regions.

In this work we have considered:
\begin{itemize}
    \item $(\nabla\cdot \svec{v})^2$,
    \item $\lVert\nabla p\rVert^2$,
\end{itemize}
mapped into the range~$[0,1]$ in both cases\chg{, i.e. normalized}.
In shocks, pressure gradients are notably large, whereas in most regions of the flow, including boundary layers and turbulence, they are limited to much lower values. Additionally, compressibility plays a significant role in supersonic \chg{flows}. To account for this effect, we include the divergence of the velocity in the feature space. 
For the sake of completeness, \cref{sec:aVar} includes examples of alternative feature spaces.

\subsection{Sensor definition}

We aim to create the clustering sensor $s_c$ from the GMM output, focusing on obtaining \chg{node- and} element-wise values from the nodal probabilities for comparison with other sensors in the literature.
\chg{We} start by sorting the clusters based on the distance between their centroids and the origin \chg{in the feature space (cluster~$k=0$ is the closest to the origin, while cluster~$k=K$ is the furthest)}. This straightforward implementation yields good results, as all variables considered in the feature space are positively correlated with the strength of the discontinuities.
\chg{With this ordering, the nodal values of the sensor are simply the identifiers of the clusters with the highest probability, normalized to remain in the range~$s_{c,i} \in [0,1]$.
In other words, for a point~$\svec{x}_i$ of the spatial discretization, the sensor takes the value}
\begin{equation}\chgenv
\label{eq:gmm:sensor_node}
    s_{c,i} = \frac{1}{K}\argmax_k \left(\frac{f_k(\stvec{u}_i)}{\sum_{k=0}^K f_k(\stvec{u}_i)}\right).
\end{equation}
The element-wise values of the sensor are calculated by \chg{taking the largest nodal value of the sensor within each element, i.e. for element~$\Omega_e$:}
\begin{equation}\chgenv
\label{eq:gmm:sensor_elem}
    s_c = \max s_{c,i}, \quad i \,:\, \svec{x}_i \in \Omega_e.
\end{equation}
\chg{While} other possibilities\chg{---}such as \chg{averaging} over the nodes of an element\chg{---are} also valid, \cref{eq:gmm:sensor_elem} adopts a more conservative approach. Detecting a single solution point is sufficient to label an entire element as problematic. 

\section{Traditional sensors}
\label{sec:sensors}

\subsection{Modal sensor of Persson and Peraire}
\label{sub:sens:modal}

A widely used sensor within the high-order community, denoted as $s_m$, relies on a modal representation of certain flow \chg{variables.} In this approach, each mode is associated with a specific spatial frequency, and since higher frequencies correspond to larger gradients, the sensor estimates the smoothness of a scalar field $u$ based on the relative weight of its highest-frequency modes in the approximation. The expression for the modal sensor is given as \chg{follows~\cite{Persson2006}}:
\begin{equation}
\label{eq:sens:modal_sensor_log}
    s_m' = \log{\frac{\langle u_h, u_h \rangle}{\langle u, u \rangle}}.
\end{equation}
\chg{Here}, $u_h$ represents the highest frequency modes of $u$, i.e., all modes that include at least the highest mode in one of the directions. In our case, we use $u = p\rho$, and we also include tests with other variables (discussed in \cref{sec:aVar}) to present a comprehensive evaluation of the performance of the sensor.

In \cref{sub:sdisc}, we explain that the solver utilized to compute the results in this work implements a nodal formulation of the spatial discretization based on Lagrange polynomials. As a result, we first perform a basis change to express the variables of interest as a linear combination of Legendre polynomials. Both bases span the same polynomial subspace, facilitating the conversion between the two representations.

\subsection{Integral sensor}
\label{sub:sens:integral}

Considering that discontinuities introduce large gradients in the solution, we adopt a simple \chg{sensor,} denoted as $s_a$, which is based on the integral of a certain variable $u$ inside each element. The expression for the sensor is given as \chg{follows~\cite{mateogabin2022entropy}}:
\begin{equation}
\label{eq:sens:avg_sensor}
    s_a' = \frac{\sqrt{\langle u,u \rangle}}{V}. 
\end{equation}

In our results, we specifically use $u = \lVert\nabla p\rVert$, but we also include some tests with other potential choices in \cref{sec:aVar}.

\subsection{Sensor scaling}

The sensors presented in \cref{sub:sens:modal,sub:sens:integral} are not confined to the interval $[0,1]$, unlike our proposed GMM sensor. To address this, we apply the \chg{following scaling technique to the raw sensor values~\cite{Persson2006}}:
\begin{equation}
\label{eq:sens:sensor_scaling}
    s = \left\{
    \begin{array}{ll}
        0 & \text{if}\;\; s' < s_0 - \Delta s, \\
        1 + \sin\frac{\pi(s' - s_0)}{2\Delta s} & \text{if}\;\; s_0 - \Delta s \leq s' \leq s_0 + \Delta s, \\
        1 & \text{if}\;\; s' > s_0 + \Delta s.
    \end{array}
    \right.
\end{equation}
\chg{The} parameters $s_0$ and $\Delta s$ serve as the center and width of the mapping from the original interval of $s'$ to the final range $s \in [0,1]$. This scaling ensures that all sensors, regardless of their original value range, are transformed to the interval $[0,1]$, allowing meaningful and consistent comparisons between different sensor outputs.

\section{Numerical discretization: HORSES3D}
\label{sec:horses}

The main objective of our research is to develop a sensor capable of detecting shock waves in both low- and high-order \chg{Navier--Stokes} solvers\chg{.
To be able to test the methodology of \cref{sec:gmm}}, we integrate our sensor with the open-source software HORSES3D~\cite{Ferrer2023} \chg{and couple it with two common stabilization approaches to provide examples of possible applications in the context of shock capturing.
The first strategy was already described in \cref{sec:ns}, and consists in the introduction of the artificial viscosity of \cref{eq:ns:gpflux} into the NS equations, modulated in intensity with the help of a sensor.
The second approach will be explained in \cref{sub:sdisc,sec:aDisc}, and consists in a node-wise blending of a high-order approximation and a finite volume scheme controlled by our proposed sensor.
This computational setup} allows us to assess the performance of the sensors in the context of a complete simulation framework. Such an integrated approach provides valuable insights into the effectiveness and robustness of the sensor in detecting shock \chg{waves.}

\subsection{Spatial discretization}
\label{sub:sdisc}

In this study, we adopt a discontinuous \chg{Galerkin~\cite{Black1999,kopriva2009implementing}} (DG) approach to discretize the spatial terms of \cref{eq:ns:advdiff}. Our physical domain is represented by a mesh of non-overlapping elements, and within each element, we approximate the values of various magnitudes using piecewise polynomials of order $P$. Since the solution is discontinuous, we introduce numerical fluxes to facilitate the transfer of information across neighboring elements. These fluxes serve as mathematical approximations to the Riemann problem generated at the discontinuities, ensuring that information is exchanged in a physically meaningful manner.

Our chosen approximation basis is tensor-product Lagrange polynomials, which means that the degrees of freedom in this scheme are simply the nodal values. Specifically, for the \chg{Navier--Stokes} equations described in \cref{sec:ns}, each node is assigned a value of the state vector $\stvec{q}_{ijk} = (\rho_{ijk}, \rho\svec{v}_{ijk}, \rho e_{ijk})^T$.

\chg{Although the details of the mathematical derivation are provided in \cref{sec:aDisc}, we briefly introduce in this section the second stabilization approach that we employ to obtain some of the results of \cref{sec:results}.
The methodology of \cref{sec:ns} consisted in the addition of an artificial term that introduces more dissipation into the mathematical formulation.
Conversely, the approach that we show herein does not modify the original equations and instead, adds dissipation by combining high- and low-order methods to evaluate the spatial semi-discretization.
In particular, the combination of our DG method with a finite volume scheme enables us to introduce dissipation at every node by means of Riemann solvers.
The idea is based on the telescopic form of the derivative operators of the DG method used in this work, providing a finite volume--like expression of the DG high-order operator,~$\mathbb{D}^{\text{DG}}$.
In a one-dimensional case, this means that the derivative at any point can be computed as~\cite{Fisher2013}
\begin{equation*}
    \mathbb{D}^{\text{DG}}_i(\stvec{f}) = \frac{\hat{\stvec{f}}_{(i,i+1)} - \hat{\stvec{f}}_{(i-1,i)}}{\omega_i},
\end{equation*}
for a certain definition of the sub-cell fluxes~$\hat{\stvec{f}} = \hat{\stvec{f}}^{\text{DG}}$ (see \cref{sec:aDisc} for the complete explanation of this expression).
However, these fluxes can also be computed using other approaches, such as a more dissipative finite volume scheme.
Therefore, a hybrid DG-FV formulation can be used to stabilize high-order approximations when the sub-cell fluxes are combined as~\cite{HENNEMANN2021109935,RUEDARAMIREZ2022105627,lin2023high}
\begin{equation}
\label{eq:sdisc:hybrid}
\begin{gathered}
    \hat{\stvec{f}}_{(i,i+1)} = \left[1 - \alpha_{(i,i+1)}\right] \hat{\stvec{f}}^{\text{DG}}_{(i,i+1)} + \alpha_{(i,i+1)} \hat{\stvec{f}}^{\text{FV}}_{(i,i+1)}, \\
    \alpha_{(i,i+1)} \in [0, \alpha_{\max}],
\end{gathered}
\end{equation}
with~$\alpha_{\max} \in [0,1]$.
We employ this approach in \cref{sub:sedov,sub:dmr}, using the values of our GMM-based sensor to compute the blending coefficients~$\alpha_{(i,i+1)}$ associated to every sub-cell flux:
\begin{equation}
\label{eq:sdisc:alpha}
    \alpha_{(i,i+1)} = \max(s_i, s_{i+1}),
\end{equation}
where~$s_i$ and~$s_{i+1}$ are the values of the sensor at the nodes~$i$ and~$i+1$, respectively.}

\subsection{Temporal discretization}
\label{sub:tdisc}

For temporal integration, we have chosen the well-known \emph{Strong Stability Preserving \chg{Runge--Kutta}} method, SSPRK33, proposed by Shu and \chg{Osher.~\cite{shu1988efficient}}
This three-stage \chg{Runge--Kutta} method is of third order and is well-suited for our \chg{purposes.
To} further enhance the stability of the time integration without significantly reducing the \chg{time-step} size, we incorporate the positivity-preserving limiter developed by Zhang and Shu~\cite{Zhang2011} at the end of every stage of the SSPRK33 integrator.
\chg{This method rescales the polynomial approximation in an element around the average when the density or the pressure become excessively small.
Specifically, defining~$\varepsilon(u)$ as
\begin{equation*}
    \varepsilon(u) = \min(\bar{u}, \varepsilon^{\star}),
\end{equation*}
with~$\bar{u}$ the average value of~$u$ in an element and~$\varepsilon^{\star}$ a user-defined parameter, the limiter ``shrinks'' the density approximation whenever the minimum value of~$\rho$ is below~$\varepsilon(\rho)$, and modifies the entire state vector when the pressure reaches values below~$\varepsilon(p)$.
The exact implementation used in this work can be found in the code repository provided at the end of the article.}
The combination of artificial viscosity with this limiter has shown promising results in stabilization.
On the one hand, the limiter can prevent negative values of density and pressure when a certain CFL condition is met; however, it does not eliminate oscillations from the solution. Additionally, it often requires excessively small time steps after several iterations. On the other hand, the artificial viscosity approach negatively impacts the viscous CFL number, resulting in the smearing of oscillations and the imposition of a smaller time step.
By utilizing both methods together, we benefit from the additional dissipation provided by the artificial flux, which effectively eliminates oscillations, while the limiter allows for the use of larger time steps. This combined approach strikes a balance between stability and efficiency, resulting in a more robust and accurate simulation of the flow dynamics.

\section{Results}
\label{sec:results}

In this section, we conduct \chg{an} analysis of our GMM sensor as described in \cref{sec:gmm}, \chg{combining it with different stabilization strategies and comparing its performance against} other well-known sensors outlined in \cref{sec:sensors}.
The results presented herein are computed using the open-source software HORSES3D~\cite{Ferrer2023} and are based on the discretization described in \cref{sec:horses}, along with the following key components.

\chg{We} employ the SSPRK33 method (detailed in \cref{sub:tdisc}) \chg{for} temporal discretization.
Note that HORSES3D employs a non-dimensional formulation of the \chg{Navier--Stokes} equations; therefore, the magnitudes given in this and the following sections are also non-dimensional.
For the viscous terms, we utilize the BR1~\cite{Bassi1997} scheme with entropy gradients, incorporating central numerical fluxes for gradients and viscous fluxes.
To enhance the robustness of the simulations, we adopt the split-form of Chandrashekar~\cite{Chandrashekar2013} to discretize the advection term.
The Riemann solver for inter-element fluxes is the two-point flux of Chandrashekar, complemented with additional dissipation from \chg{a matrix method.~\cite{Ismail2009,Chandrashekar2013}}

\chg{Although} the tests are two-dimensional, the high Reynolds numbers involved ensure the appearance of turbulent regions. It is important to note that turbulence is inherently a three-dimensional phenomenon, and thus these two-dimensional regions do not fully capture the complete behavior of the flow. However, for the purpose of our study, these regions suffice, as they introduce gradients that must be differentiated from the ones representing shocks.

With this comprehensive setup, we proceed to \chg{assess} the performance of our novel sensor \chg{with four test cases.
In all of them, our proposed sensor is updated every ten time steps to save on computational time (we further discuss this strategy in \cref{sub:perf}).
The results of \cref{sub:sedov,sub:dmr} showcase the nodal resolution of the sensor of \cref{eq:gmm:sensor_node}, capable of detecting only the degrees of freedom that require stabilization.
For these two sections we use the hybrid DG-FV approach of \cref{sub:sdisc}, with~$\alpha_{\max}=0.5$ in \cref{eq:sdisc:alpha}, and computing the finite volume stabilization fluxes,~$\hat{\stvec{f}}^{\text{FV}}$, with the same Riemann solver that we employ for the inter-element numerical fluxes.
In \cref{sub:inviscid,sub:viscous} we compare our GMM-based sensor} against other established sensors, providing valuable insights into the effectiveness and reliability of our proposed approach in various scenarios. \chg{In this cases, however, the classical sensors that we use for comparison only return element-wise values and therefore, we employ the maximum nodal value in each element as the final sensor value (see \cref{eq:gmm:sensor_elem}).}

\subsection{\chg{Sedov blast}}
\label{sub:sedov}

\chg{This first numerical experiment simulates the evolution of a two-dimensional explosion.
Beginning with a zero-velocity flow field, and a smooth peak in density and pressure around the origin,}
\begin{equation*}\chgenv
\begin{gathered}
    \rho(t=0) = 1 + \mathcal{N}_1(r; 0.25), \\
    p(t=0) = 10^{-2} + \mathcal{N}_1(r; 0.15), \\
    \mathcal{N}_1(r; \sigma) = \frac{1}{4 \pi \sigma^2} \exp \left(-\frac{r^2}{2\sigma^2} \right), \\
    r^2 = x^2 + y^2,
\end{gathered}
\end{equation*}
\chg{the flow develops a shock wave that propagates radially.
The setup enables us to assess the behavior of our sensor in smooth flows and with fast-moving shock waves.}

\chg{We solve the flow in a square domain with sides of length 2, and divide it into $64\times 64$ elements.
The solution is approximated by polynomials of order~$P=4$, leading to 102,400 degrees of freedom.
All the boundaries are set as slip walls, i.e. symmetry conditions, generating reflections towards the end of the simulation at~$t=1.5$.
The solver is run for 3,000 iterations with a time-step size of $\Delta t = 5\times 10^{-4}$, for a maximum CFL number based on the distance~$h$ of \cref{eq:ns:gpvisc} of CFL$_i \approx 0.01$.
The minimum values of density and pressure allowed by the limiter are determined by~$\varepsilon^{\star}=10^{-13}$.}

\begin{figure*}[htpb]\chgenv
    \subfloat[$t = 1.0$]{
    \begin{minipage}{0.3\textwidth}
        \includegraphics[width=\textwidth]{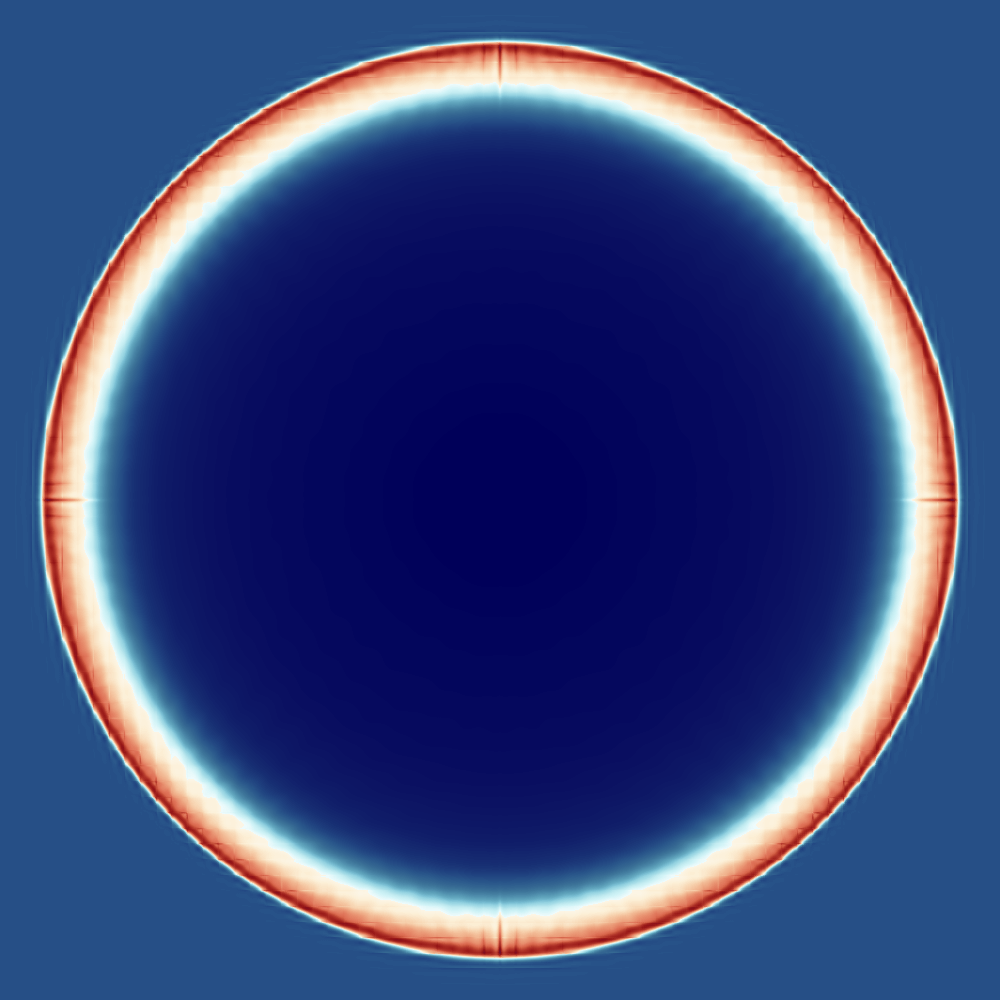}
    \end{minipage}\qquad
    \begin{minipage}{0.3\textwidth}
        \includegraphics[width=\textwidth]{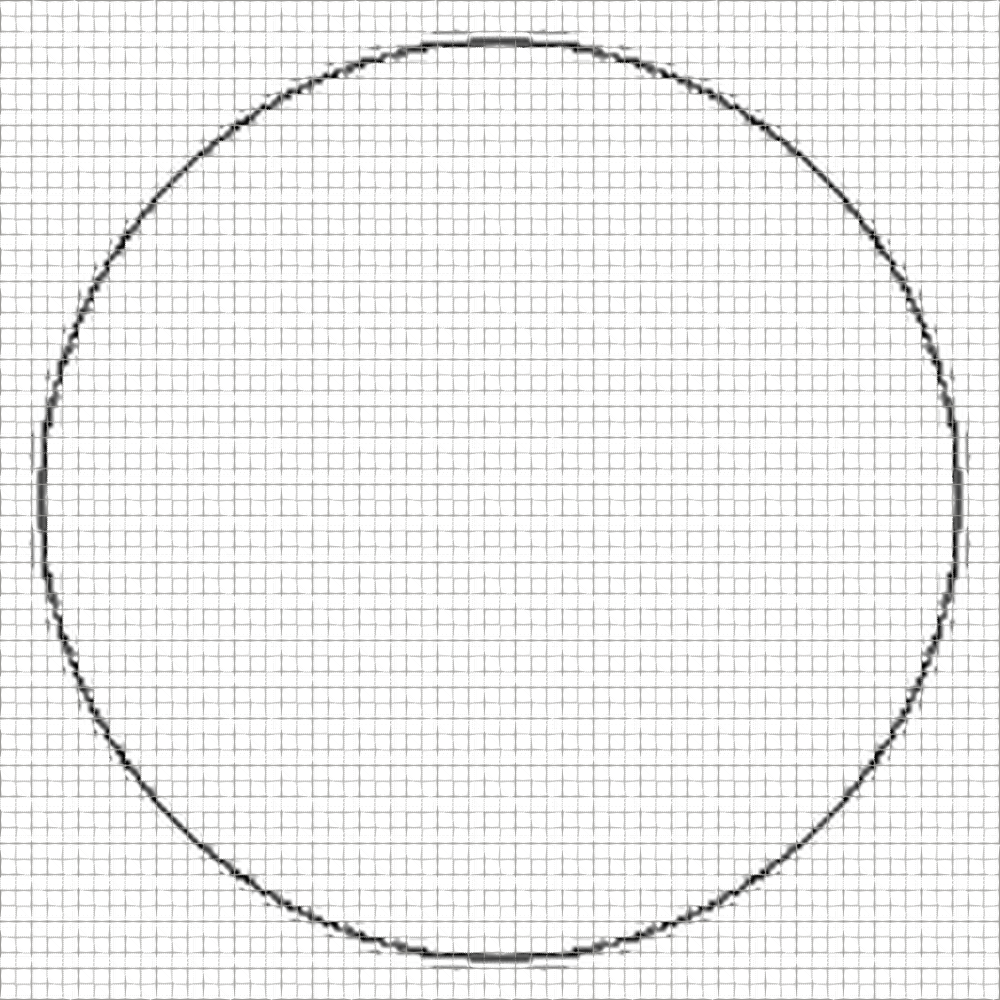}
    \end{minipage}
    }\\\medskip
    \subfloat[$t = 1.5$]{
    \begin{minipage}{0.3\textwidth}
        \includegraphics[width=\textwidth]{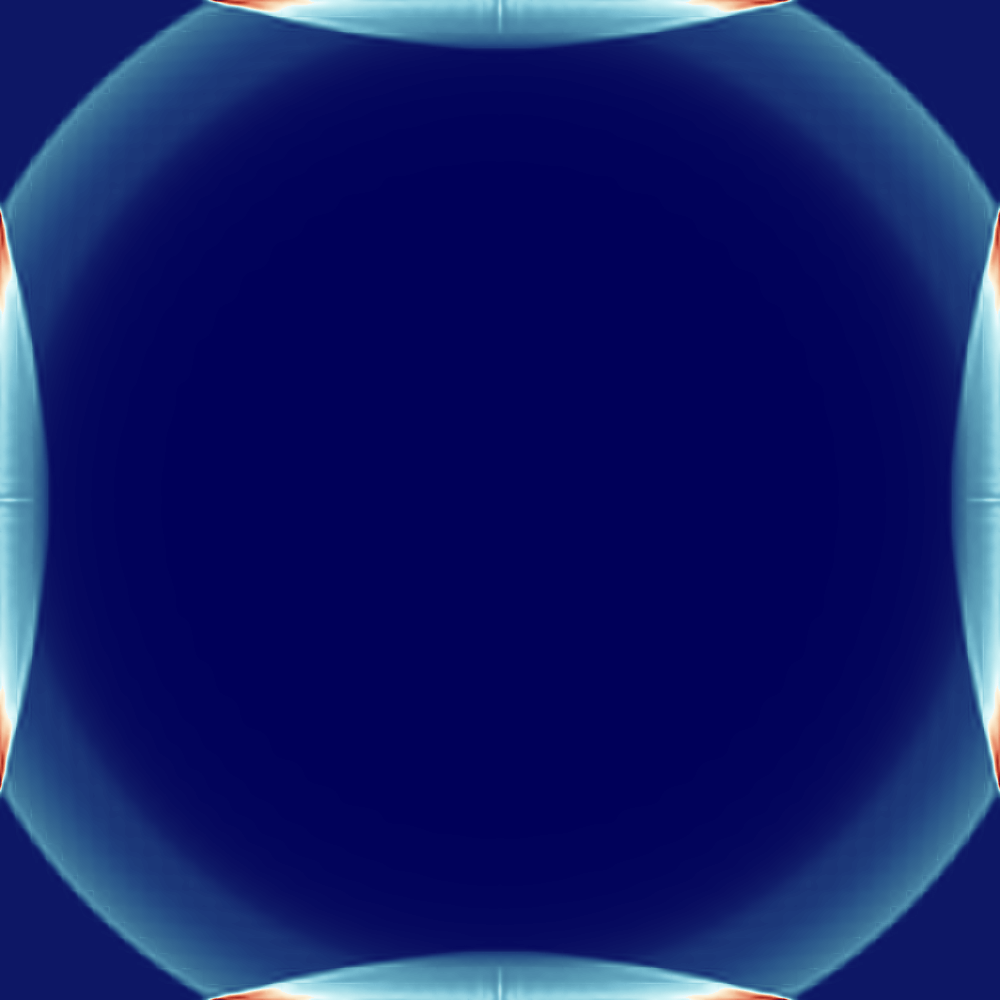}
    \end{minipage}\qquad
    \begin{minipage}{0.3\textwidth}
        \includegraphics[width=\textwidth]{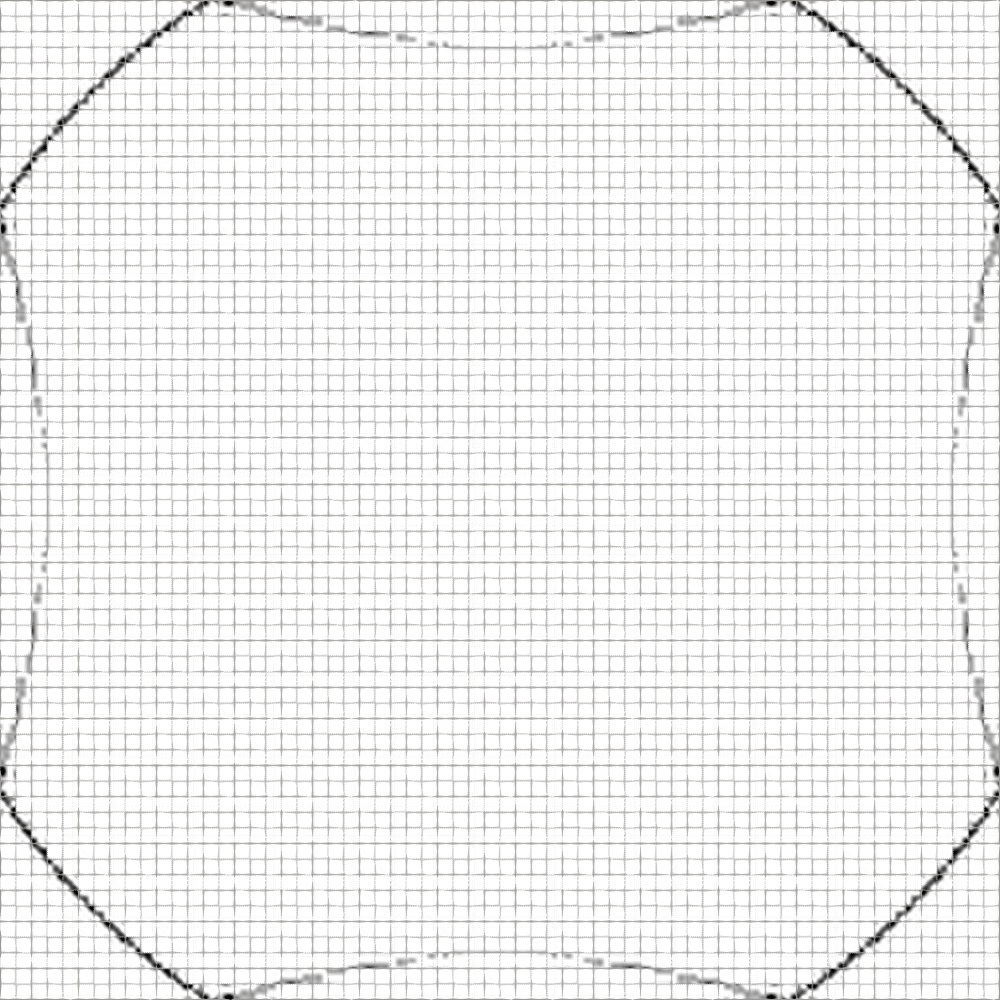}
    \end{minipage}
    }
    \caption{Density field (left) and nodal GMM sensor (right) of \cref{eq:gmm:sensor_node} applied to the Sedov blast case, using $\lVert\nabla p\rVert^2, (\nabla\cdot \svec{v})^2$ and four clusters.}
    \label{fig:res:blast_shock}
\end{figure*}

\chg{The sensor correctly detects the position of the shock inside the elements that contain it, as presented in \cref{fig:res:blast_shock}, and the four clusters provide a certain degree of resolution regarding the intensity of the discontinuity.
Although the shock is moving, the use of an explicit time-stepping approach imposes strong constraints on the size of the time steps, and the ten-iterations delay in the computation of the sensor does not affect the final results.}

\chg{Unfortunately, the sensing algorithm that we propose always uses normalized variables and thus, it has no information to determine if discontinuities are present in the flow field.
For this reason its use is limited to cases where the flow always contains shock waves, as in a flow without them the sensor will still classify regions according to the intensity of the gradients.}
\begin{figure}[htpb]\chgenv
    \centering
    \begin{minipage}{0.3\textwidth}
        \includegraphics[trim=180 180 180 180,clip,width=\textwidth]{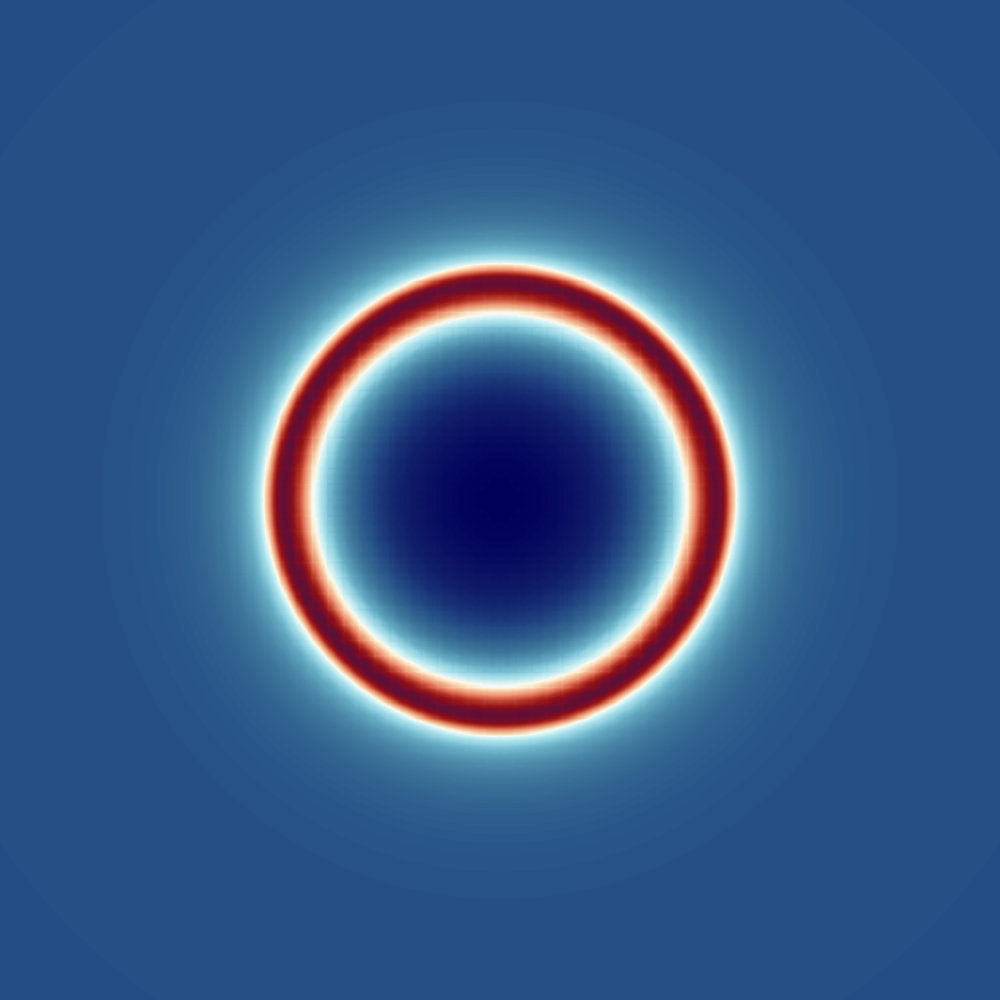}
    \end{minipage}\\\medskip
    \begin{minipage}{0.3\textwidth}
        \includegraphics[trim=180 180 180 180,clip,width=\textwidth]{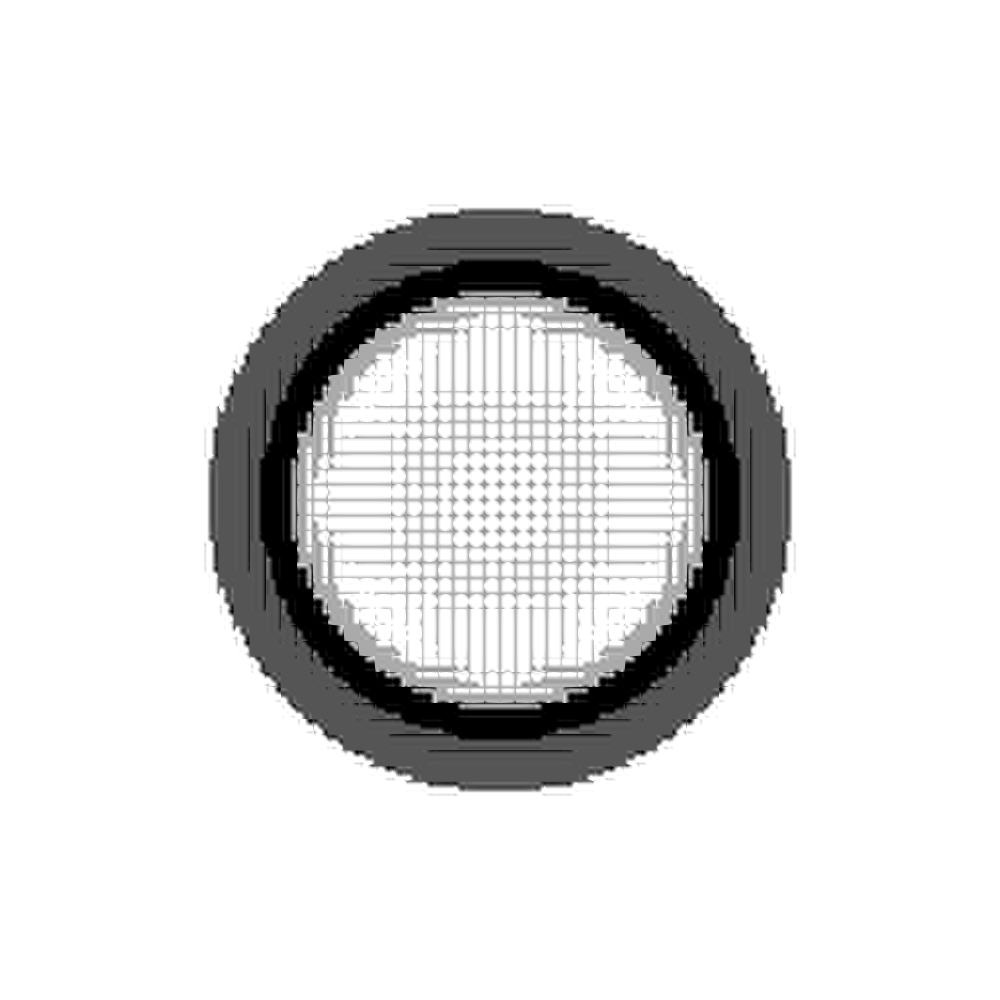}
    \end{minipage}
    \caption{Density field (top) and GMM sensor (bottom) at~${t=0.25}$ of the Sedov blast case.}
    \label{fig:res:blast_smooth}
\end{figure}
\chg{This is clearly seen in \cref{fig:res:blast_smooth}, where the initial Gaussian distribution of density has evolved into a ring-shaped structure but has not yet formed a discontinuity at the front.
When the sensor is used for shock capturing in a smooth case, its impact on the final solution is determined by the stabilization method employed.
In this particular simulation, sensed regions are blended with a sub-cell finite volume scheme.
Consequently, the advection operator is more dissipative and the scheme is less accurate, but still conserves its sub-cell resolution.
Although in this work we are not concerned about the performance of the sensor in cases without shocks, we understand that this issue must be addressed before adopting this approach for shock capturing in more generic cases.
Indicators based on the intensity of the gradients can mitigate it, but more research needs to be done as such approaches also increase the complexity of the algorithm.}

\subsection{\chg{Double Mach reflection}}
\label{sub:dmr}

\chg{When a moving shock wave encounters a wedge, the supersonic flow is reflected, forming a characteristic pattern before reaching a stationary state.
The double Mach reflection test case simulates the evolution of the flow when a Mach 10 shock wave hits a~$30^{\circ}$ wedge.}

\chg{To simplify the geometry of the case, the spatial domain is a rectangle with dimensions~$3.25\times 1$, and the surface of the wedge covers the bottom boundary for~${x \in [1/6, 3.25]}$.
Thus, the reference frame is rotated and the shock wave travels diagonally, entering the domain from the top-left corner and moving according to the following expression:}
\begin{equation}\chgenv
\label{eq:res:wave_pos}
    x_w(y,t) = \frac{1}{6} + y \tan \phi + \frac{10 t}{\cos \phi}, \quad \phi = \frac{\pi}{6}.
\end{equation}
\chg{Discarding the influence of the wedge, the position of any point on the surface of the discontinuity at a time~$t$ is given by \cref{eq:res:wave_pos} as $(x_w(y,t), y)$.}

\chg{The domain is tessellated into~$117\times 36$ square elements, all of the same size.
Using polynomials of degree~$P=4$ to approximate the solution, this numerical setup contains 105,300 degrees of freedom.
The surface of the wedge is modeled as a slip wall, and the rest of the boundaries directly impose the freestream conditions given by \cref{eq:res:wave_pos,tab:res:dmr}.}
\begin{table}[htpb]\chgenv
    \centering
    \caption{Freestream conditions on both sides of the incoming shock wave in the double Mach reflection case.~\cite{HENNEMANN2021109935}}
    \begin{ruledtabular}
    \begin{tabular}{ccc}
        Variable & $x \leq x_w$ & $x > x_w$ \\
        $\rho$ & 8 & 1.4 \\
        $u$ & 7.145 & 0 \\
        $v$ & -4.125 & 0 \\
        $p$ & 116.5 & 1 \\
    \end{tabular}
    \end{ruledtabular}
    \label{tab:res:dmr}
\end{table}
\chg{As this case involves higher speeds, by~$t=0.2$ the shock has already covered most of the domain and consequently, the time-step size must also be smaller than in the previous test.
We use~$\Delta t = 5\times 10^{-6}$ (maximum CFL$_i \approx 0.01$) and run 40,000 iterations, with the high-order limiter set to~$\varepsilon^{\star}=10^{-13}$.}

\begin{figure}[htpb]\chgenv
    \centering
    \begin{minipage}{0.48\textwidth}
        \includegraphics[width=\textwidth]{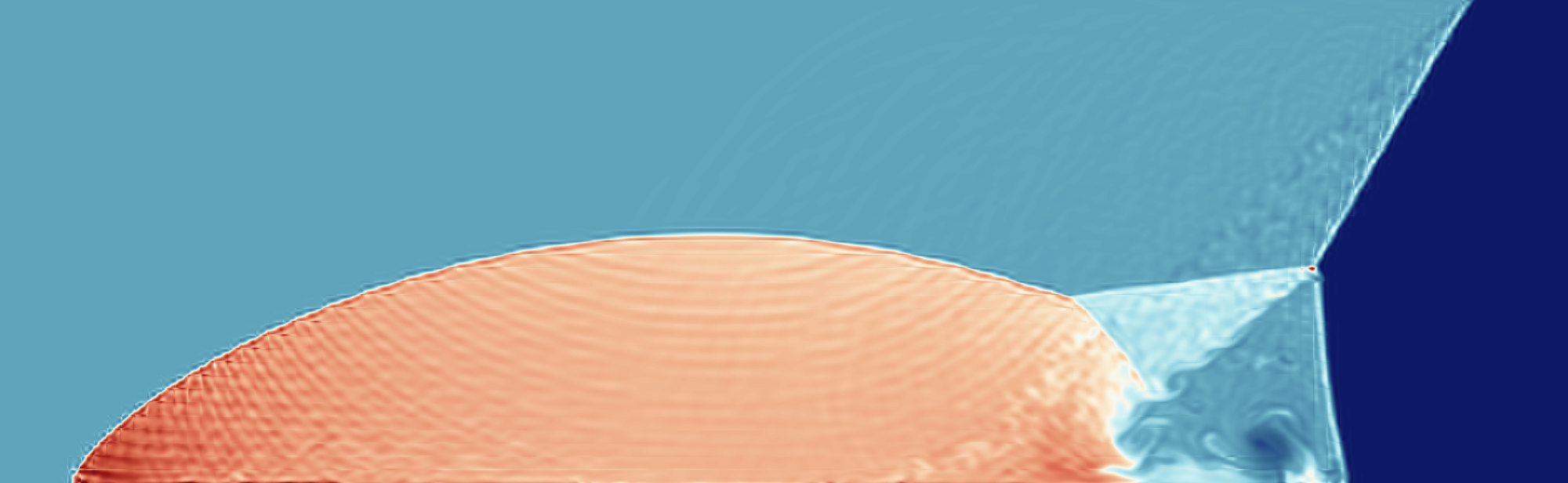}
    \end{minipage}\\\medskip
    \begin{minipage}{0.48\textwidth}
        \includegraphics[width=\textwidth]{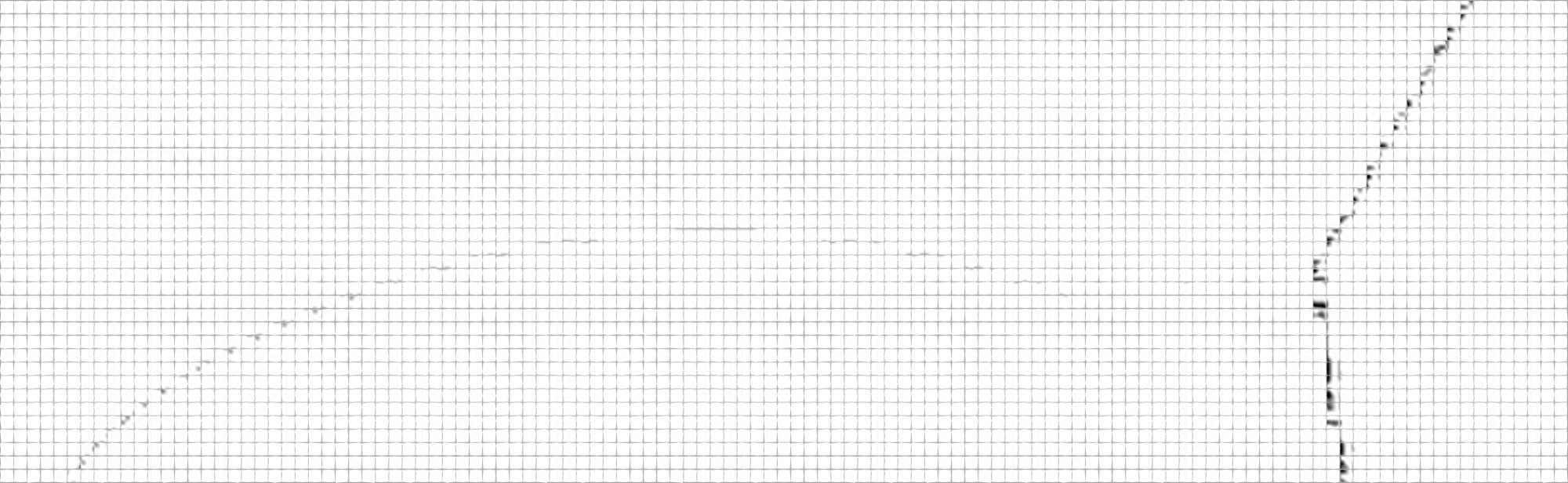}
    \end{minipage}
    \caption{Density field (left) and nodal GMM sensor (right) of \cref{eq:gmm:sensor_node} applied to the double Mach reflection case, using $\lVert\nabla p\rVert^2, (\nabla\cdot \svec{v})^2$ and four clusters.}
    \label{fig:res:dmr}
\end{figure}

\chg{As showcased in \cref{fig:res:dmr}, the discontinuities of this case are stronger and the hybrid stabilization approach is not able to remove all the oscillations.}
\chg{Consequently, additional dissipation is not added across the entire shock wave, but only at the specific nodes where oscillations are strong enough to destabilize the simulation.
Although the element-wise formulation of \cref{eq:gmm:sensor_elem} is more stable, we show this case because it proves that our GMM approach can also be of use in non-optimal setups.}

\chg{The results of \cref{fig:res:dmr} are computed using four clusters, and the solution at most of the points is calculated with the purely high-order scheme.
The few nodes that need stabilization are correctly detected around the contour of the system of shock waves that appears, where the largest discontinuities are located.
In addition, the sensor is able to separate the two main types of shocks that evolve in this case: the strong initial shock wave that moves to the right at Mach 10 and the weaker one that, given enough time, forms an oblique shock at the front of the wedge.
As less dissipation is introduced in the domain, the vortical structures of the region of interaction between the main shock wave and the wedge do not vanish, and the sub-cell resolution of the DG scheme is fully exploited to capture some of theses vortices.
Apart from the unwanted oscillations, no other non-physical behavior is observed.
In particular, the beginning of the shock at $\svec{x}=(1/6,0)$ is smooth, and the interaction point between the initial fast-moving wave and the rest of the features is stabilized throughout the entire simulation.}

\subsection{Inviscid flow around a cylinder at Mach 3}
\label{sub:inviscid}

The objective of this test is to evaluate the performance of the sensors in a purely hyperbolic scenario, where boundary layers are absent and large gradients consistently indicate the existence of discontinuities or regions of strong turbulence.
\chg{We therefore utilize a two-dimensional setup,~\cite{ryujin-2021-1}} featuring a Mach 3 flow inside a planar channel encountering a cylindrical obstacle.

\begin{figure}[htpb]
    \centering
    \includegraphics[width=0.45\textwidth]{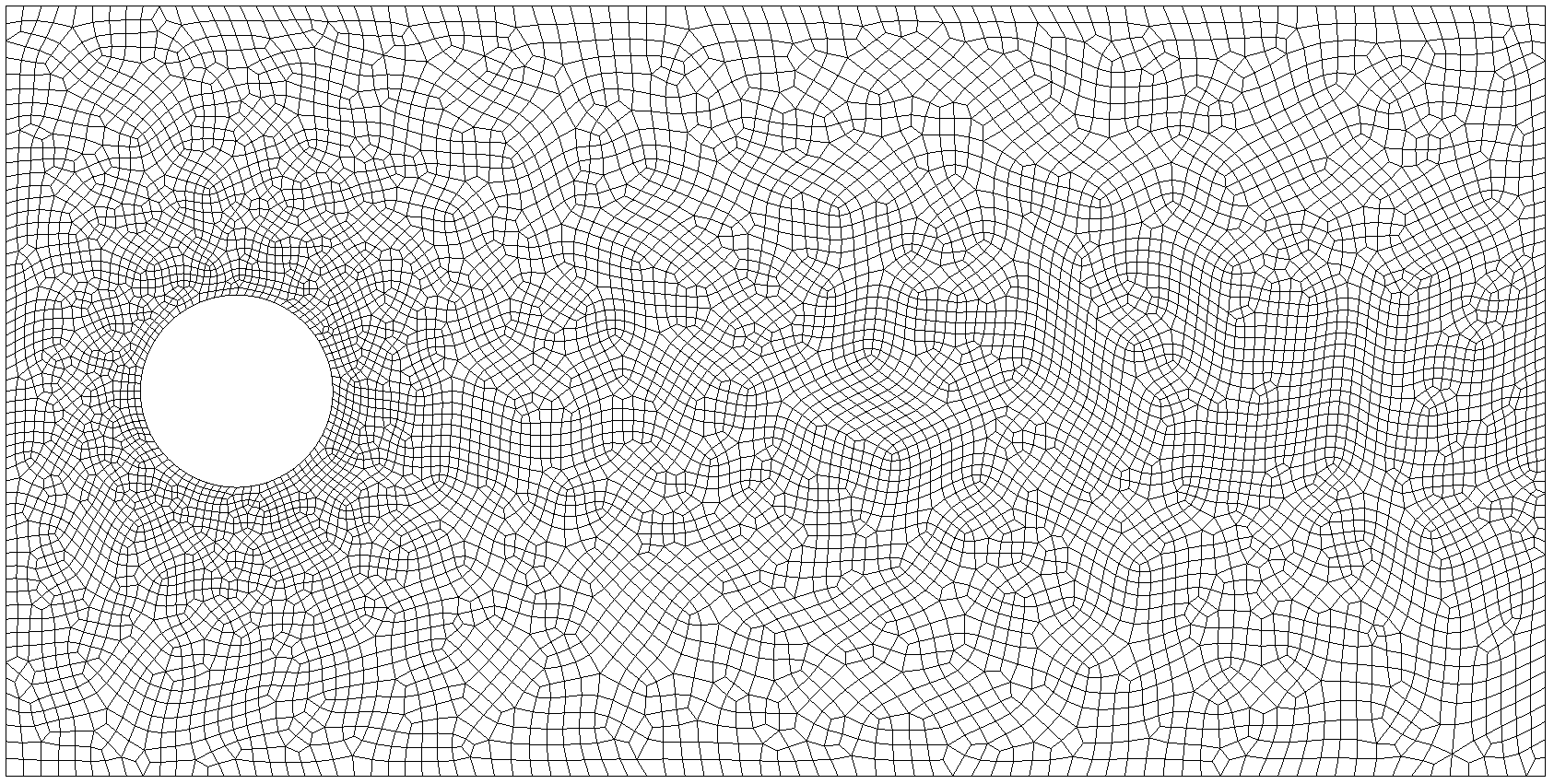}
    \caption{Mesh used to compute the flow around a Mach 3 cylinder with no viscosity.}
    \label{fig:res:mesh_inviscid}
\end{figure}

For the simulations, we employ a rectangular, unstructured mesh with boundaries $\svec{x} \in [-1.2,6.8] \times [-2,2]$, and place a cylinder of diameter one at the center, $\svec{x}_c = (0,0)$. As illustrated in \cref{fig:res:mesh_inviscid}, the domain is divided into 8,145 elements, with smaller sizes implemented near the cylinder and the wake region to improve resolution. The mesh was generated with GMSH v4.11, and second-order elements are used to properly describe the surface of the cylinder. Within each element, we approximate the solution using polynomials of order 4, resulting in a total of 203,625 degrees of freedom.
\chg{Time is descretized in 300,000 intervals of size $\Delta t = 2\times 10^{-4}$, representing a maximum CFL$_i \approx 0.025 - 0.1$ depending on the element size.
At every stage of the time integrator, the limiter ensures the positivity of the density and pressure using~$\varepsilon^{\star}=10^{-5}$.
Although higher than in the previous cases, this value is well below the minima of the flow field.
It is, however, required to stabilize the simulations during the initial iterations.}

\begin{figure*}[htpb]
    \subfloat[]{
        \includegraphics[width=0.4\textwidth]{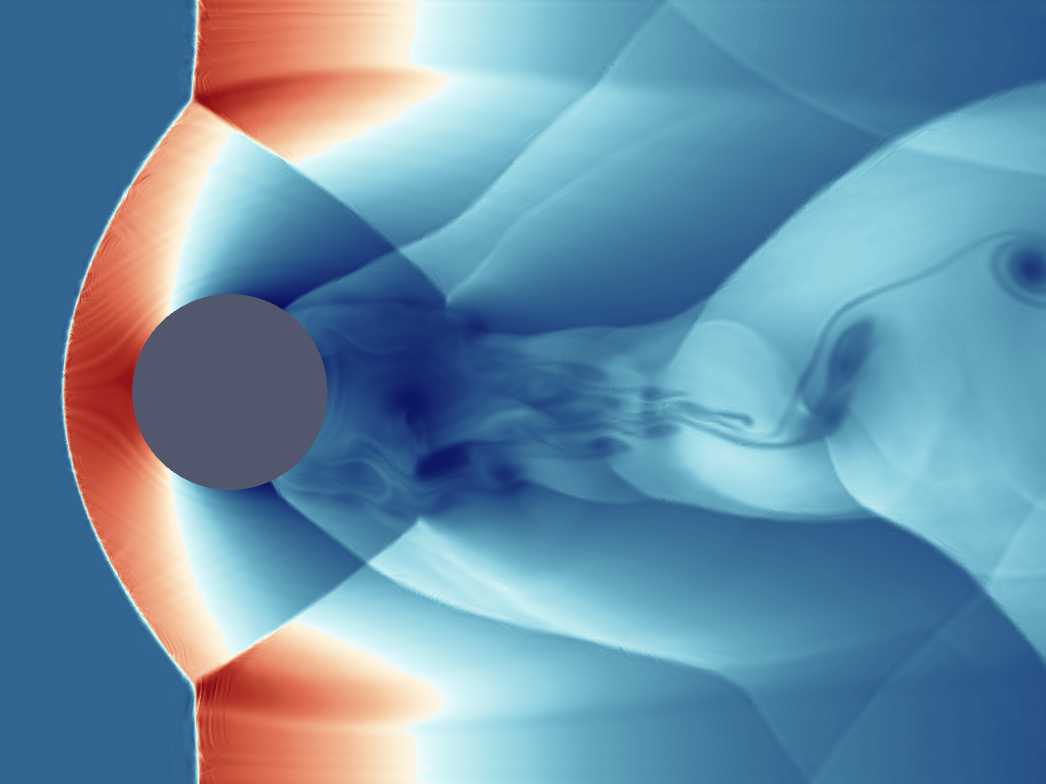}
    }\qquad
    \subfloat[]{
        \includegraphics[width=0.4\textwidth]{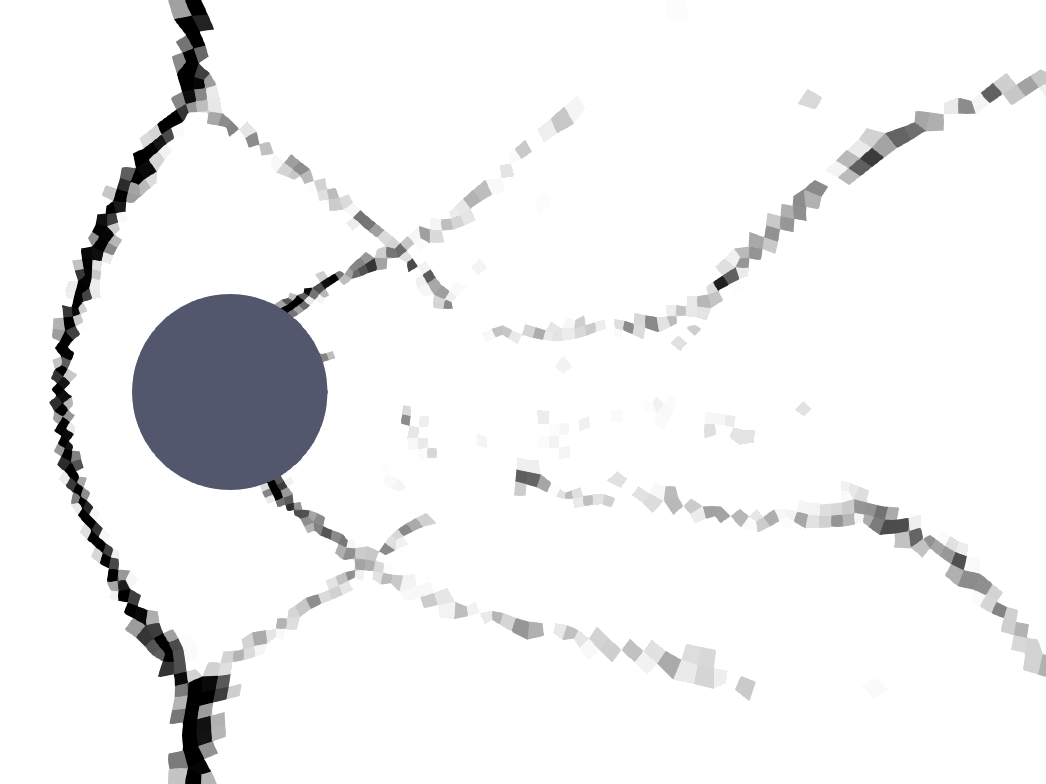}
    }\\
    \subfloat[]{
        \includegraphics[width=0.4\textwidth]{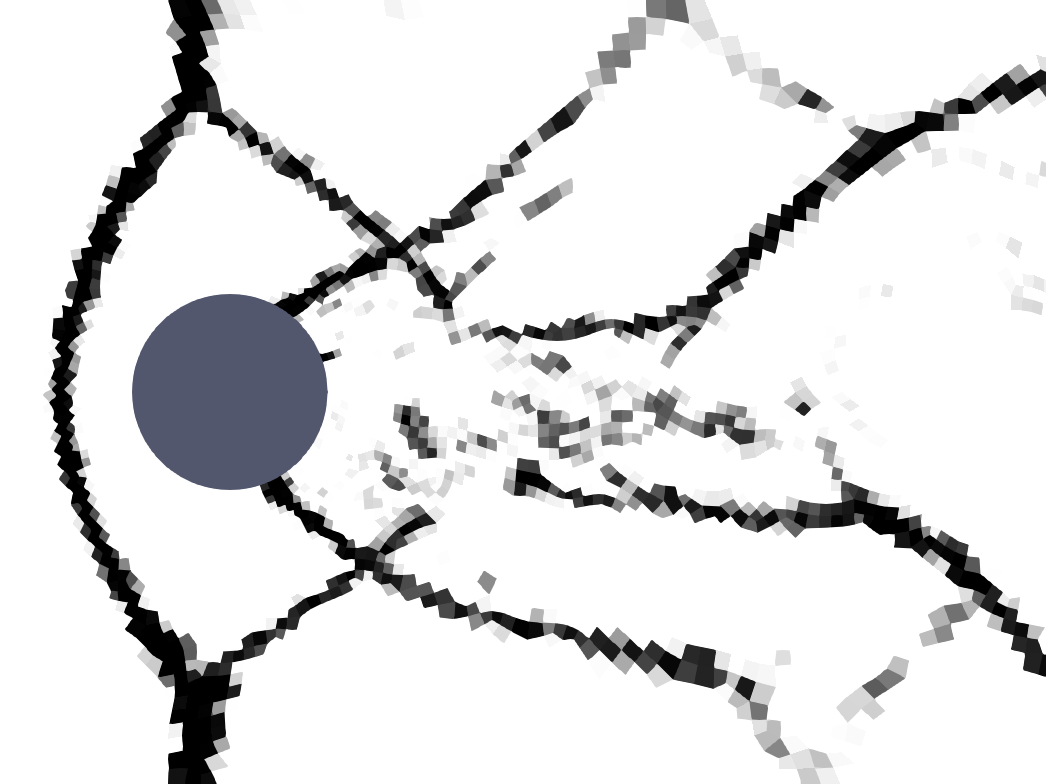}
    }\qquad
    \subfloat[]{
        \includegraphics[width=0.4\textwidth]{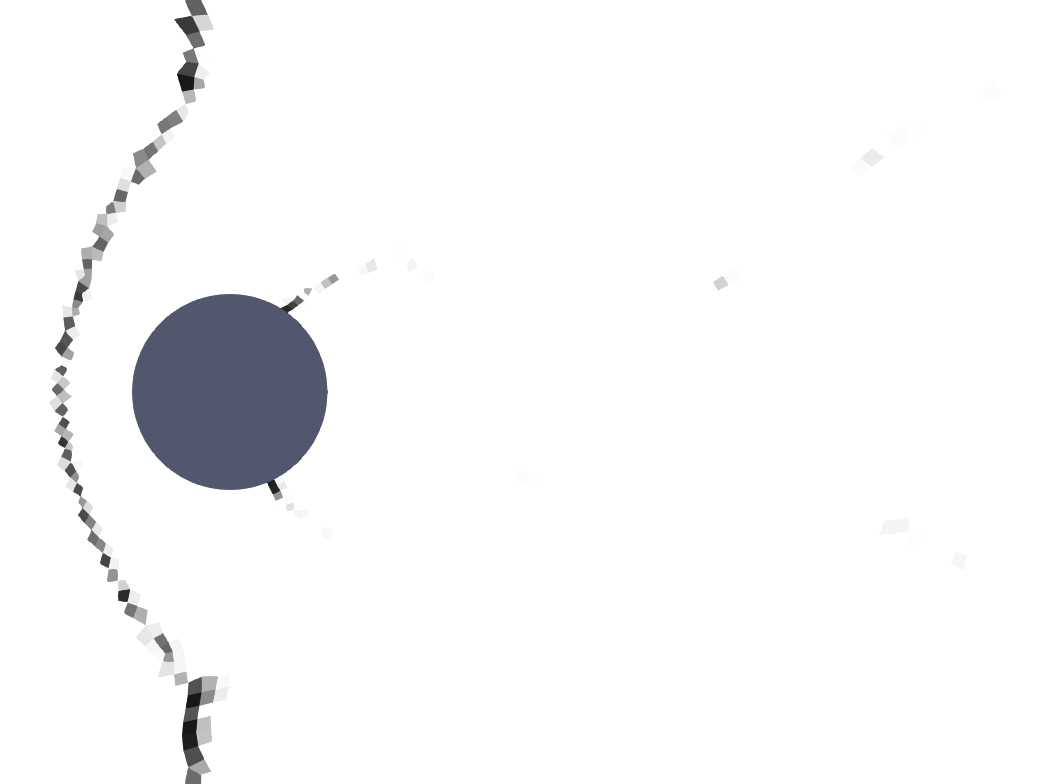}
    }
    \caption{Inviscid case after 300,000 iterations with the modal sensor of \cref{sub:sens:modal}, using $p\rho$. a) density field, b) sensor with $s_0 = -2.5$ and $\Delta s = 1$. Sensor applied to the last iteration with $s_0 = -3.5$, $\Delta s = 1$ (c), and with $s_0 = -1.5$, $\Delta s = 1$ (d).}
    \label{fig:res:inviscid_modal}
\end{figure*}

\begin{figure*}[htpb]
    \subfloat[]{
        \includegraphics[width=0.4\textwidth]{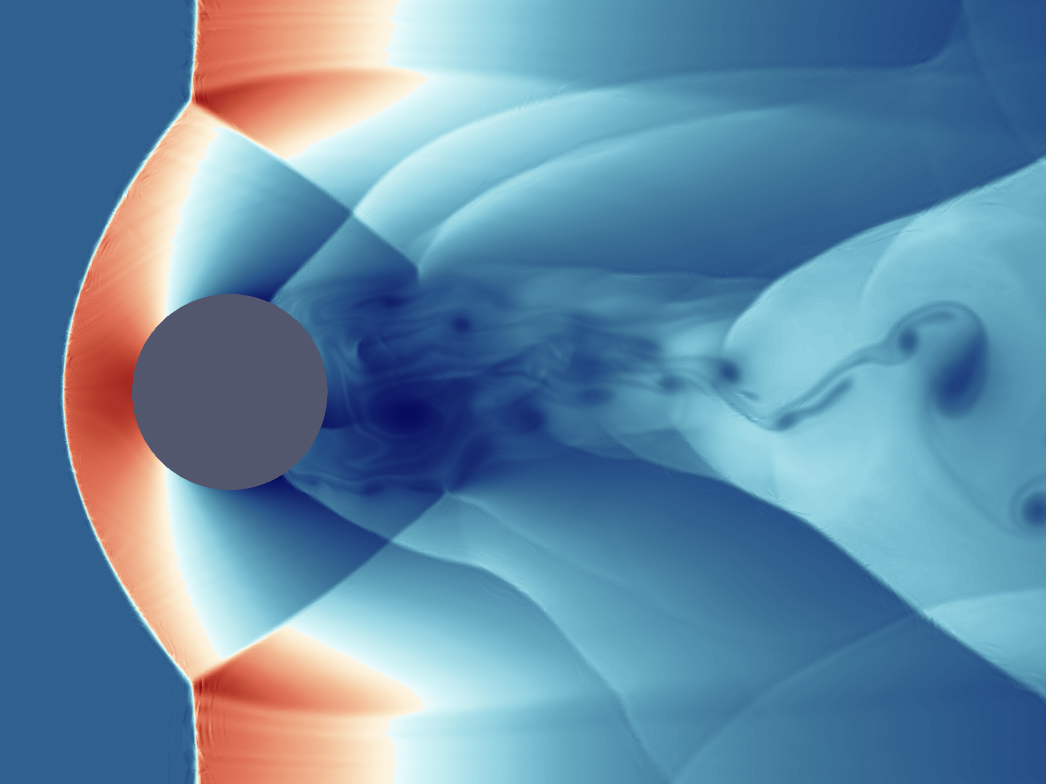}
    }\qquad
    \subfloat[]{
        \includegraphics[width=0.4\textwidth]{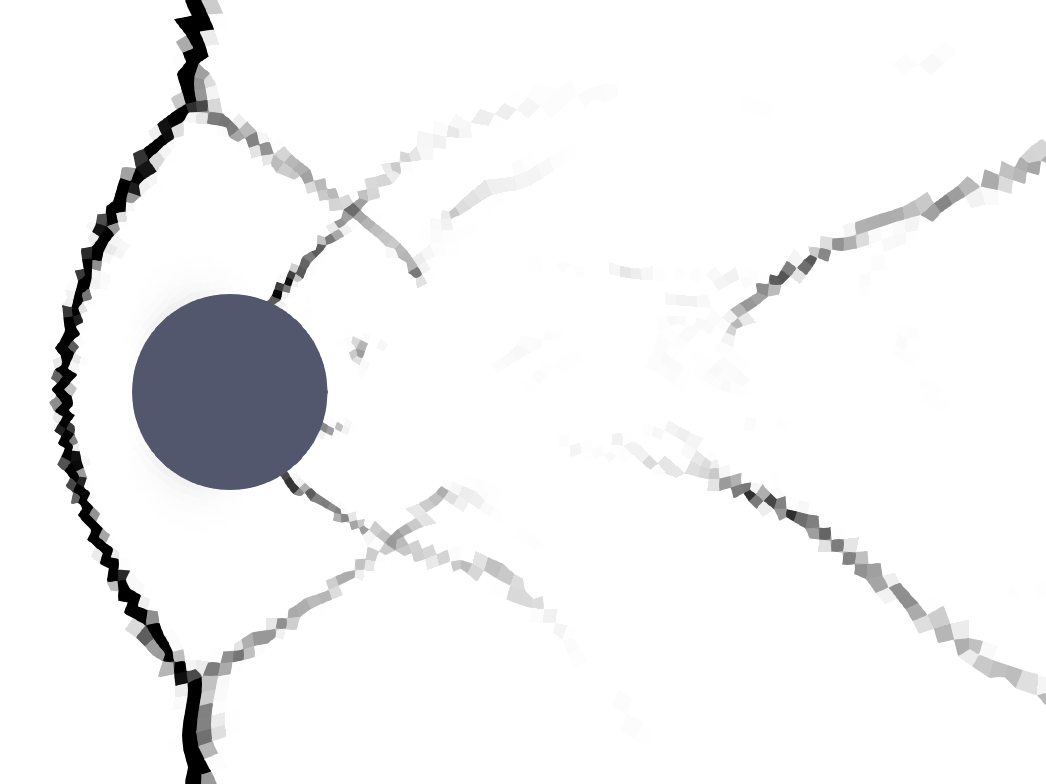}
    }\\
    \subfloat[]{
        \includegraphics[width=0.4\textwidth]{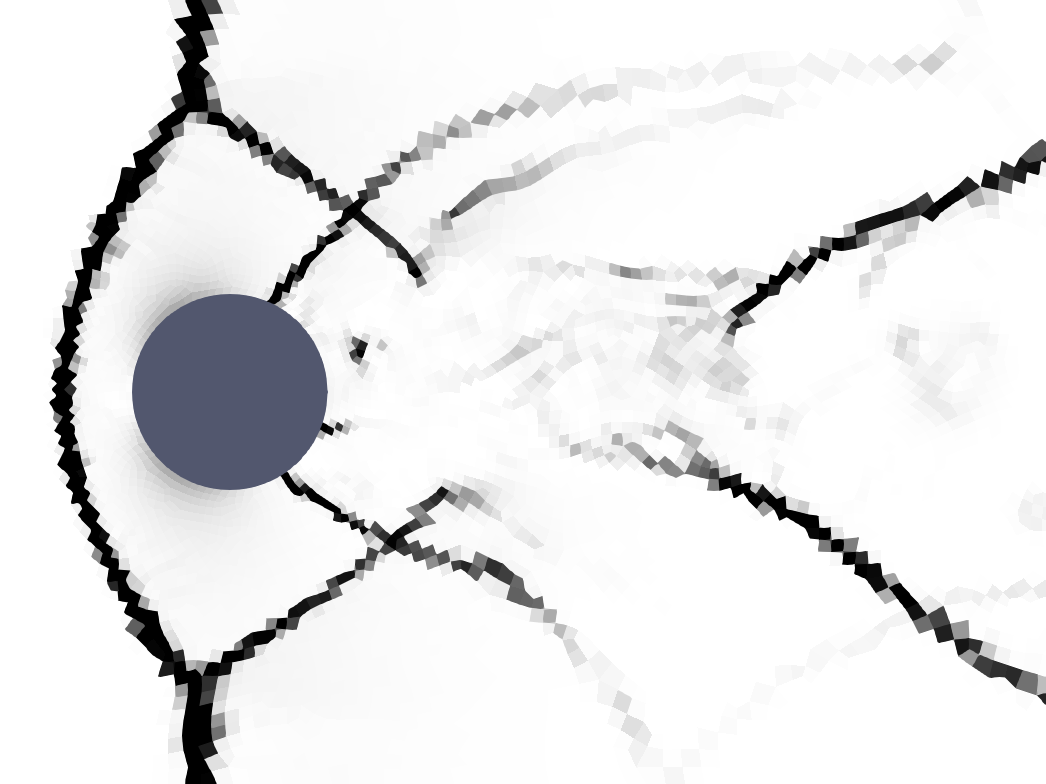}
    }\qquad
    \subfloat[]{
        \includegraphics[width=0.4\textwidth]{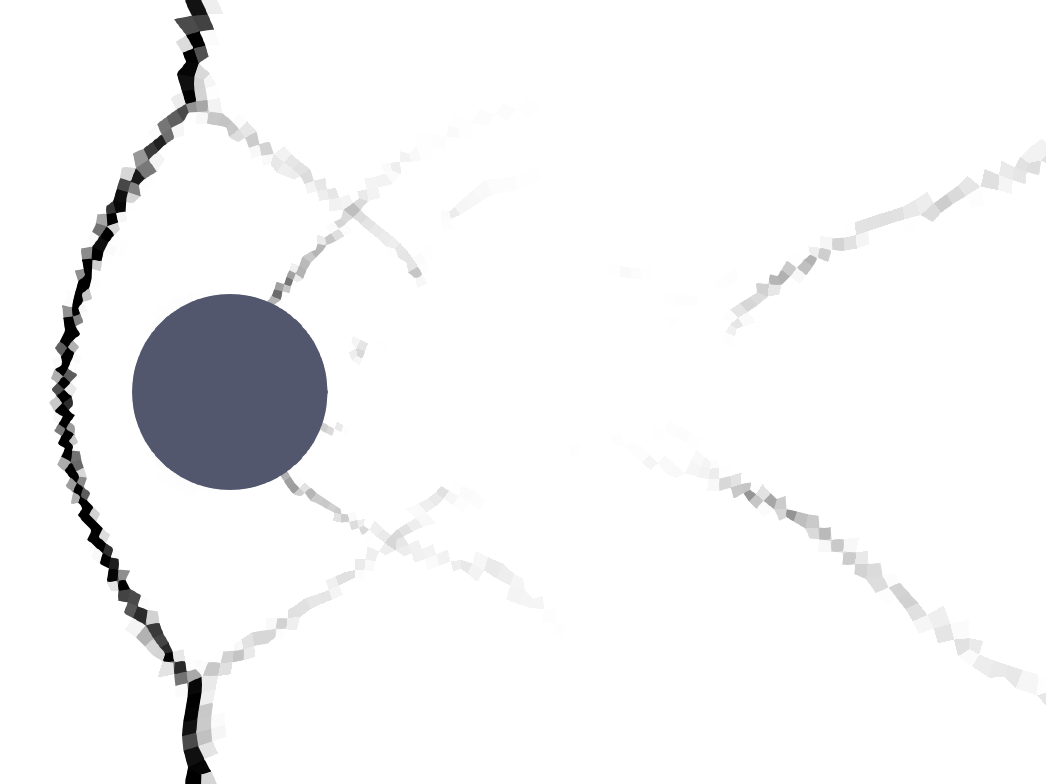}
    }
    \caption{Inviscid case after 300,000 iterations with the integral sensor of \cref{sub:sens:integral}, using $\lVert\nabla p\rVert^2$. a) density field, b) sensor with $s_0 = 5.25$ and $\Delta s = 4.75$. Sensor applied to the last iteration with $s_0 = 2.5$, $\Delta s = 2.5$ (c), and with $s_0 = 8$, $\Delta s = 7$ (d).}
    \label{fig:res:inviscid_integral}
\end{figure*}

\begin{figure*}[htpb]
    \subfloat[]{
        \includegraphics[width=0.4\textwidth]{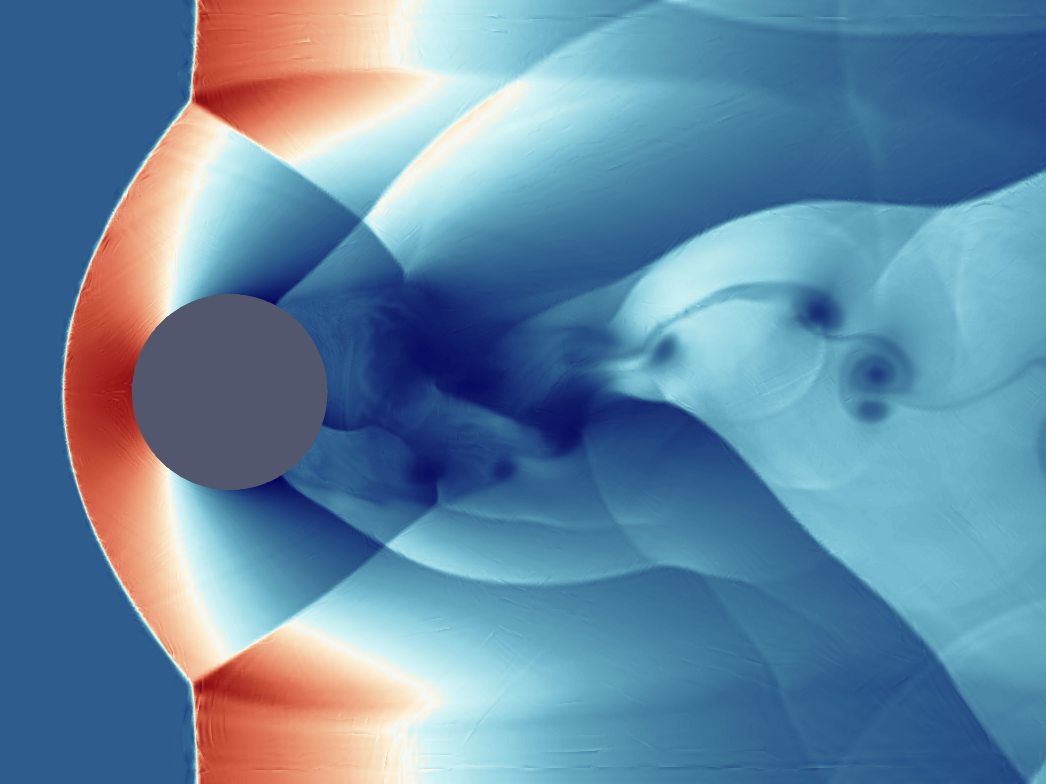}
    }\qquad
    \subfloat[]{
        \includegraphics[width=0.4\textwidth]{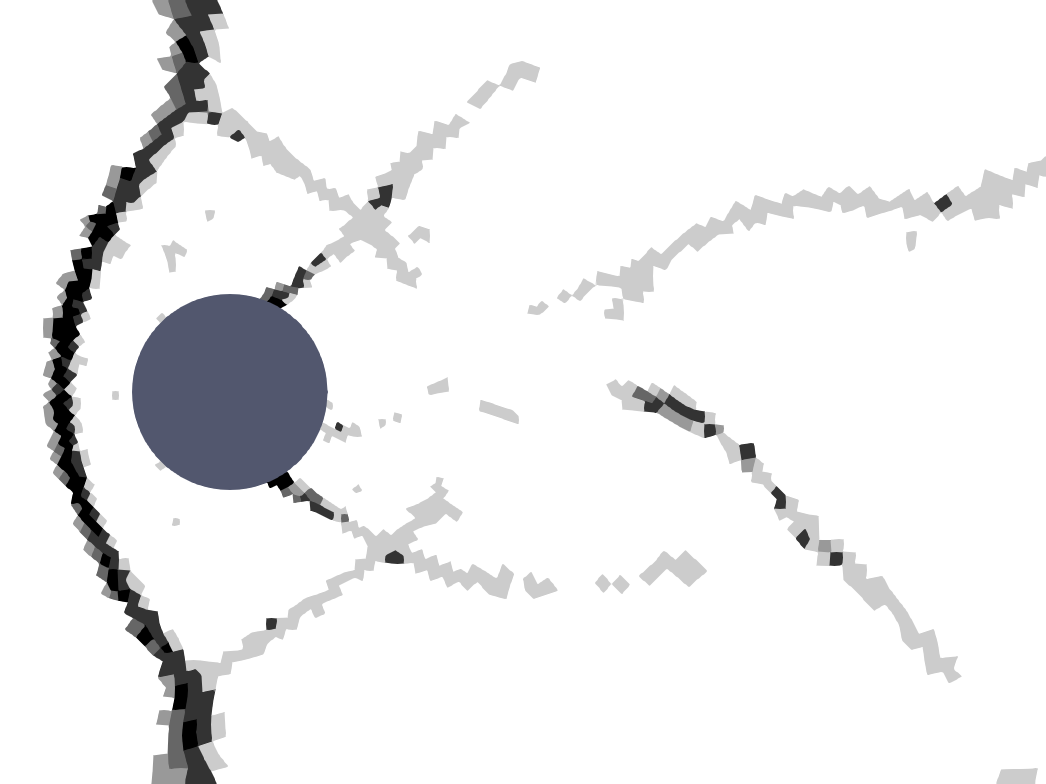}
    }
    \caption{Inviscid case after 300,000 iterations with our adaptive GMM sensor of \cref{sec:gmm}, using $\lVert\nabla p\rVert^2, (\nabla\cdot \svec{v})^2$. a) density field, b) sensor with six clusters.}
    \label{fig:res:inviscid_gmm}
\end{figure*}

For the boundary conditions, we enforce slip-wall (symmetry) conditions at the top and bottom boundaries, as well as on the surface of the cylinder.
\chg{On the left side, we implement an inflow condition at Mach 3, while an outflow condition is applied on the right side using Riemann invariants.
If the Mach number normal to the boundary is higher than one, no information travels inwards and the state on the outer side of the boundary faces is copied from the interior.
Conversely, for lower Mach numbers, the outer state is~$\stvec{q}=(\rho_0,\rho\svec{v}_0,\rho e_0)$ with\cite{Carlson2011,Mengaldo2014,mateogabin2022entropy}
\begin{equation*}
\begin{gathered}
    \rho_0 = \rho \left( 1 + \frac{p_0/p-1}{\gamma} \right), \\
    \svec{v}_0 = \svec{v}_t + \svec{v}_{0,n}, \\
    \svec{v}_{0,n} = r^+ - \frac{2c_0}{\gamma-1}, \quad r^+ = \svec{v}_n + \frac{2c}{\gamma-1}, \\
    \rho e_0 = \frac{p_0}{\gamma - 1} + \frac{1}{2}\rho_0\lVert\svec{v}_0\rVert^2.
\end{gathered}
\end{equation*}
where~$p_0$ is the exterior pressure,~$c$ and~$c_0$ are the speed of sound on both sides of the boundary, and~$\svec{v}_t$ is the tangent component of the velocity.}

The artificial viscosity is defined with a constant value of $\mu_0 = 0.1$. To compute the final viscosity, we scale it based on the sensor value and the mesh size, following \cref{eq:ns:gpvisc}. This approach allows us to dynamically adjust the viscosity to account for regions with varying levels of discontinuities and turbulence.
\chg{With this scaling, the viscous CFL number ranges between CFL$_v \approx 3\times 10^{-3}$ and CFL$_v \approx 6\times 10^{-4}$.
Therefore, the additional viscosity does not impose further constraints to the size of the time steps.}

\begin{table}[htpb]
    \centering
    \caption{Parameters of the sensors for the inviscid cylinder.}
    \begin{ruledtabular}
    \begin{tabular}{lccc}
        Sensor & $s_0$ & $\Delta s$ & \# of clusters \\
        \hline
        Modal & -2.5 & 1 & - \\
        Integral & 5.25 & 4.75 & - \\
        GMM & - & - & 6 \\
    \end{tabular}
    \end{ruledtabular}
    \label{tab:res:inviscid_params}
\end{table}

All simulations exhibit similar overall behavior; however, slight differences in the artificial viscosity among the different methods result in the triggering of turbulence in slightly distinct ways. As a consequence of the chaotic nature of turbulence, the instantaneous snapshots presented in \cref{fig:res:inviscid_modal,fig:res:inviscid_integral,fig:res:inviscid_gmm} show notable discrepancies in the wake region after $t = 60$. This phenomenon is well known, and it is why turbulence is typically evaluated in terms of averaged quantities.

However, in our case, the primary interest lies in effectively capturing shocks rather than analyzing these turbulent effects. Therefore, our figures showcase instantaneous snapshots that allow for a straightforward assessment of the performance of the sensor, particularly in terms of detecting and representing shocks in the flow field.

The three methods demonstrate good performance with the values specified in \cref{tab:res:inviscid_params} (sub-figures a and b), effectively disregarding almost the entire wake region and accurately detecting the main shock waves. In this scenario, the absence of a boundary layer around the cylinder delays the detachment point, causing the flow to accelerate until it reaches Mach numbers similar to those at the inlet. Subsequently, the flow experiences a sudden deceleration, eventually reaching a near-zero velocity just behind the cylinder.

The most challenging region of this test case is the strong shock that appears at the detachment point. This shock is responsible for the development of the wake downstream, where it interacts with the reflections of the main shock wave generated in front of the cylinder. Capturing and representing this region accurately is crucial to the overall fidelity of the simulation results.

Our novel GMM sensor performs exceptionally well in this test case. The automatic clustering capability of the GMM enables it to adapt to the feature space effectively.
It can accurately detect regions of decreasing intensity near shocks while also correctly grouping all smooth regions into the same cluster.
As a result, the sensor provides a smoother transition between detected and undetected regions, which is a crucial aspect in element-wise artificial viscosity models since discontinuities in the viscosity field can introduce \chg{errors.~\cite{barter2010shock}}

Furthermore, the sensor distribution obtained using the GMM-based algorithm shows excellent agreement with the distributions from the modal and integral sensors. This highlights a significant advantage of our method: it requires minimal parameter tuning since the algorithm is adaptive and regions are consistently and accurately identified without the need for fine-tuning.

The only parameter that needs to be introduced is the number of clusters, and its impact is minimal since the algorithm automatically adjusts to the data distribution. Having more clusters results in smoother transitions between smooth and non-smooth regions, but the sensor remains effective with a relatively small number of clusters.
We examine this statement in\ \cref{sub:stats}, demonstrating that dissipation is only added where necessary across a wide range of cluster numbers.

To illustrate the importance of finding appropriate values for the sensor thresholds, we modify the parameter~$s_0$ and~$\Delta s$ of \cref{eq:sens:sensor_scaling} as shown in the c and d sub-figures of \cref{fig:res:inviscid_modal,fig:res:inviscid_integral}. This emphasizes the need to carefully select sensor thresholds to ensure accurate identification and representation of shocks and other flow features.

\subsection{Viscous flow around a cylinder at Mach 2}
\label{sub:viscous}

This test introduces several characteristics of real flows, making it more complex than the previous case. In the previous test, viscosity was only included in shocks, which was shown to be critical for stabilizing under-resolved features and obtaining a robust scheme. However, in this case, the flow is viscous everywhere, with a Reynolds number of~$10^5$ based on the diameter of the cylinder.

The main objective of this test is to assess the performance of our sensor in simulations with boundary layers, as their proper representation is crucial for the overall stability of the discretization. While shock sensors are designed to detect discontinuities, they may also identify other under-resolved regions of the flow. In the context of turbulent flows, this may lead to a failure to match the real energy spectrum. However, for boundary layers, the challenge lies in maintaining robustness in the numerical solution.

Elements in boundary layers are typically stretched significantly to resolve normal gradients while minimizing the number of degrees of freedom. However, this approach impacts negatively the maximum allowed time step, which is already small in high-order schemes with explicit time integration. The introduction of artificial viscosity into these elements worsens the situation, resulting in even shorter time steps.

In this test case, we aim to examine how our sensor performs in the presence of boundary layers and ensure that it effectively identifies and treats under-resolved regions while maintaining the overall stability and accuracy of the simulation.

\begin{figure}[htpb]
    \centering
    \includegraphics[width=0.4\textwidth]{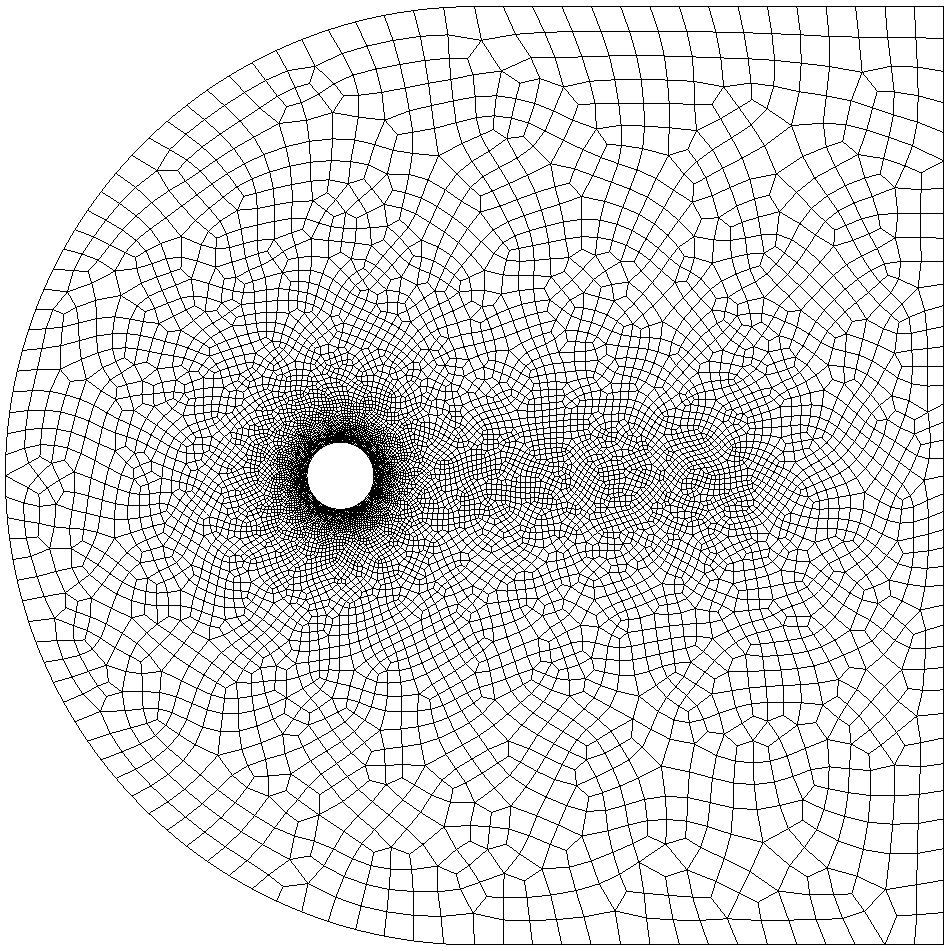}
    \caption{Mesh used to compute the flow around a Mach 2 cylinder at $\text{Re}=100,000$.}
    \label{fig:res:mesh_viscous}
\end{figure}

The mesh, as shown in \cref{fig:res:mesh_viscous}, represents a free flow at Mach 2 with a cylinder placed at the origin with a diameter of one.
The unstructured mesh has bounds $x \in [-5,9]$ and $y \in [-7,7]$, and was generated using GMSH v4.11.
It comprises 9,713 elements of second order, concentrated around the cylinder and in the wake region.
The polynomial approximation used is of order four, resulting in a total of 242,825 degrees of freedom.
\chg{Time integration is performed in 300,000 steps of size $\Delta t = 2\times 10^{-4}$, with a maximum inviscid CFL$_i \approx 0.16$ for the smaller elements.
Considering the Reynolds number of this case, the viscous CFL number has a maximum value CFL$_v \approx 0.12$.
Nevertheless, since this estimates consider the characteristic length introduced in \cref{eq:ns:gpvisc}, these values may be misleading in highly-stretched elements, and dissipative effects can have a relevant influence in the maximum allowed time-step size.
As in the inviscid case of the previous section, the high-order limiter is configured with $\varepsilon^{\star}=10^{-5}.$}

We apply inflow boundary conditions on the semicircular left side of the mesh and outflow boundary conditions on the top, right, and bottom sides. The surface of the cylinder is modeled with a no-slip boundary condition. Since the flow is viscous in this case, we found that a viscous constant~$\mu_0 = 0.08$ for \cref{eq:ns:gpvisc} provides the best results.
\chg{This additional viscosity can increment the previous CFL estimates in $\Delta$CFL$_v \approx 0.004$.
Although the change is small in comparison, the shape of the mesh elements is not fully considered in this calculations, and the dissipation introduced in regions of the flow with low speeds and deformed elements can represent an important constraint.}

\begin{figure}[htpb]
    \centering
    \includegraphics[width=0.48\textwidth]{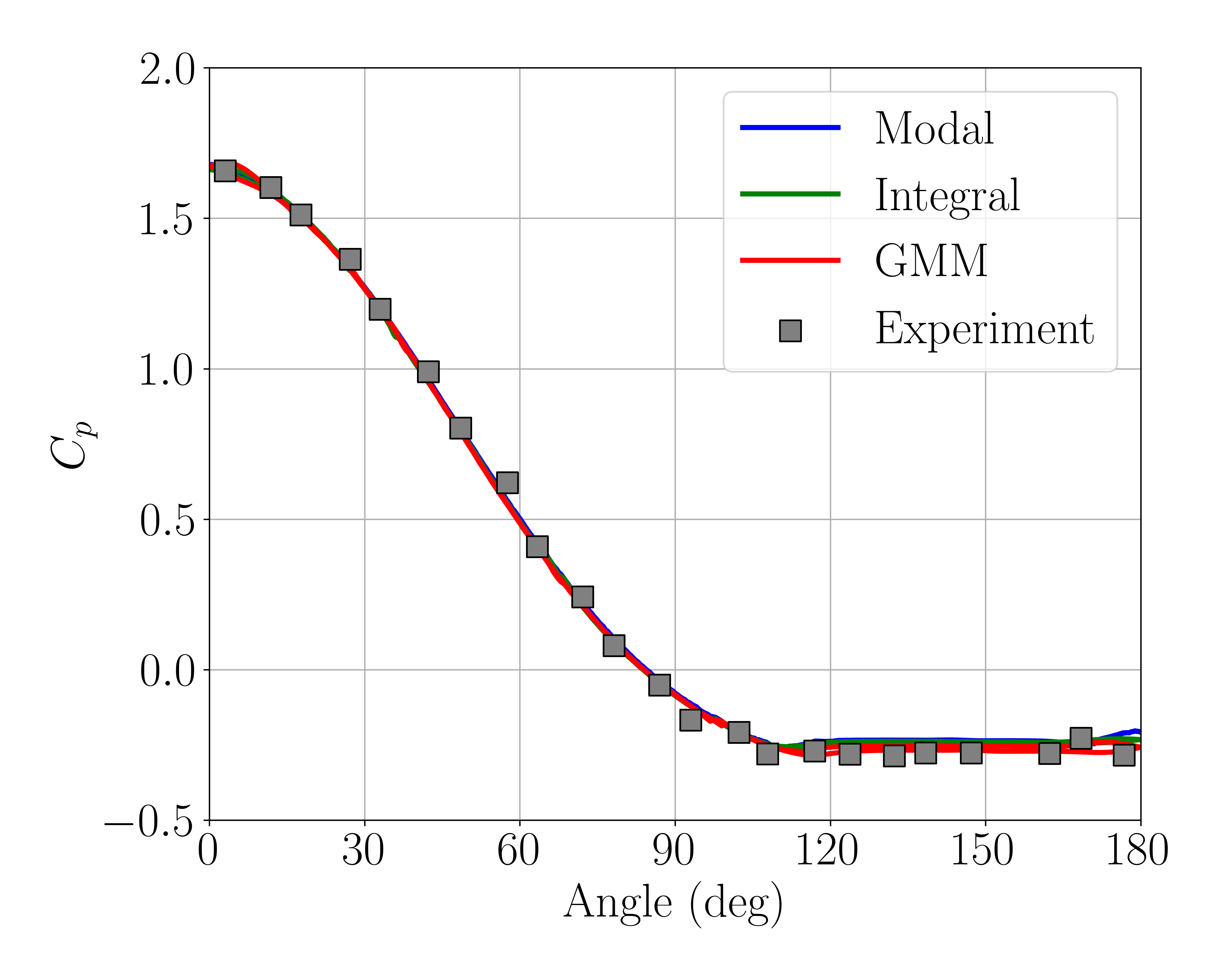}
    \caption{\chg{Experimental~\cite{Gowen1952} and numerical (averaged over 20 time units) pressure coefficient around the Mach 2 cylinder}. The angle is measured from the stagnation point, i.e. the left-most position of the surface of the cylinder.}
    \label{fig:res:cp}
\end{figure}

This test case models a more realistic configuration, allowing for quantitative comparisons with results from the literature. It has been extensively studied both experimentally and \chg{numerically,~\cite{Gowen1952,Bashkin2000,Bashkin2002}} and we have obtained satisfactory results for the pressure distribution around the cylinder surface, as shown in \cref{fig:res:cp}.

\begin{figure*}[htpb]
    \subfloat[]{
        \includegraphics[width=0.4\textwidth]{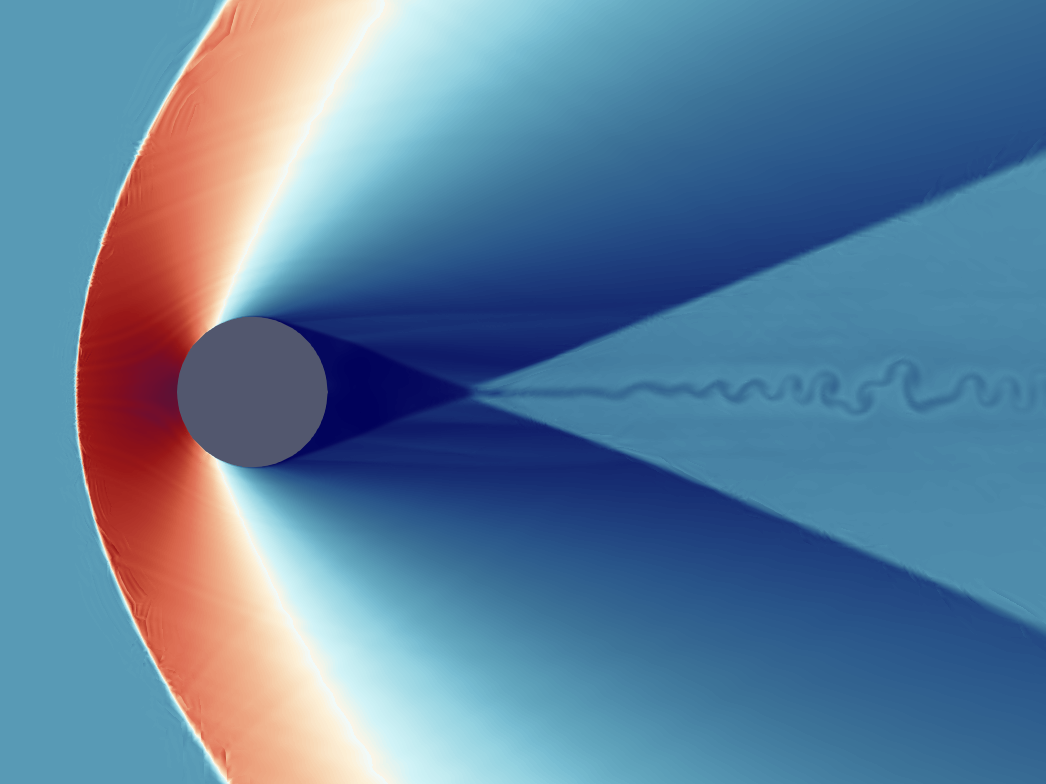}
    }\qquad
    \subfloat[]{
        \includegraphics[width=0.4\textwidth]{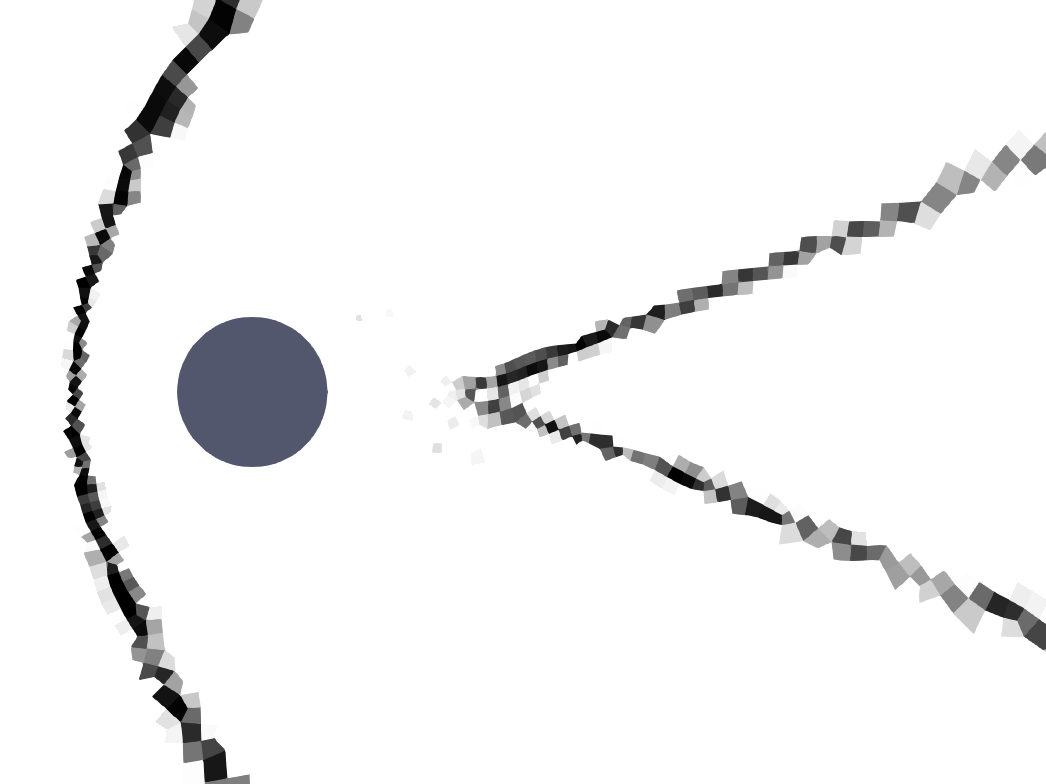}
    }\\
    \subfloat[]{
        \includegraphics[width=0.4\textwidth]{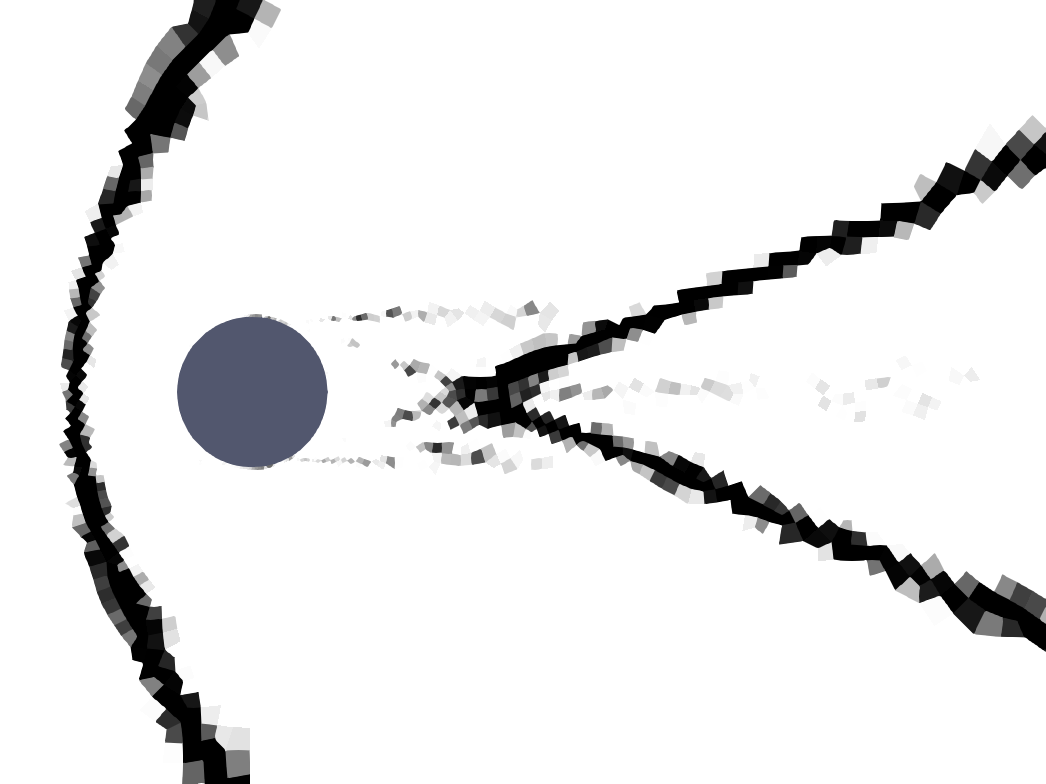}
    }\qquad
    \subfloat[]{
        \includegraphics[width=0.4\textwidth]{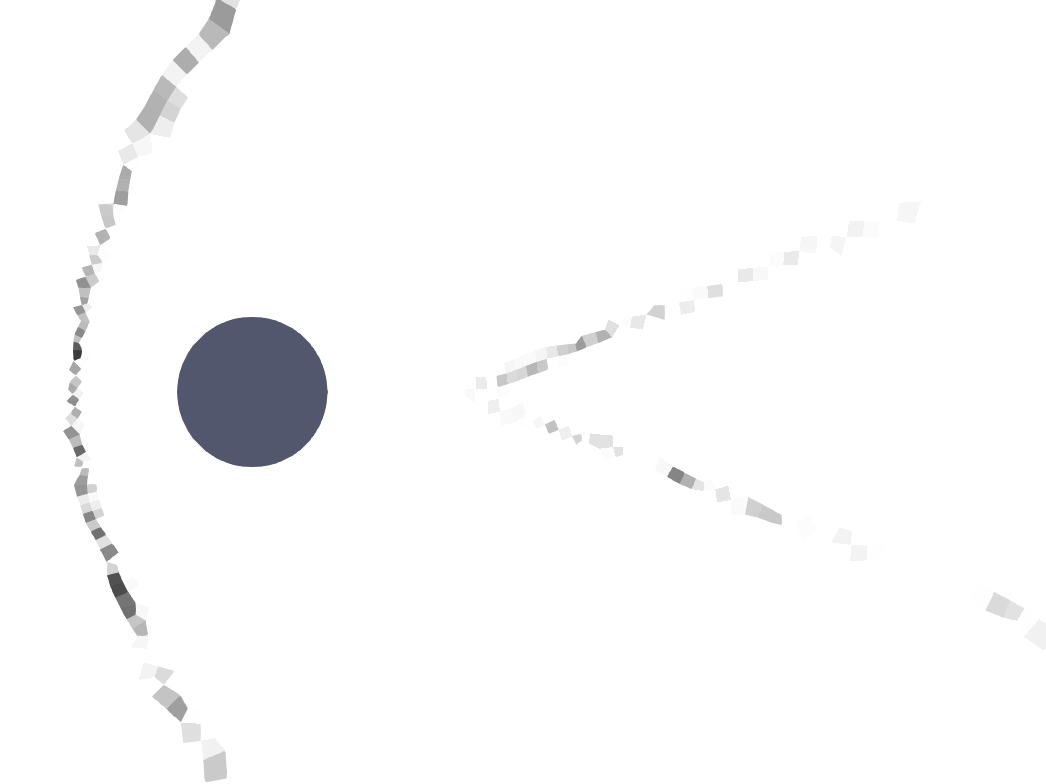}
    }
    \caption{Viscous case after 300,000 iterations with the modal sensor of \cref{sub:sens:modal}, using $p\rho$. a) density field, b) sensor with $s_0 = -2.5$ and $\Delta s = 1$. Sensor applied to the last iteration with $s_0 = -3.5$, $\Delta s = 1$ (c), and with $s_0 = -1.5$, $\Delta s = 1$ (d).}
    \label{fig:res:viscous_modal}
\end{figure*}

\begin{figure*}[htpb]
    \subfloat[]{
        \includegraphics[width=0.4\textwidth]{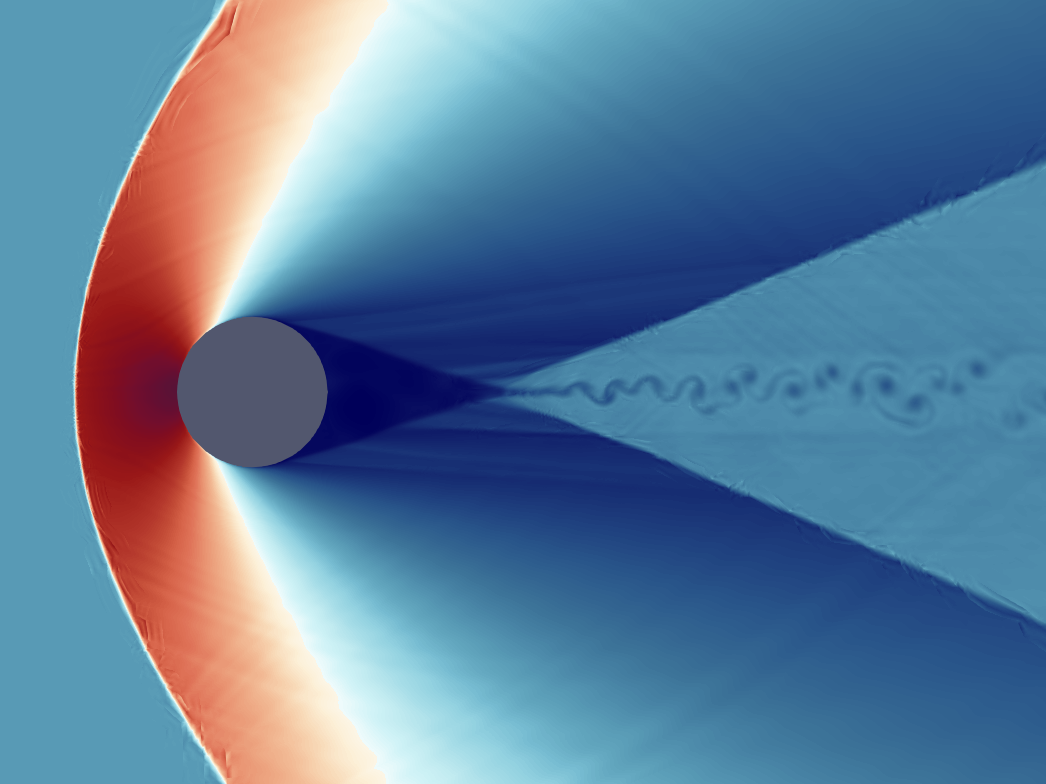}
    }\qquad
    \subfloat[]{
        \includegraphics[width=0.4\textwidth]{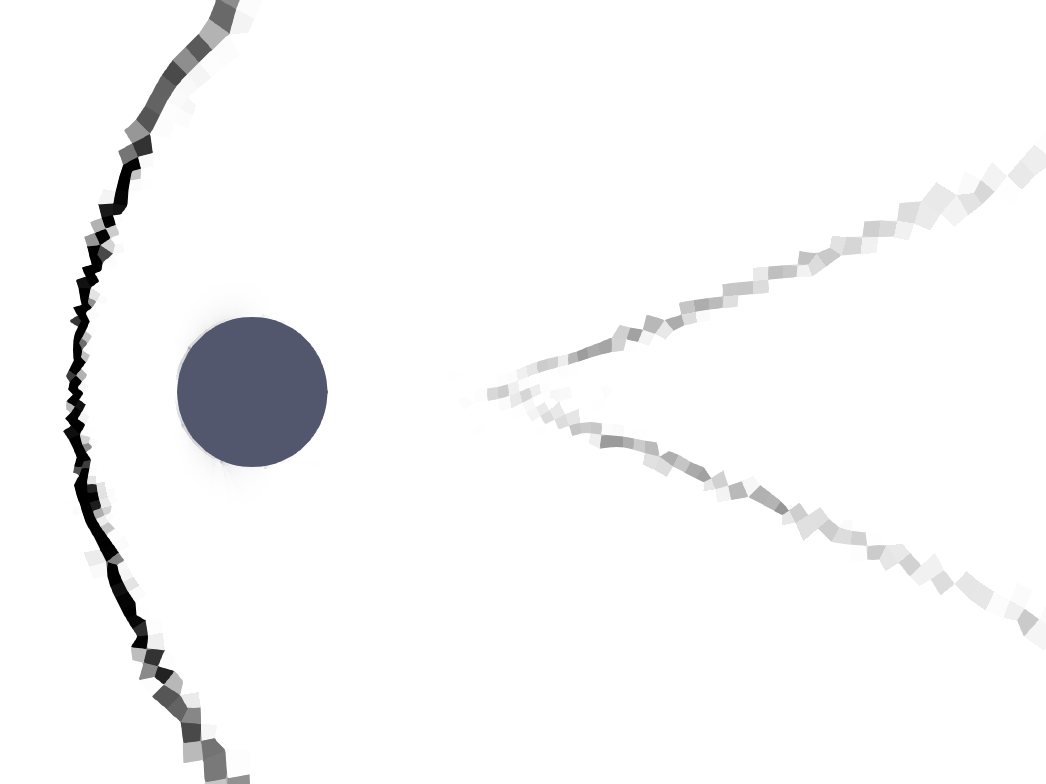}
    }\\
    \subfloat[]{
        \includegraphics[width=0.4\textwidth]{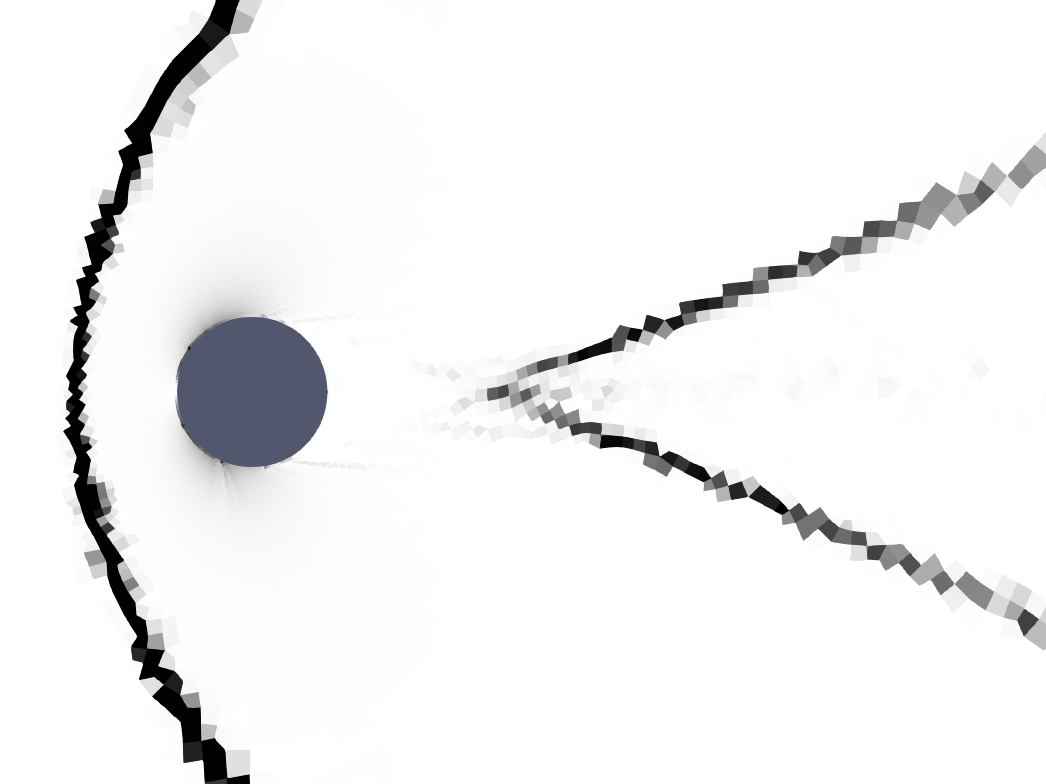}
    }\qquad
    \subfloat[]{
        \includegraphics[width=0.4\textwidth]{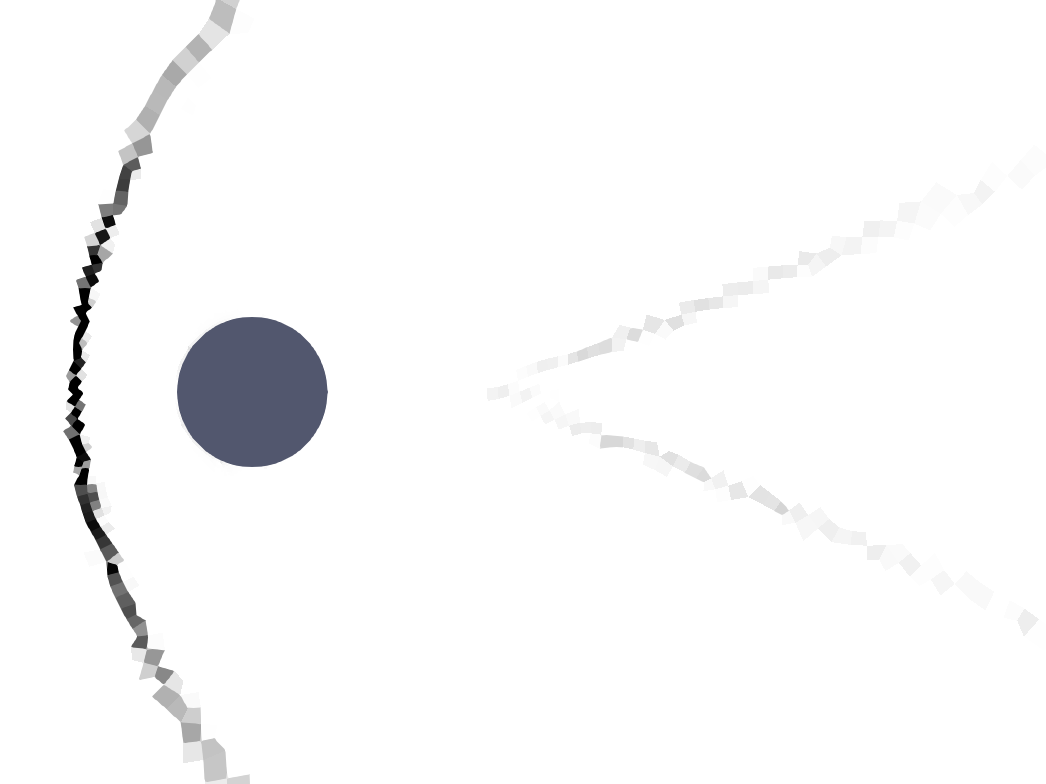}
    }
    \caption{Viscous case after 300,000 iterations with the integral sensor of \cref{sub:sens:integral}, using $\lVert\nabla p\rVert^2$. a) density field, b) sensor with $s_0 = 5.25$ and $\Delta s = 4.75$. Sensor applied to the last iteration with $s_0 = 2.5$, $\Delta s = 2.5$ (c), and with $s_0 = 8$, $\Delta s = 7$ (d).}
    \label{fig:res:viscous_integral}
\end{figure*}

\begin{figure*}[htpb]
    \subfloat[]{
        \includegraphics[width=0.4\textwidth]{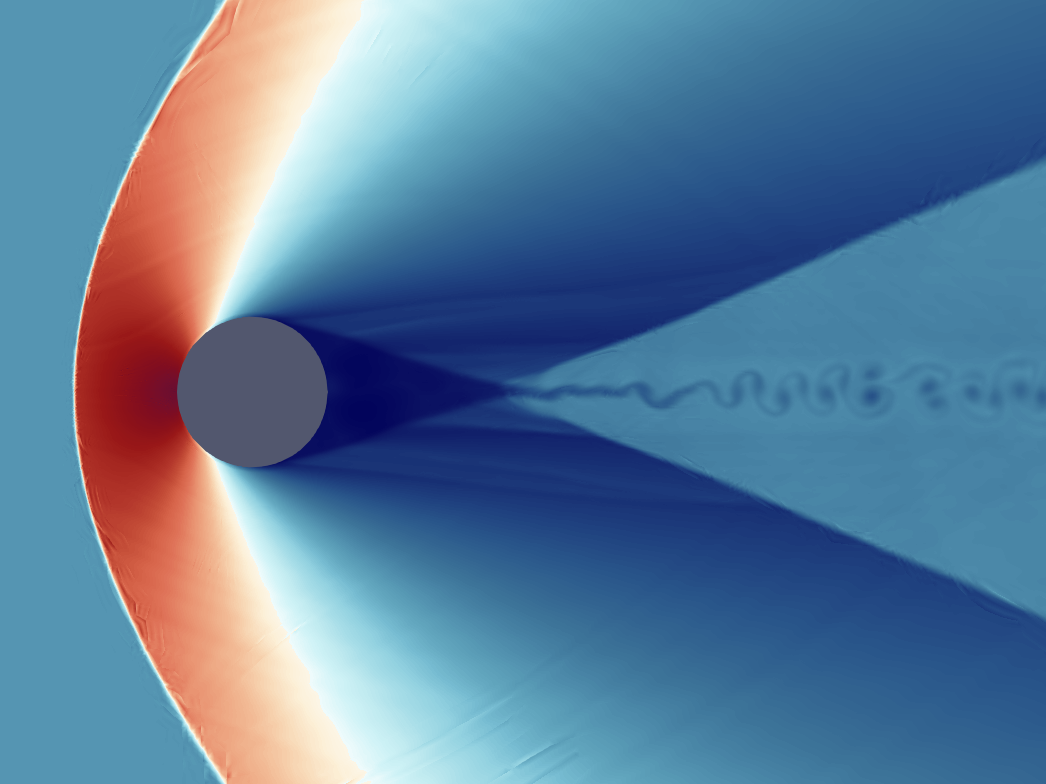}
    }\qquad
    \subfloat[]{
        \includegraphics[width=0.4\textwidth]{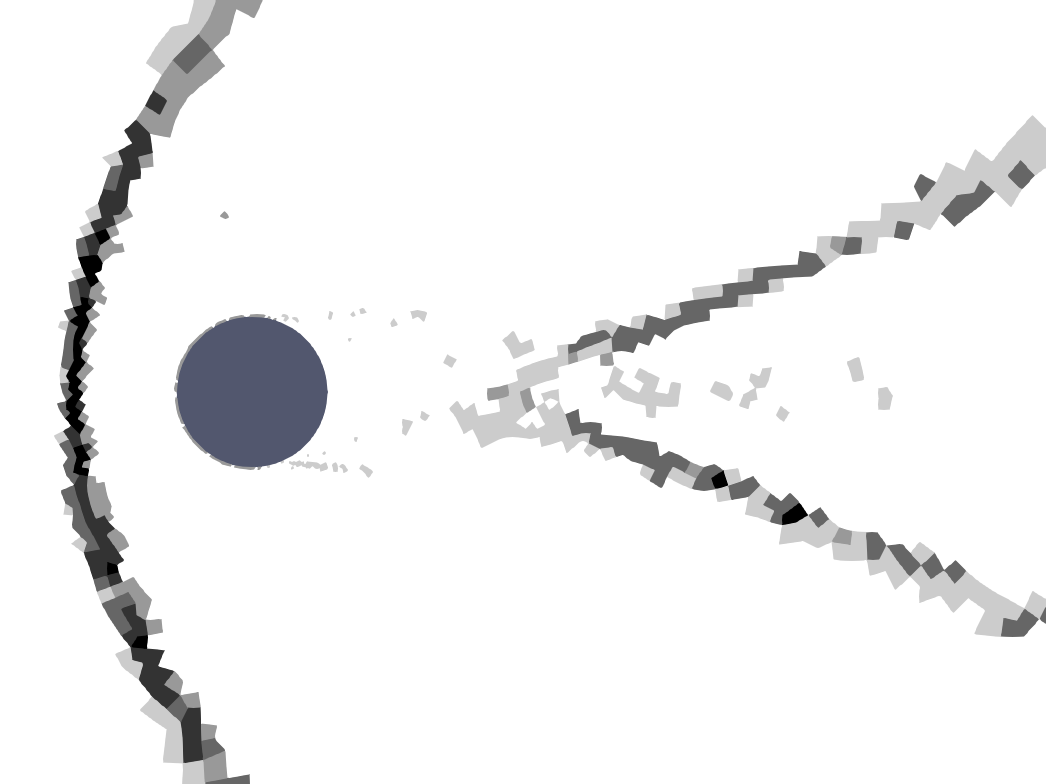}
    }
    \caption{Viscous case after 300,000 iterations with our adaptive GMM sensor of \cref{sec:gmm}, using $\lVert\nabla p\rVert^2, (\nabla\cdot \svec{v})^2$. a) density field, b) sensor with six clusters.}
    \label{fig:res:viscous_gmm}
\end{figure*}

The simulations were carried out using the sensors specified in \cref{tab:res:viscous_params}. While it is coincidental that the parameter values match those of \cref{tab:res:inviscid_params}, we invite the reader to refer to \cref{sec:aVar} for additional examples in which the values differ.
In any case, the $C_p$ distribution shown in \cref{fig:res:cp}---averaged over~$10^5$ time steps or 20 time units---closely matches the experimental \chg{data~\cite{Gowen1952}} with the three sensors. The detachment point, at an angle of approximately $\theta_s \simeq 110^{\circ}$, is also well \chg{captured.~\cite{Gowen1952,Bashkin2000,Bashkin2002}}
This angle is measured from the stagnation point and, since the case is symmetric, has the same sign on both sides of the cylinder.

\begin{table}[htpb]
    \centering
    \caption{Parameters of the sensors for the viscous cylinder.}
    \begin{ruledtabular}
    \begin{tabular}{lccc}
        Sensor & $s_0$ & $\Delta s$ & \# of clusters \\
        \hline
        Modal & -2.5 & 1 & - \\
        Integral & 5.25 & 4.75 & - \\
        GMM & - & - & 6 \\
    \end{tabular}
    \end{ruledtabular}
    \label{tab:res:viscous_params}
\end{table}

Before the detachment point, the pressure gradient across the boundary layer is smooth, allowing for an accurate approximation of the pressure distribution. However, in the wake region, a recirculation bubble forms, and the pressure field is not captured with the same accuracy.
It is also important to note that we do not expect a perfect match, as the mesh is not specifically optimized to capture all the scales involved in the boundary layer. The $y^{+}$ value of our discretization is around 20, indicating that the element size and distribution of high-order nodes are not fine enough to accurately resolve the viscous stresses close to the wall.
As a result, we were unable to accurately compare the total drag with the experimental data, but our pressure distribution results remain accurate and in good agreement with the experiments.

In \cref{fig:res:viscous_modal,fig:res:viscous_integral,fig:res:viscous_gmm} (sub-plots a and b), we present the density plots and sensor distributions for the three methods. All three approaches successfully capture the main shock wave with sub-cell resolution, and the oscillations in the wake region are not dissipated.
\chg{Upon} analysis of the sensor values, we observe that no additional viscosity is introduced in the smooth regions or in the wake downstream. This lack of additional viscosity explains the high level of detail observed in the results, such as the accurate representation of detachment shocks, small vortices in the wake, or thin shock waves. 

As expected, the sensors perform well in capturing the shock waves in this test case, given their successful performance in the inviscid scenario. However, the main challenge in this setup lies in not detecting the boundary layer around the cylinder.
\chg{Boundary} layers and shock waves are physically distinct phenomena, but they both impose significant computational demands and can induce oscillations and non-physical behaviors in high-order approximations. In our DGSEM, no-slip boundary conditions are imposed weakly, which means that the discontinuity between the flow near the boundary and the actual boundary condition can grow if the approximation is excessively coarse. This phenomenon is particularly evident in supersonic simulations with transient flow configurations, leading to high numerical fluxes that introduce oscillations in the boundary layer.
At this stage, even a small addition of artificial viscosity can lead to a simulation crash.

Taking into account this, it is evident that the modal and integral sensors in \cref{fig:res:viscous_modal,fig:res:viscous_integral} needed precise threshold selection to avoid incorrect detection patterns, as deviations from these values resulted in inaccuracies (see sub-figures c and d).
In contrast, our GMM sensor does not suffer from this limitation, as the clustering process is automatic, and the discrimination between smooth and non-smooth regions remains independent of the number of clusters used.
Numerical evidence supporting this is presented in \cref{sub:stats}.

\subsection{Analysis of the feature space and sensitivity to the number of clusters}
\label{sub:stats}

The preceding test cases have served as a qualitative assessment of our GMM-based sensor, particularly when compared to other established sensors in the literature.
In this section, our objective is to provide a clearer understanding of how the number of clusters affects the properties of the sensors.

A common metric used to assess the goodness of fit of a model to the given data is the Bayesian Information Criterion (BIC), defined as
\begin{equation*}
    \text{BIC} = -2\log L + N_p\log N,
\end{equation*}
where $\log L$ is the logarithm of the likelihood, $N_p$ is the number of free parameters of the model, and $N$ is the number of points in the feature space. For the \chg{GMM}, the number of free parameters is given by
\begin{equation*}
    N_p = K + Kv + K\frac{v(v+1)}{2} - 1,
\end{equation*}
using the definitions of \cref{eq:gmm:gm}, and $\log L$ takes the form of \cref{eq:gmm:log_likelihood}. Another well-known metric is the Akaike Information Criterion (AIC), defined as
\begin{equation*}
    \text{AIC} = -2\log L + 2N_p.
\end{equation*}
The difference between these two criteria is $N_p(2 - \log N)$, and in our case, it is several orders of magnitude smaller than the log-likelihood. As both methods would yield the same conclusions, we continue with the BIC in the following discussion.\\

We begin by extracting the solution of the viscous flow case at $t = 60$ obtained using the GMM sensor, as depicted in \cref{fig:res:viscous_gmm}.
Using a Python code and the scikit-learn \chg{library,~\cite{sklearn}} we compute the Bayesian Information Criterion (BIC) for a varying number of clusters, ranging from one to six.

\begin{figure}[htpb]
    \centering
    \includegraphics[width=0.45\textwidth]{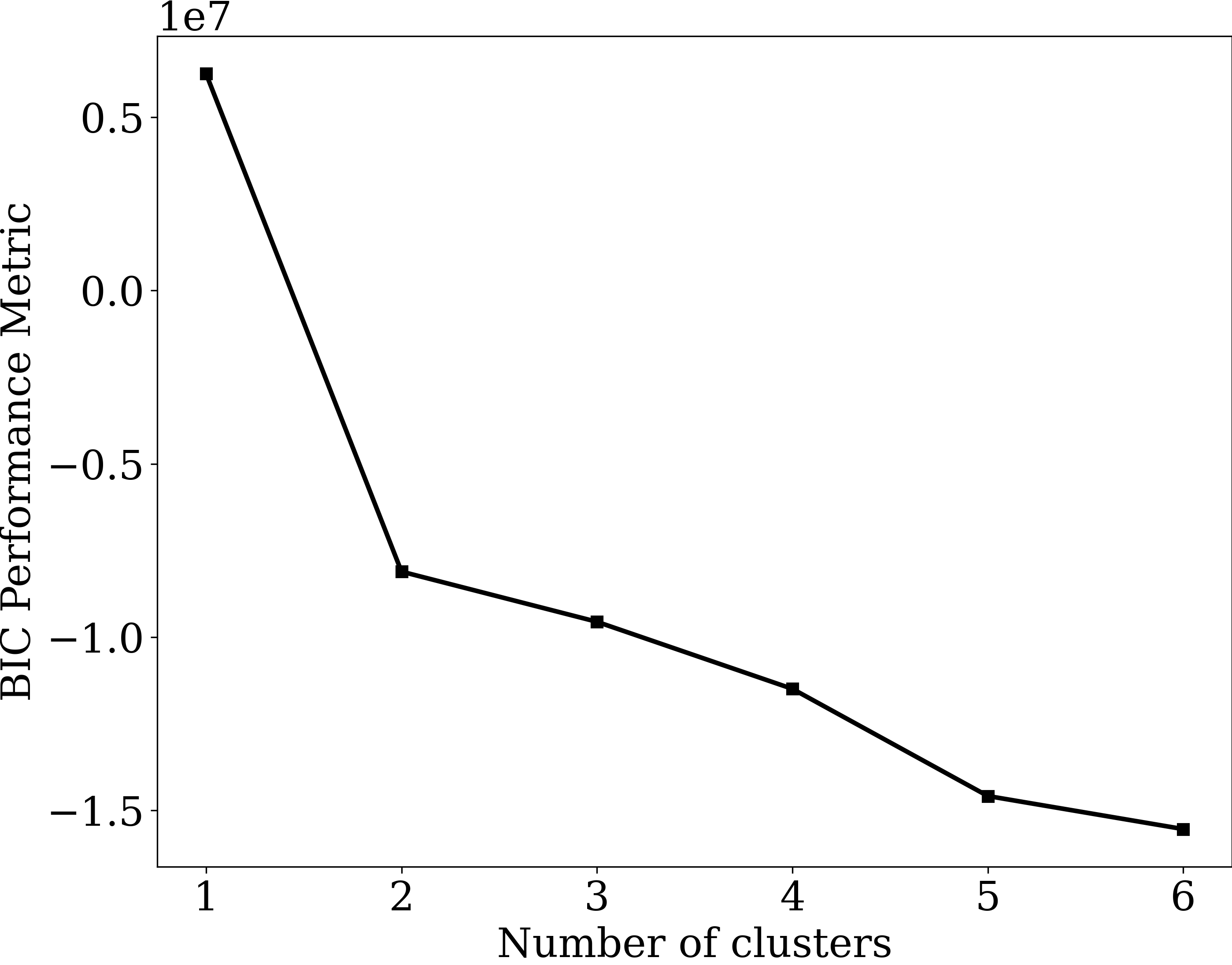}
    \caption{Bayesian information criterion as a function of the number of clusters.}
    \label{fig:stat:bic}
\end{figure}

\begin{table}[htpb]\chgenv
    \centering
    \caption{AIC and BIC values shown in \cref{fig:stat:bic}.}
    \begin{ruledtabular}
    \begin{tabular}{ccc}
         \# of clusters & AIC & BIC \\
         \hline
         1              & 6,250,414   & 6,254,626   \\
         2              & -8,119,936  & -8,111,501  \\
         3              & -9,571,977  & -9,559,320  \\
         4              & -11,511,725 & -11,494,846 \\
         5              & -14,607,819 & -14,586,717 \\
         6              & -15,566,407 & -15,541,083 \\
    \end{tabular}
    \end{ruledtabular}
    \label{tab:stat:aic_bic}
\end{table}

The plot in \cref{fig:stat:bic} clearly demonstrates that the use of two clusters is sufficient to capture most of the information, since the rate of descent in the BIC decreases significantly after this point. This rule-of-thumb is often referred to as the ``elbow method'', and takes into account that the BIC balances lower values of the log-likelihood with higher numbers of parameters to avoid favoring overfitted models. In this specific application of the GMM for a shock-capturing sensor, we have found that increasing the number of clusters beyond two does not negatively impact the performance of the artificial viscosity approach used in this work.

\begin{figure}[htpb]
    \centering
    \includegraphics[width=0.48\textwidth]{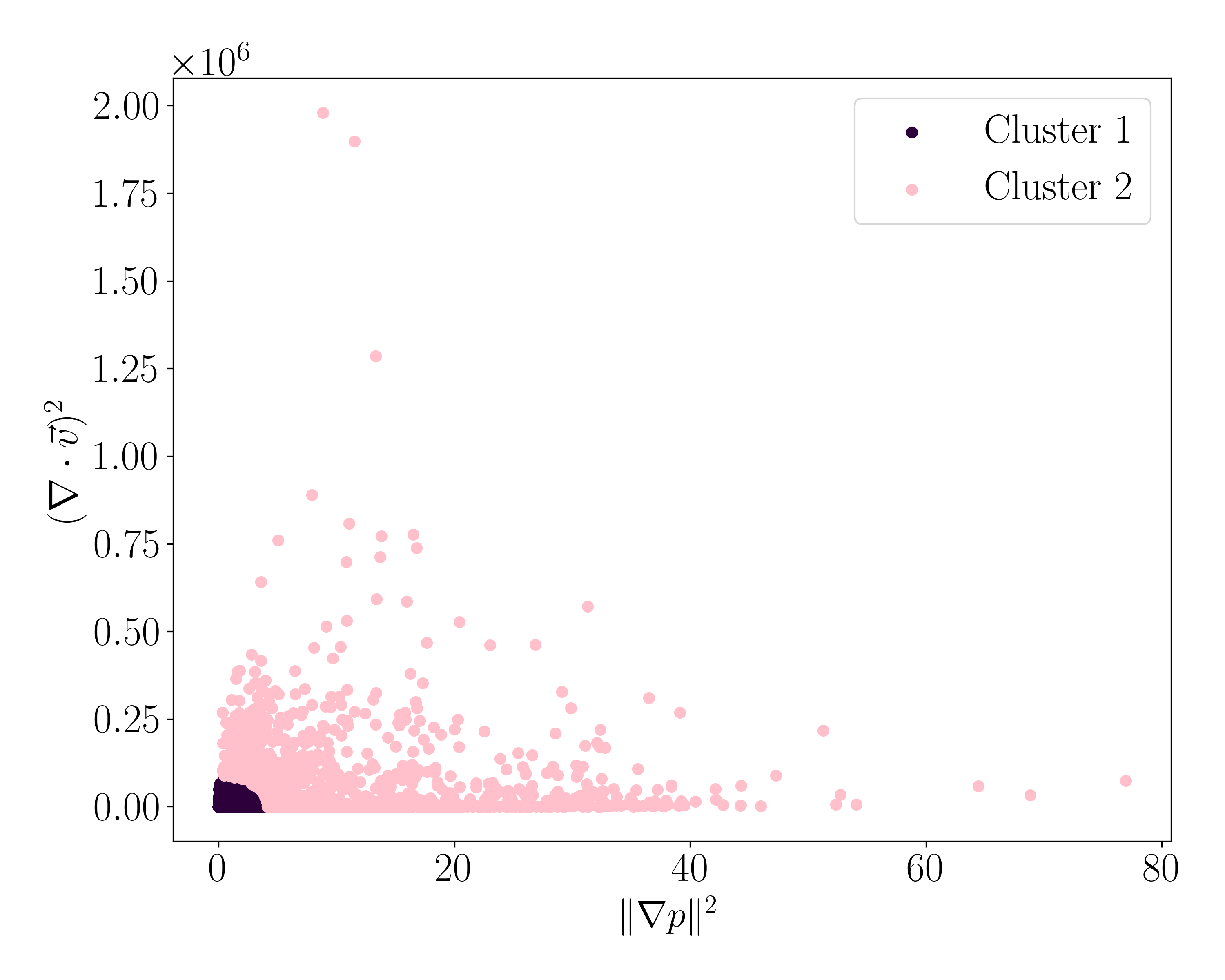}
    \caption{GMM sensor with two clusters applied to the viscous case.}
    \label{fig:stat:feature_two}
\end{figure}
\begin{figure}[htpb]
    \centering
    \includegraphics[width=0.48\textwidth]{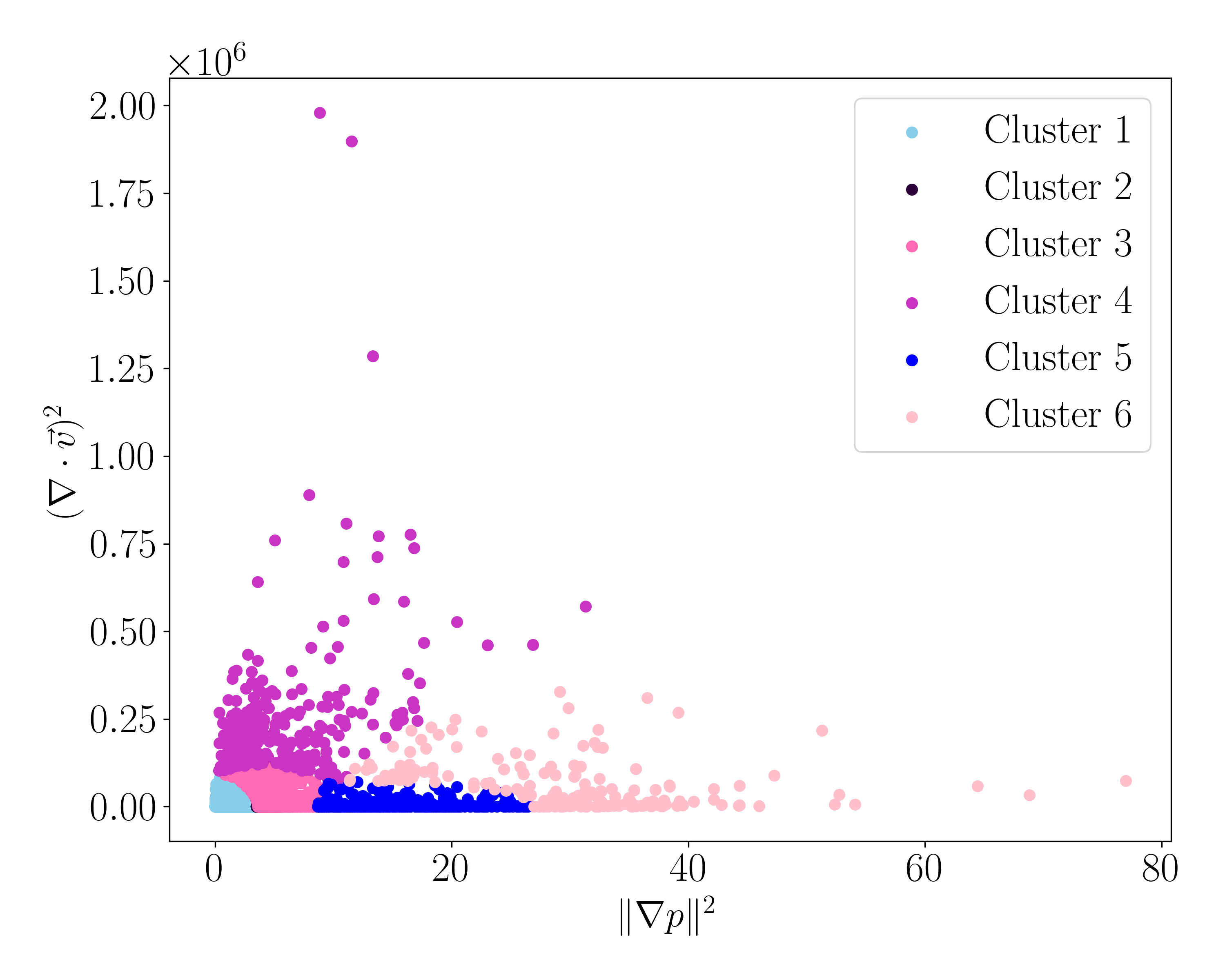}
    \caption{GMM sensor with six clusters applied to the viscous case.}
    \label{fig:stat:feature_six}
\end{figure}
The Gaussian mixture models with one to six clusters effectively differentiate the two main regions: smooth flow and discontinuities. As shown in \cref{fig:stat:feature_two}, cluster 1 encompasses the vast majority of points representing smooth regions, while cluster 2 includes the fewer nodes associated with shock waves.
Both the BIC and the feature-space plot support the idea that this division of the feature space is the most representative of the dataset. Therefore, when applying the GMM shock sensor to a specific numerical discretization, using two clusters as a default value is likely to yield reliable and accurate results.
However, as we have explained, the artificial viscosity method we have used in this work benefits from smooth spatial transitions in the viscosity field. Therefore, using only two clusters might not be optimal.

In \cref{fig:stat:feature_six}, we present the feature space and the groups made by the GMM when using six clusters. It is evident that the first cluster has an almost identical shape, indicating that the differences between this cluster and the others are significant enough to ensure a clear distinction between smooth and non-smooth regions. Consequently, we can add more levels to the sensor by increasing the number of clusters.
In our simulations, we could not exceed ten clusters because the adaption step we introduced caused some clusters to collapse. The ideal number of clusters might depend on the specific case being simulated, particularly on the ratio of nodes in discontinuous regions compared to nodes in smooth regions. However, this issue does not significantly impact our discussion, as our primary goal is to describe a sensor algorithm capable of detecting shock waves with a certain level of refinement. Based on our experience, using four to six clusters provides the best results.



\subsection{Performance scaling}
\label{sub:perf}

In this section, we shift our focus to investigate the computational cost of our sensor. While it offers improved robustness in the numerical discretization, there is a trade-off in terms of increased computational requirements. To assess the performance of the GMM sensor, we performed a strong scaling test, comparing its efficiency with different parallelization strategies available in HORSES3D\chg{---including} both threads (OpenMP) and processes (MPI)\chg{---}. In our testing, we evaluated the performance of the sensor under different combinations of parallelization methods.
\begin{figure}[htpb]
    \centering
    \includegraphics[width=0.4\textwidth]{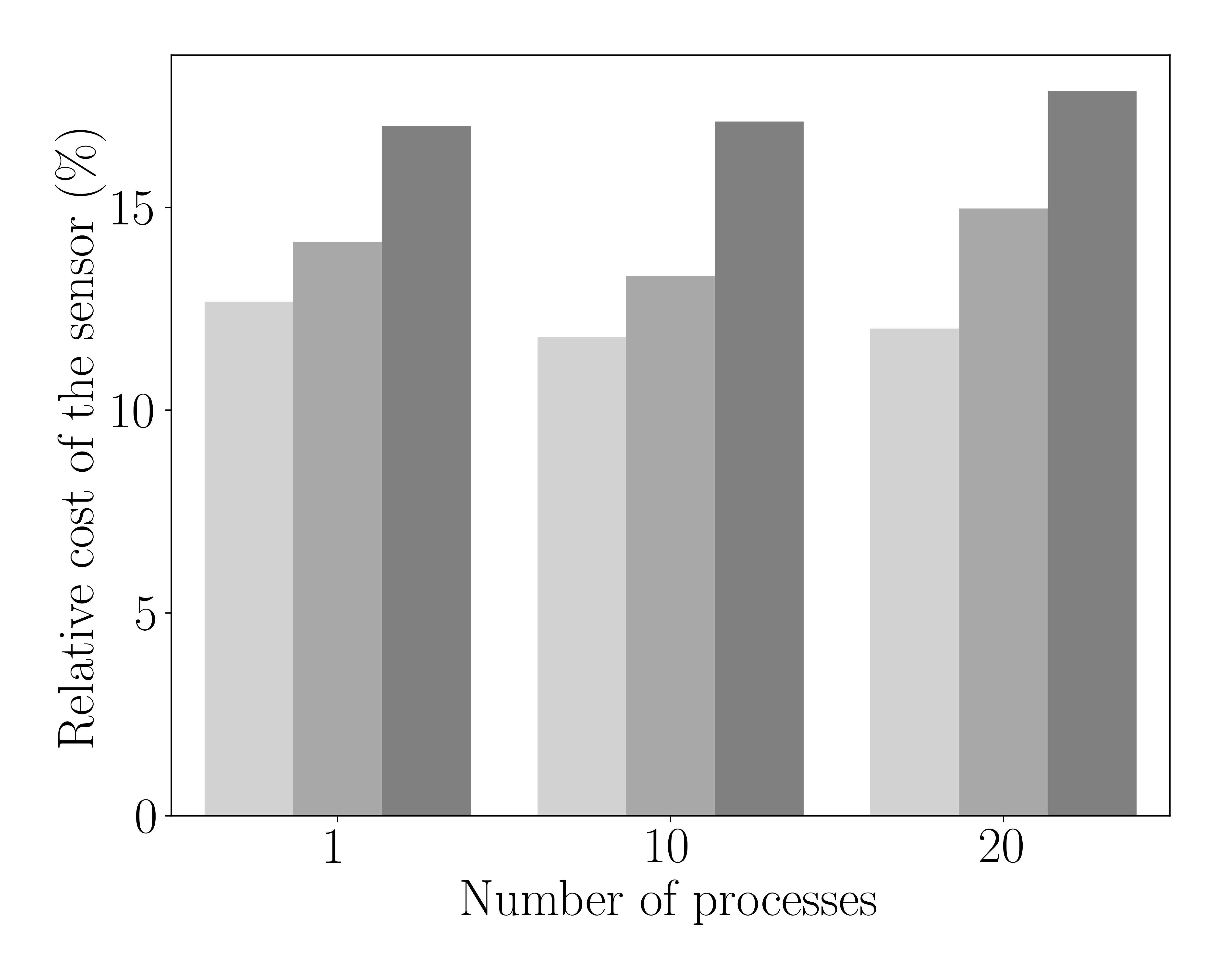}
    \caption{Cost of evaluating the sensor relative to the total time of one time step in the case of the viscous cylinder. The measurements are performed over 20 time steps. \chg{Light gray}: 1 thread, \chg{gray}: 2 threads, \chg{dark gray}: 4 threads.}
    \label{fig:perf:cost}
\end{figure}
The results are presented in \cref{fig:perf:cost}, which illustrates the relative computational cost of the GMM-based sensor within a single time step. To generate these data, we used the solution obtained from the GMM-based sensor in \cref{sub:viscous} (see \cref{fig:res:viscous_gmm}) and computed the metric for 20 iterations. The sensor computation time was then divided by the total time taken for the same 20 time steps.

Interestingly, we observed that the cost of the sensor does not vary with the number of processes used. This observation aligns with our explanation in \cref{sec:gmm}, where we discussed that data transfer between processes is not a significant issue with the GMM sensor. Since only small amounts of information need to be exchanged, the performance remains consistent regardless of the number of processes employed.

However, we did observe that the multithreaded performance of our implementation could be further optimized. The other operations performed by the software during a single time step scale better when increasing the number of threads. Consequently, there is a performance hit of around 10\% to 20\% when computing the sensor every time step.
This effectively results in an overhead of 2\% maximum in the total simulation time if the sensor is updated every 10 iterations.
We justify this by recalling that we use an explicit time integration scheme (as described in \cref{sub:tdisc}) and this typically involves small time-step sizes.
Therefore, the solution undergoes minimal changes after a few time steps, and the sensor output remains virtually constant.
\begin{table*}[htpb]\chgenv
    \centering
    \caption{Relative cost (\%) of the sensors with respect to the computation of one time step in the case of the viscous cylinder.}
    \begin{ruledtabular}
    \begin{tabular}{lccc|ccc|ccc}
         & \multicolumn{3}{c|}{1 process} & \multicolumn{3}{c|}{10 processes} & \multicolumn{3}{c}{20 processes} \\
        \cline{2-10}
         & 1 thread & 2 threads & 4 threads & 1 thread & 2 threads & 4 threads & 1 thread & 2 threads & 4 threads \\
        \hline
        GMM & 12.68 & 14.16 & 17.02 & 11.80 & 13.31 & 17.12 & 12.02 & 14.97 & 17.86 \\
        Modal & 0.71 & 0.65 & 0.58 & 0.71 & 0.54 & 0.39 & 0.40 & 0.40 & 0.52 \\
        Integral & 0.61 & 0.56 & 0.53 & 0.51 & 0.69 & 0.61 & 0.41 & 0.59 & 0.34 \\
    \end{tabular}
    \end{ruledtabular}
    \label{tab:perf:cost}
\end{table*}
The cost of the other traditional sensors considered in this work is at least an order of magnitude lower \chg{(see \cref{tab:perf:cost})}, but the impact is nevertheless low in all the cases, and the advantages of our proposed sensor offset this overhead.

Lastly, it is essential to emphasize that \cref{fig:perf:cost} does not provide insight into the scalability of HORSES3D. Instead, it solely compares the performance of the new GMM-based sensor against the rest of the code. The constant cost observed with varying processes merely indicates that the GMM-based sensor scales well within the solver.

\section{Conclusions}
\label{sec:conclusions}

In this article, we have introduced a shock detection algorithm based on \chg{GMMs} and integrated it into an existing high-order \chg{CFD} solver. Unlike many common sensors, our method requires only one parameter, the maximum number of clusters, and provides default values that yield satisfactory results in most cases. The adaptive nature of the algorithm allows it to modify this parameter if the user-provided value is not suitable for the simulation, making it easy to use and apply.

The Bayesian formulation of our sensor allows it to leverage previous information, resulting in improved accuracy and performance. This feature makes it suitable for integration into the workflow of PDE solvers. We have successfully incorporated it into our open-source solver HORSES3D~\cite{Ferrer2023} as part of \chg{two common stabilization strategies.}
The algorithm has demonstrated its effectiveness in challenging cases, including \chg{Mach 10 flows and} moderately high Reynolds numbers. Although we have presented a simple approach in this article, it has proven to be effective in enhancing the robustness of existing software.

We recognize that there is potential for further improvement and refinement of our approach. For example, \chg{Bayesian} inference techniques could be utilized to implement better prediction models for the number of clusters in the GMM. Exploring different feature spaces might also yield more suitable choices for shock detection. Although we have conducted various tests with different variables, a more comprehensive study of the influence of the feature space dimension and variable selection is left for future work.

Our research findings indicate that unsupervised machine learning methods hold promise in improving the performance of complex CFD codes. Such codes often encounter varied and complex geometries and flow configurations, which can be challenging to handle using traditional supervised algorithms that rely on training datasets. On the contrary, unsupervised methods, such as GMM-based shock detection, can adapt to diverse scenarios and offer robust solutions without the need for extensive training data. As the field of machine learning advances, we envision even greater opportunities to enhance CFD simulations and address real-world cases.

\begin{acknowledgments}
Andr\'es Mateo has received funding from Universidad Polit\'ecnica de Madrid under the Programa Propio PhD programme.
Kenza Tlales was supported by Grant 080 Bis/PG/Espagne/2020 2021 of Ministère de l'Enseignement Supérieur et de la Recherche Scientifique, République Algérienne Démocratique et Populaires. 
Gonzalo Rubio, Esteban Ferrer and Eusebio Valero acknowledge the funding received by the Grant NextSim / AEI /10.13039/501100011033 and H2020, GA-956104.  
Gonzalo Rubio, Esteban Ferrer and Eusebio Valero acknowledge the funding received by the Grant DeepCFD (Project No. PID2022-137899OB-I00) funded by MCIN/AEI/10.13039/501100011033/ and by ERDF A way of making Europe. 
Esteban Ferrer would like to thank the support of Agencia Estatal de Investigación (for the grant "Europa Excelencia 2022" Proyecto EUR2022-134041/AEI/10.13039/501100011033) y del Mecanismo de Recuperación y Resiliencia de la Unión Europea, and the Comunidad de Madrid and Universidad Politécnica de Madrid for the Young Investigators award: APOYO-JOVENES-21-53NYUB-19-RRX1A0.
Finally, all authors gratefully acknowledge Universidad Politécnica de Madrid (\url{www.upm.es}) for providing computing resources on Magerit Supercomputer.

\end{acknowledgments}

\section*{Author declarations}

\subsection*{Conflict of interest}

The authors have no conflicts to disclose.

\subsection*{Author contributions}

\textbf{Andr\'es Mateo-Gab\'in:} Conceptualization; Data curation; Methodology; Software; Visualization; Writing -- original draft. \textbf{Kenza Tlales:} Data curation; Formal analysis; Software; Visualization. \textbf{Eusebio Valero:} Conceptualization; Funding acquisition; Project administration. \textbf{Esteban Ferrer:} Conceptualization; Funding acquisition; Methodology; Project administration. \textbf{Gonzalo Rubio:} Conceptualization; Funding acquisition; Methodology; Supervision; Writing -- review \& editing.

\section*{Data availability statement}

The data that support the findings of this study are openly available in \url{https://github.com/Andres-MG/2023_gmm_shock_sensor}.

\appendix

\section{K-means clustering}
\label{sec:aKM}

The k-means algorithm defines clusters based on their centroids, and nodes are assigned to the closest ones. It produces hyperspherical clusters, making it suitable for feature spaces where the different groups are well defined and quasi-isotropic.
The number of clusters is fixed at the beginning of the algorithm, and the final distribution is found iteratively, usually converging after a few steps (see \cref{alg:aKM:kmeans}).
As an iterative method, it is necessary to have an initial state to begin the iterations.
Therefore, if no information is available, we start the algorithm with a random distribution of clusters.
However, if there is known data, we restart the algorithm from that starting point.
In our work, this is the case when we use it to enhance the results of an unconverged GMM pass.

\begin{algorithm}
    \caption{K-means.}
    \label{alg:aKM:kmeans}
    \DontPrintSemicolon
    \KwIn{$nclusters$, \chg{$max\_iters$}, $x$}
    \KwOut{$\bar{x}$, $clusters$}
    \BlankLine

    Initialize $\bar{x}$\;
    \BlankLine

    $clusters \gets GetClusters(x, \bar{x})$\;
    $prevclusters \gets clusters$\;
    \BlankLine

    \For{$i$ in 1, \dots, \chg{max\_iters}}{
        $\bar{x} \gets GetCentroids(x, clusters)$\;
        $clusters \gets GetClusters(x, \bar{x})$\;
        \BlankLine

        \eIf{$clusters = prevclusters$}{
            Leave the loop. The algorithm has converged\;
        }{
            $prevclusters \gets clusters$\;
        }
    }
\end{algorithm}

The implementation is similar to the one described for the GMM in \cref{sec:gmm}, and two steps are performed at each iteration.
First, the centroids of the clusters are computed as the average position of all the points assigned to each group in the last iteration.
Then, this new cluster distribution is used to recompute the distances between nodes and centroids, regrouping the points with the new definitions.
The algorithm stops when the centroids are not updated in two consecutive iterations.

\section{Details on the numerical formulation}
\label{sec:aDisc}

\subsection{\chg{Navier--Stokes} equations}
\label{sub:ans}

\Cref{eq:ns:advdiff} states the generic form of an advection-diffusion equation, but does not introduce the specific form of the various terms.
The advective,~$\bvec{f}_e$, and viscous,~$\bvec{f}_v$, fluxes of the \chg{Navier--Stokes} equations, are defined as
\begin{equation*}
\begin{gathered}
    \stvec{f}_e = \left(\begin{array}{c}
        \rho u \\ \rho u^2 + p \\ \rho uv \\ \rho uw \\ \rho hu
    \end{array}\right), \quad
    \stvec{f}_v = \left(\begin{array}{c}
        0 \\ \tau_{11} \\ \tau_{21} \\ \tau_{31} \\ \svec{\tau}_1 \cdot \svec{v} + q_1
    \end{array}\right), \\
    \stvec{g}_e = \left(\begin{array}{c}
        \rho v \\ \rho uv \\ \rho v^2 + p \\ \rho vw \\ \rho hv
    \end{array}\right), \quad
    \stvec{g}_v = \left(\begin{array}{c}
        0 \\ \tau_{12} \\ \tau_{22} \\ \tau_{32} \\ \svec{\tau}_2 \cdot \svec{v} + q_2
    \end{array}\right), \\
    \stvec{h}_e = \left(\begin{array}{c}
        \rho w \\ \rho uw \\ \rho vw \\ \rho v^2 + p \\ \rho hw
    \end{array}\right), \quad
    \stvec{h}_v = \left(\begin{array}{c}
        0 \\ \tau_{13} \\ \tau_{23} \\ \tau_{33} \\ \svec{\tau}_3 \cdot \svec{v} + q_3
    \end{array}\right).
\end{gathered}
\end{equation*}
These variables are related by the expressions
\begin{equation*}
    \rho e = \rho e_i + \frac{1}{2}\rho\lvert\svec{v}\rvert^2, \quad
    e_i = \frac{p/\rho}{\gamma - 1}, \quad
    p = \rho RT,
\end{equation*}
and the viscous flux is defined in terms of the stress tensor and the heat flux,
\begin{equation*}
\begin{gathered}
    \tau_{ij} = \mu \left(\frac{\partial v_i}{\partial x_j} + \frac{\partial v_j}{\partial x_i} - \frac{2}{3}\nabla \cdot \svec{v}\ \delta_{ij}\right), \quad
    \svec{\tau}_i = (\tau_{1i}, \tau_{2i}, \tau_{3i}), \\
    \svec{q} = \kappa \nabla T, \quad
    \kappa = \theta \mu R, \quad \theta = \frac{\gamma}{(\gamma - 1)\text{Pr}}.
\end{gathered}
\end{equation*}
The Prandtl number, Pr, is characteristic of the medium, and we use a value of 0.72 for the air.

These definitions include conservation laws for the mass, momentum and energy, but not the entropy.
However, it is also an important physical principle that the entropy of a closed system cannot decrease.
To ensure that our numerical approximations also accommodate this law, we introduce the concept of mathematical entropy,~$S$.
This entropy is defined as a scaled version of the physical entropy,~$s$,
\begin{equation*}
    S = -\frac{\rho s}{\gamma - 1}, \quad s = \ln p - \gamma \ln \rho.
\end{equation*}
The derivative of this entropy function with respect to the conservative variables,~$\stvec{q}$, is a new set of variables called \emph{entropy variables},~$\stvec{w}$,
\begin{equation*}
    \stvec{w} = \nabla_{\stvec{q}}S = \left(\frac{\gamma-s}{\gamma - 1}-\frac{\rho \lvert\svec{v}\rvert^2}{2p}, \frac{\rho u}{p}, \frac{\rho v}{p}, \frac{\rho w}{p}, -\frac{\rho}{p}\right)^T.
\end{equation*}
We use these entropy variables to compute the artificial viscosity flux of \cref{eq:ns:gpflux}, ensuring that the discretization is entropy-stable.~\cite{Guermond2014,mateogabin2022entropy}

\subsection{Spatial discretization}

In this work, we follow a discontinuous Galerkin (DG) approach to discretize the spatial part of \cref{eq:ns:advdiff}.
First, the physical domain is tessellated into non-overlapping elements, and each of them is mapped to the so-called standard element,~$E$, defined as~$\svec{\xi} \in [-1, 1]^3$.
For each element,~$e$, the mapping~$\svec{x} = \svec{X}\left(\svec{\xi}\right)$ relates the local coordinates in the reference element, $\svec{\xi} = \left(\xi, \eta, \zeta\right)^T \in E$, to the physical space, $\svec{x} = \left(x,y,z\right)^T \in e$, and allows the definition of \cref{eq:ns:advdiff} in terms of operators defined in the reference element.
From this mapping we can compute the local reference frame (covariant,~$\svec{a}_i$, and contravariant,~$\svec{a}^i$) in each element and the Jacobian, that gives an idea of the deformation of the element with respect to the standard space,
\begin{equation*}
    \svec{a}_i = \frac{\partial \svec{X}}{\partial \xi_i}, \quad
    \jac\svec{a}^i = \svec{a}_j \times \svec{a}_k, \quad
    \jac = \jac\svec{a}^i \cdot \svec{a}_i.
\end{equation*}

Defining the matrices~$\smat{M}$ with columns~$\svec{M}_i = \jac\svec{a}^i$, and~$\bmat{M}=\smat{M}\otimes\stmat{I}_5$ ($\stmat{I}_5$ is the~$5\times 5$ identity matrix), state and block vectors can be easily projected from the covariant to the contravariant basis.
In the case of gradients,
\begin{equation*}
\begin{gathered}
    \jac\nabla f = \smat{M}\nablaxi f, \quad
    \jac\nabla \cdot \svec{f} = \nablaxi \cdot \left(\smat{M}^T\svec{f}\right) = \nablaxi \cdot \csvec{f},\\
    \jac\nabla \stvec{f} = \bmat{M}\nablaxi \stvec{f}, \quad
    \jac\nabla \cdot \bvec{f} = \nablaxi \cdot \left(\bmat{M}^T\bvec{f}\right) = \nablaxi \cdot \cbvec{f},
\end{gathered}
\end{equation*}
where~$\csvec{f}$ and~$\cbvec{f}$ are the projections into the contravariant basis of a state and a block vector, respectively.

The weak form of \cref{eq:ns:advdiff} is obtained in several steps.
First, to avoid computing second derivatives, the gradients of the entropy variables,~$\stvec{w}$, are computed in a separate equation,~$\bvec{g}=\nabla\stvec{w}$.
Then, the combination of \cref{eq:ns:advdiff} with this definition of the gradients is multiplied by two test functions,~$\stvecg{\phi}$ and~$\bvecg{\psi}$, and integrated over the reference element.
After application of the chain rule, we obtain the weak form that needs to be solved for each of the elements in the entire domain,~$\Omega$,
\begin{widetext}
\begin{equation}
\label{eq:ns:advdiff_ldg_weak}
\begin{gathered}
    \left\langle \jac\stvec{q}_t,\stvecg{\phi}\right\rangle_E + \int_{\partial E}\stvecg{\phi}^T\cbvec{f}_e\cdot\hat{n}\diff\hat{S} - \left\langle\cbvec{f}_e,\nablaxi\stvecg{\phi}\right\rangle_E = \int_{\partial E}\stvecg{\phi}^T\left(\cbvec{f}_v + \cbvec{f}_a\right)\cdot\hat{n}\diff\hat{S} - \left\langle\cbvec{f}_v + \cbvec{f}_a,\nablaxi\stvecg{\phi}\right\rangle_E, \\
    \left\langle \jac\bvec{g},\bvecg{\psi}\right\rangle_E = \int_{\partial E}\stvec{w}^T\cbvecg{\psi}\cdot\hat{n}\diff\hat{S} - \left\langle\stvec{w},\nablaxi\cdot\cbvecg{\psi}\right\rangle_E.
\end{gathered}
\end{equation}
\end{widetext}
In \cref{eq:ns:advdiff_ldg_weak},~$\hat{n}$ and~$\mathrm{d}\hat{S}$ are the unit normal vector and
the surface differential of the six planar faces of~$E$ (e.g. for the
faces~$\xi=\pm 1$,~$\hat{n}=(\pm 1,0,0)^T$ and~$\diff\hat{S}=\diff\eta\diff\zeta$).
Since in a DG scheme there are no continuity constraints between the elements, these are used to exchange information between contiguous elements by solving the Riemann problem that they define.
Therefore, we introduce so-called \emph{numerical fluxes},~$\stvec{f}^{\star}(\stvec{q}_l, \stvec{q}_r, \hat{n})$, in the interfaces of the elements to approximate these one-dimensional Riemann problems.

The next step is introducing polynomial approximations of order~$P$.
Since we employ a \emph{Discontinuous Galerkin Spectral Element Method} \chg{(DGSEM)~\cite{Black1999,kopriva2009implementing}}, the polynomial basis comprises the Lagrange polynomials,~$\{l_i(x)\}_{i=0}^P$, associated to a set of nodes defined in the reference element.
For example, the approximation of~$\stvec{q}$ in a three-dimensional case for an element of the tessellation is
\begin{equation*}
\begin{gathered}
    \stvec{q} \approx \stvec{Q} = \sum_{i,j,k=0}^P \stvec{Q}_{ijk} l_i(\xi)l_j(\eta)l_k(\zeta), \\
    \stvec{Q}_{ijk} = \stvec{q}\left(\xi_i, \eta_j, \zeta_k\right),
\end{gathered}
\end{equation*}
and in the following we use uppercase letters for discretized variables.
We also approximate the integrals with \chg{Gauss--Lobatto} quadratures, using the nodes of the quadrature as interpolation points.
In two and three dimensions, the interpolation nodes are defined as a tensor-product extension of the one-dimensional case.

To further increase the robustness of the discretization, we consider split-form derivative operators to reduce the adverse effects of aliasing.
In the simple case of~$f = ab$, its split-form derivative would be
\begin{equation}
\label{eq:disc:split}
    \frac{\partial f}{\partial x} = \alpha\frac{\partial (ab)}{\partial x} + (1 - \alpha)\left(b\frac{\partial a}{\partial x} + a\frac{\partial b}{\partial x}\right),
\end{equation}
for some value of~$\alpha$.
It can be \chg{proved} that the derivative operator of the DGSEM can be used to define the discretized form of this split-form operators.
Using a two-point flux,~\chg{$F^{\#}_{(ij)} = F^{\#}(Q_i, Q_j)$}, that connects all the nodes in the direction of the \chg{derivative,~\cite{Gassner2016}}
\begin{equation}\chgenv
\label{eq:disc:twopoint}
    \left.\frac{\partial f}{\partial x}\right|_{x=\xi_i} \approx \sum_{j=0}^P D_{ij}F_j = \sum_{j=0}^P 2D_{ij} F^{\#}_{(ij)}.
\end{equation}
\chg{In higher dimensions, the parenthesis always indicate the indices that differ between the two terms of the flux.
For instance, in three dimensions,~${\stvec{F}^{\#}_{(i\alpha)jk} = \stvec{F}^{\#}(\stvec{Q}_{ijk}, \stvec{Q}_{\alpha jk})}$.}
The application of all these previous concepts into \cref{eq:ns:advdiff_ldg_weak} yields the expression of the time derivative for each degree of freedom:
\begin{widetext}
\begin{subequations}\chgenv
\label{eq:disc:advdiff_dgsem}
\begin{equation}
\begin{gathered}
    \jac_{ijk}\stvec{Q}_{t,ijk} + \mathbb{D}^{\#}_{ijk}\left(\cbvec{F}_e\right) + \mathbb{F}_{ijk}\left(\stvec{F}_e^{\star} - \bvec{F}_e\cdot{\svec{n}}\right) = \mathbb{D}_{ijk}\left(\cbvec{F}_v + \cbvec{F}_a\right) + \mathbb{F}\left(\stvec{F}_v^{\star} + \stvec{F}_a^{\star}\right), \\
    \jac_{ijk}\bvec{G}_{ijk} = \bmat{M}\cdot\mathbb{D}^G_{ijk}\left(\stvec{W}\right) + \mathbb{F}\left(\bvec{W}^{\star} - \stvec{W}\cdot\svec{n}\right),
\end{gathered}
\end{equation}
\begin{equation}
\label{eq:disc:advdiff_dgsem:defs}
\begin{gathered}
    \mathbb{D}_{ijk}\left(\bvec{F}\right) = \sum_{\alpha=0}^P D_{i\alpha} \stvec{F}_{\alpha jk} + \sum_{\alpha=0}^P D_{j\alpha} \stvec{G}_{i\alpha k} + \sum_{\alpha=0}^P D_{k\alpha} \stvec{H}_{ij\alpha}, \\
    \mathbb{D}^{\#}_{ijk}\left(\bvec{F}\right) = \sum_{\alpha=0}^N 2D_{i\alpha}\stvec{F}^{\#}_{(i\alpha)jk} + \sum_{\alpha=0}^N 2D_{j\alpha}\stvec{G}^{\#}_{i(j\alpha)k} + \sum_{\alpha=0}^N 2D_{k\alpha}\stvec{H}^{\#}_{ij(k\alpha)}, \\
    \mathbb{D}^G_{ijk}\left(\stvec{F}\right) = \left(\sum_{\alpha=0}^P D_{i\alpha} \stvec{F}_{\alpha jk}, \sum_{\alpha=0}^P D_{j\alpha} \stvec{F}_{i\alpha k}, \sum_{\alpha=0}^P D_{k\alpha} \stvec{F}_{ij\alpha}\right)^T, \\
    \mathbb{F}_{ijk}\left(\stvec{F}^{\star}\right) = \frac{\stvec{F}^{\star}_{Njk}l_i(1) - \stvec{F}^{\star}_{0jk}l_i(-1)}{\omega_i} + \frac{\stvec{F}^{\star}_{iNk}l_j(1) - \stvec{F}^{\star}_{i0k}l_j(-1)}{\omega_j} + \frac{\stvec{F}^{\star}_{ijN}l_k(1) - \stvec{F}^{\star}_{ij0}l_k(-1)}{\omega_k},
\end{gathered}
\end{equation}
\end{subequations}
\end{widetext}
\chg{where~$\omega_i$} is the quadrature weight associated to the~$i$-th node.

\chg{In this work we also utilize a different formulation of the advection operator,~$\mathbb{D}^{\#}$, as introduced in \cref{sub:sdisc}.
Using the telescopic property of the DGSEM split-from derivative, the advection operator can also be written as
\begin{equation}
\label{eq:disc:fv}
\begin{aligned}
    \mathbb{D}^{\#}_{ijk}\left(\bvec{F}\right) &= \frac{\hat{\stvec{F}}_{(i,i+1)jk} - \hat{\stvec{F}}_{(i-1,i)jk}}{\omega_i} \\
    &+ \frac{\hat{\stvec{G}}_{i(j,j+1)k} - \hat{\stvec{G}}_{i(j-1,j)k}}{\omega_j} \\
    &+ \frac{\hat{\stvec{H}}_{ij(k,k+1)} - \hat{\stvec{H}}_{ij(k-1,k)}}{\omega_k},
\end{aligned}
\end{equation}
resembling a finite volume scheme for an element of size $\omega_i \times \omega_j \times \omega_k$ and interface fluxes $\hat{\stvec{F}}$, $\hat{\stvec{G}}$ and $\hat{\stvec{H}}$.
For \cref{eq:disc:fv} to be equivalent to its definition in \cref{eq:disc:advdiff_dgsem:defs}, the interface fluxes must take a specific value~\cite{Fisher2013}:
\begin{equation}
\label{eq:disc:ho_fluxes}
\begin{gathered}
    \hat{\stvec{F}}^{\text{DG}}_{(i,i+1)} = \sum_{k=i+1}^N\sum_{l=0}^i 2\omega_l D_{lk}\stvec{F}^{\#}_{(lk)}, \quad i = 0, \dots, N - 1, \\
    \hat{\stvec{F}}_{(-1,0)} = \stvec{F}_0, \quad \hat{\stvec{F}}_{(N,N+1)} = \stvec{F}_N.
\end{gathered}
\end{equation}
However, this framework enables further possibilities, as this analogy with finite volume methods can be helpful to enhance the stability.
If dissipative Riemann solvers are used to compute the sub-cell fluxes instead of \cref{eq:disc:ho_fluxes}, the discretization of \cref{eq:disc:advdiff_dgsem} adds numerical viscosity not only through the face operator~$\mathbb{F}$, but also with the action of the volume term~$\mathbb{D^{\#}}$.}

\chg{The approach that we follow in this work (specifically in \cref{sub:sedov,sub:dmr}) combines a high- and a low-order formulation to provide a hybrid DG-FV scheme~\cite{HENNEMANN2021109935,RUEDARAMIREZ2022105627,lin2023high} with sub-cell resolution.
Since the resulting methodology should be equivalent to the high-order scheme wherever possible, the weight of each component in the hybrid method is regulated by a shock sensor, according to \cref{eq:sdisc:hybrid,eq:sdisc:alpha}.}

\section{Tests with other variables}
\label{sec:aVar}

This section collects several results obtained for the cases of \cref{sub:inviscid,sub:viscous} with variables different from those proposed in the main text.
The purpose of this appendix is first to show other alternatives that may work better when used in combination with numerical setups different from ours. And secondly, to give a deeper insight into alternative feature spaces and show why the variables chosen in this work perform well in both inviscid and viscous cases.

\begin{figure*}[htpb]
    \subfloat[$p$ with $s_0 = -3$ and $\Delta s = 1$.]{
    \begin{minipage}{0.4\textwidth}
        \includegraphics[width=\textwidth]{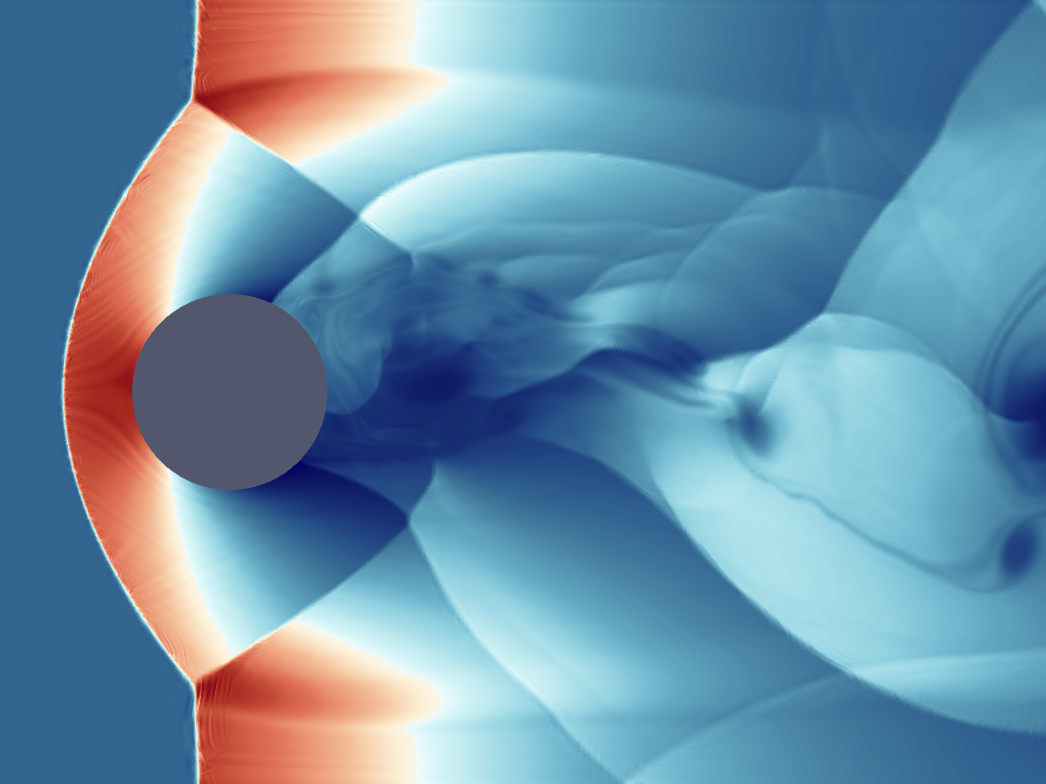}
    \end{minipage}\qquad
    \begin{minipage}{0.4\textwidth}
        \includegraphics[width=\textwidth]{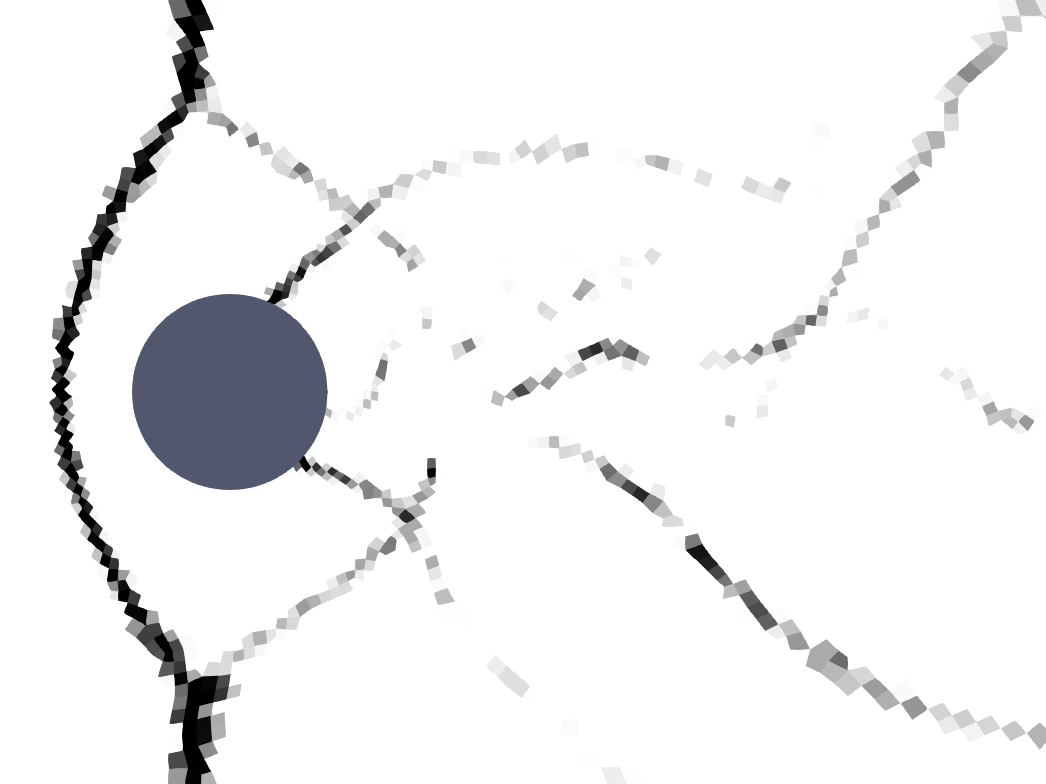}
    \end{minipage}
    }\\
    \subfloat[$\rho$ with $s_0 = -3.5$ and $\Delta s = 1$.]{
    \begin{minipage}{0.4\textwidth}
        \includegraphics[width=\textwidth]{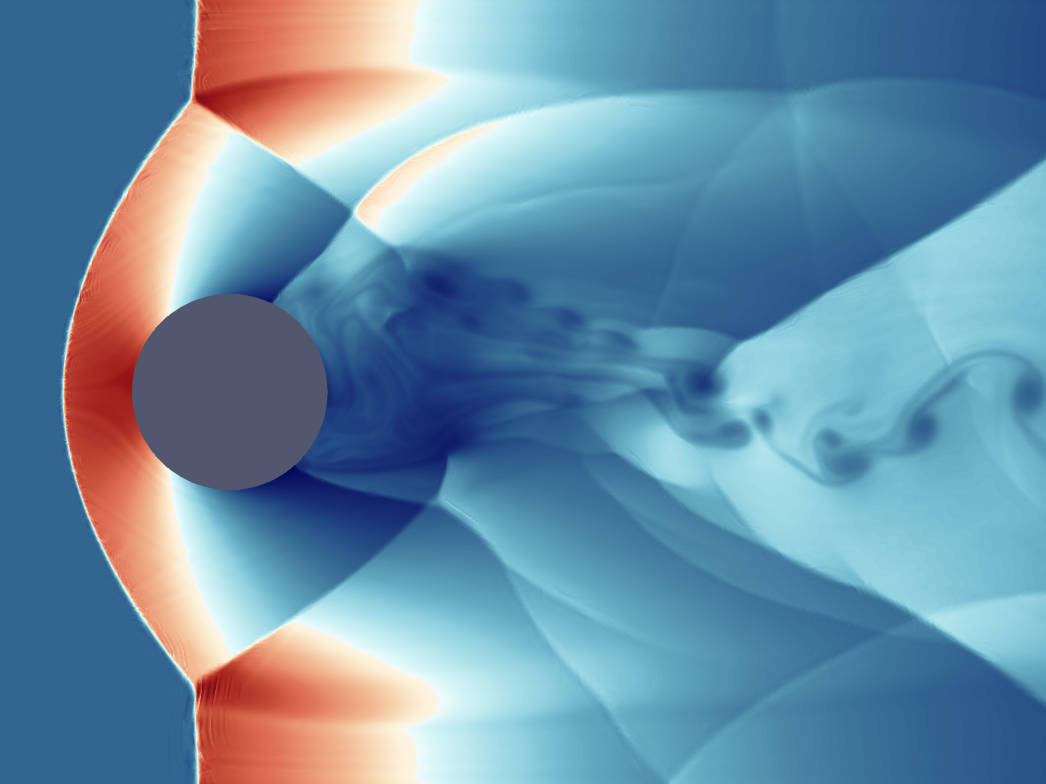}
    \end{minipage}\qquad
    \begin{minipage}{0.4\textwidth}
        \includegraphics[width=\textwidth]{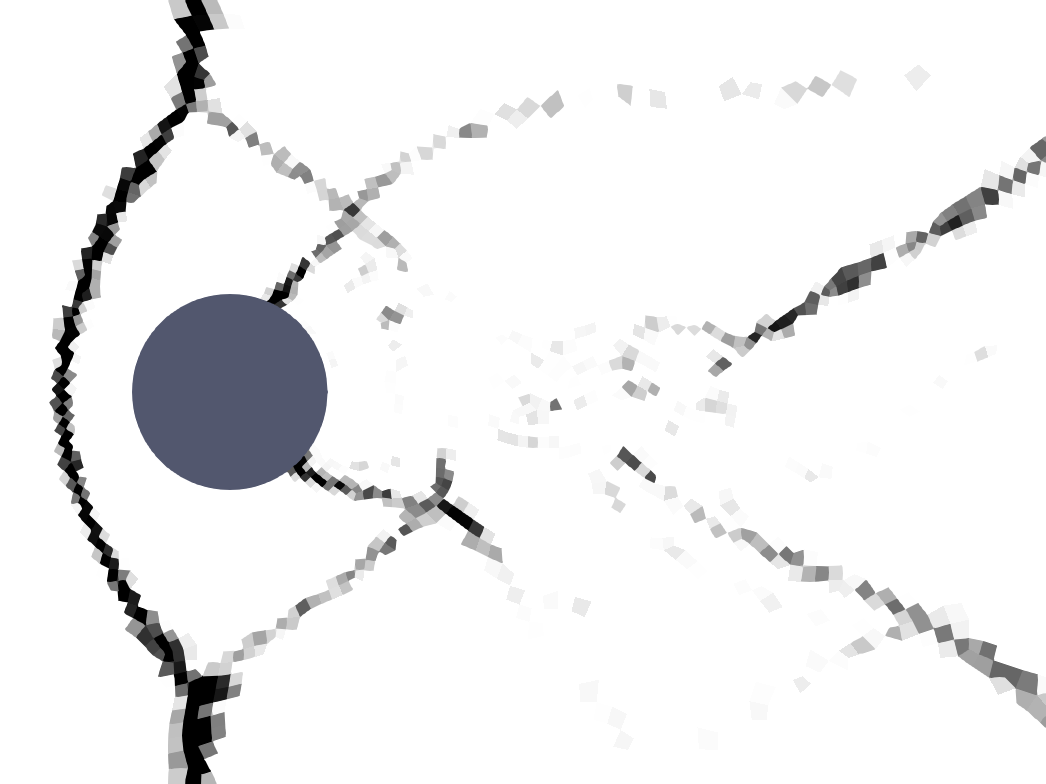}
    \end{minipage}
    }
    \caption{Inviscid case after 300,000 iterations with the modal sensor of \cref{sub:sens:modal}.}
    \label{fig:other:inviscid_modal}
\end{figure*}

\begin{figure*}[htpb]
    \subfloat[$p$ with $s_0 = -3.5$ and $\Delta s = 1$.]{
    \begin{minipage}{0.4\textwidth}
        \includegraphics[width=\textwidth]{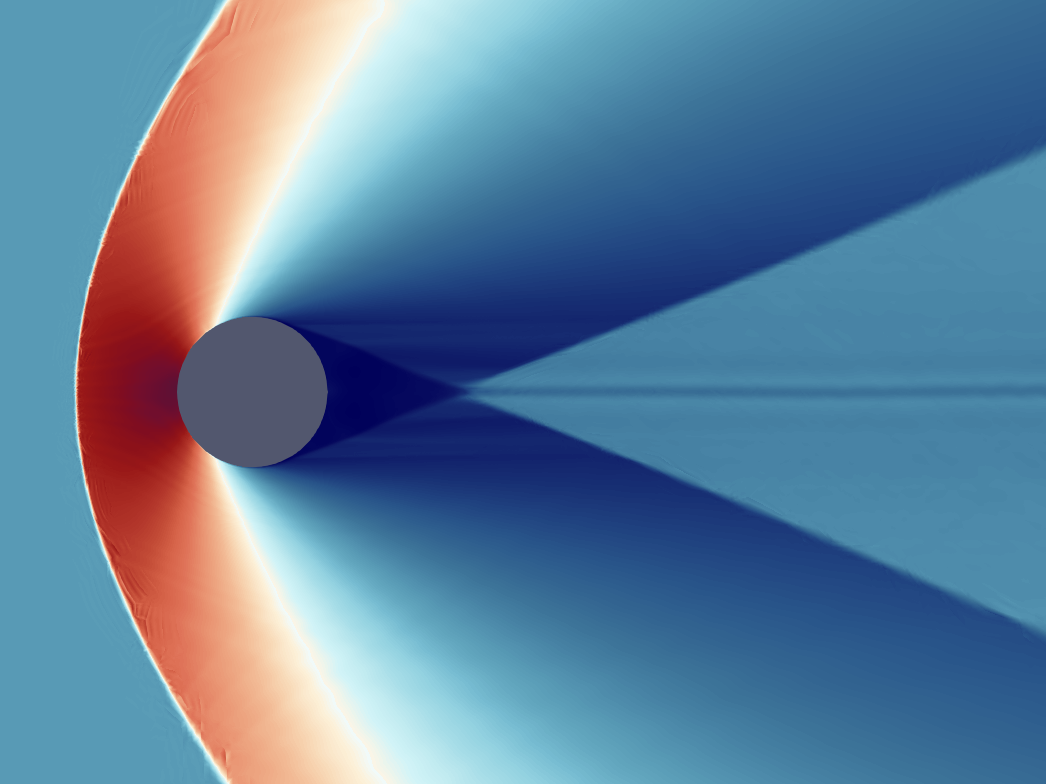}
    \end{minipage}\qquad
    \begin{minipage}{0.4\textwidth}
        \includegraphics[width=\textwidth]{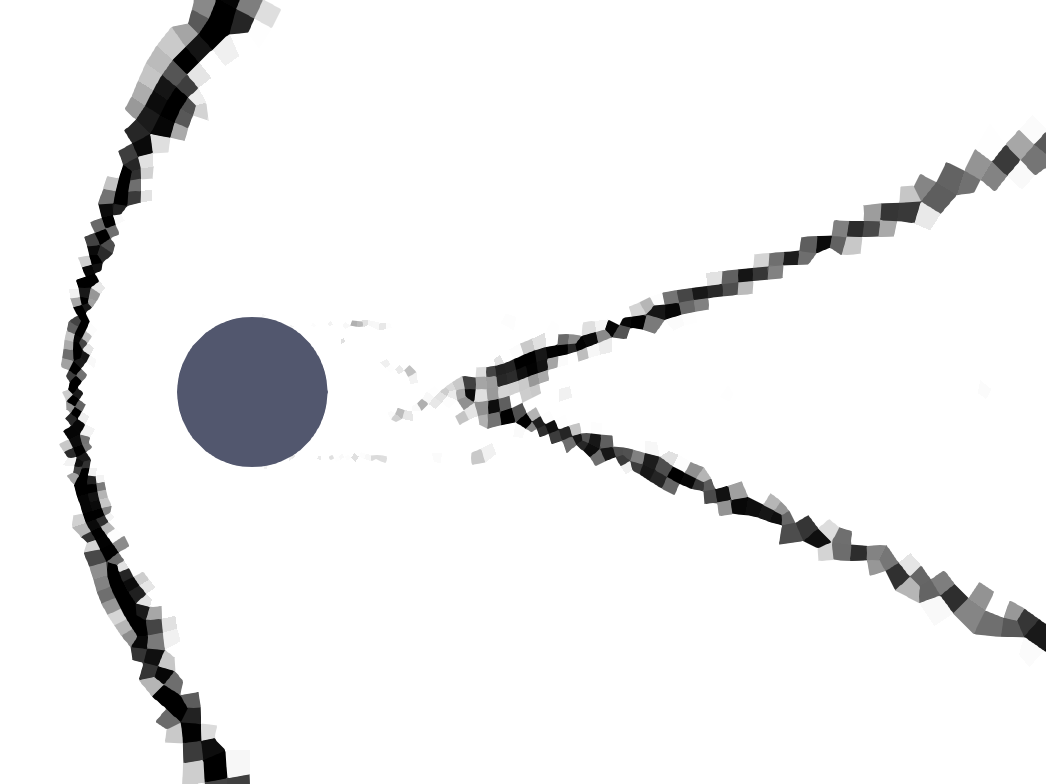}
    \end{minipage}
    }\\
    \subfloat[$\rho$ with $s_0 = -3.5$ and $\Delta s = 1$.]{
    \begin{minipage}{0.4\textwidth}
        \includegraphics[width=\textwidth]{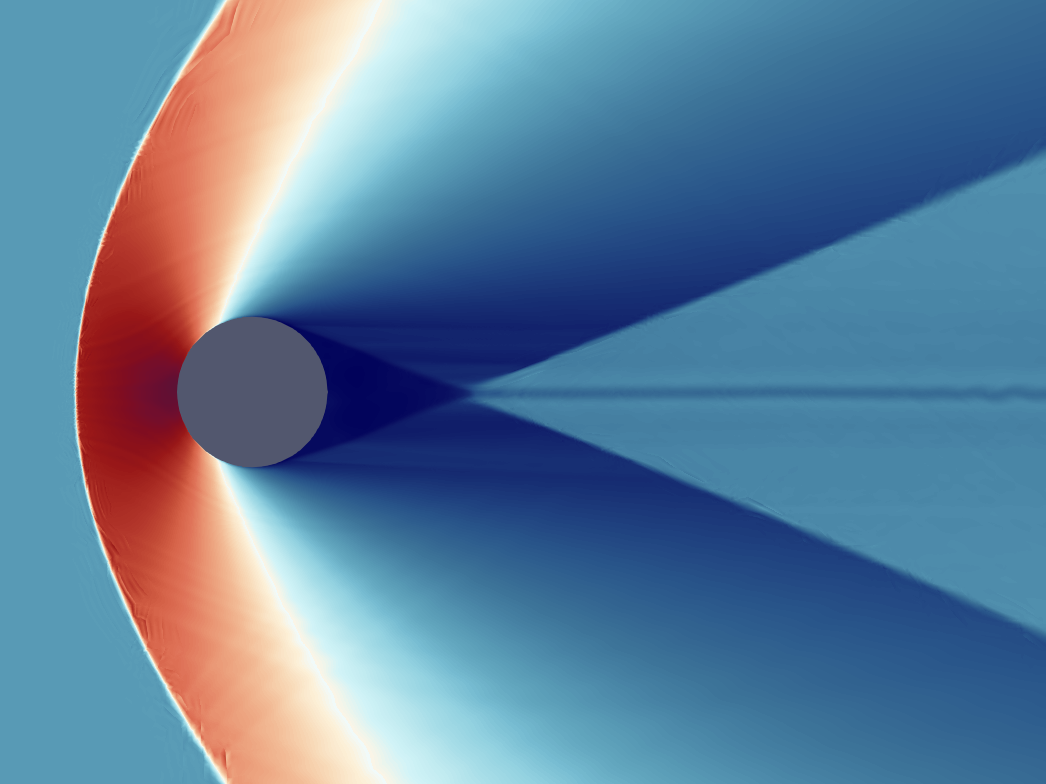}
    \end{minipage}\qquad
    \begin{minipage}{0.4\textwidth}
        \includegraphics[width=\textwidth]{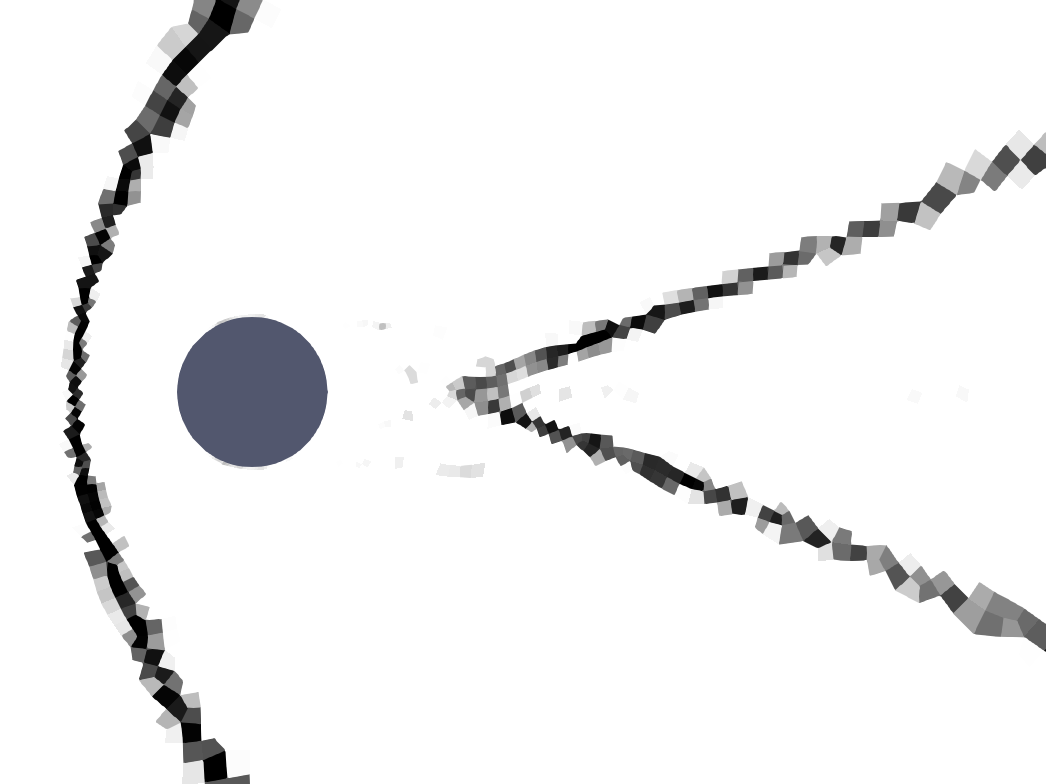}
    \end{minipage}
    }
    \caption{Viscous case after 300,000 iterations with the modal sensor of \cref{sub:sens:modal}.}
    \label{fig:other:viscous_modal}
\end{figure*}

\Cref{fig:other:inviscid_modal,fig:other:viscous_modal} display the solution of the inviscid and viscous test cases, respectively, when using the modal sensor  of \cref{sub:sens:modal}.
Both,~$p$ and~$\rho$ seem to give good results in the inviscid setup, with good sub-cell accuracy in the shocks and no dissipation in the wake.
However, the discretization is over-dissipative for the viscous case, and the vortices that appear behind the cylinder are not visible with these sensors.
The simulation shows a steady flow that does not capture the physics of the problem.

\begin{figure*}[htpb]
    \subfloat[$(\nabla\cdot \svec{v})^2$ with $s_0 = 125$ and $\Delta s = 75$.]{
    \begin{minipage}{0.4\textwidth}
        \includegraphics[width=\textwidth]{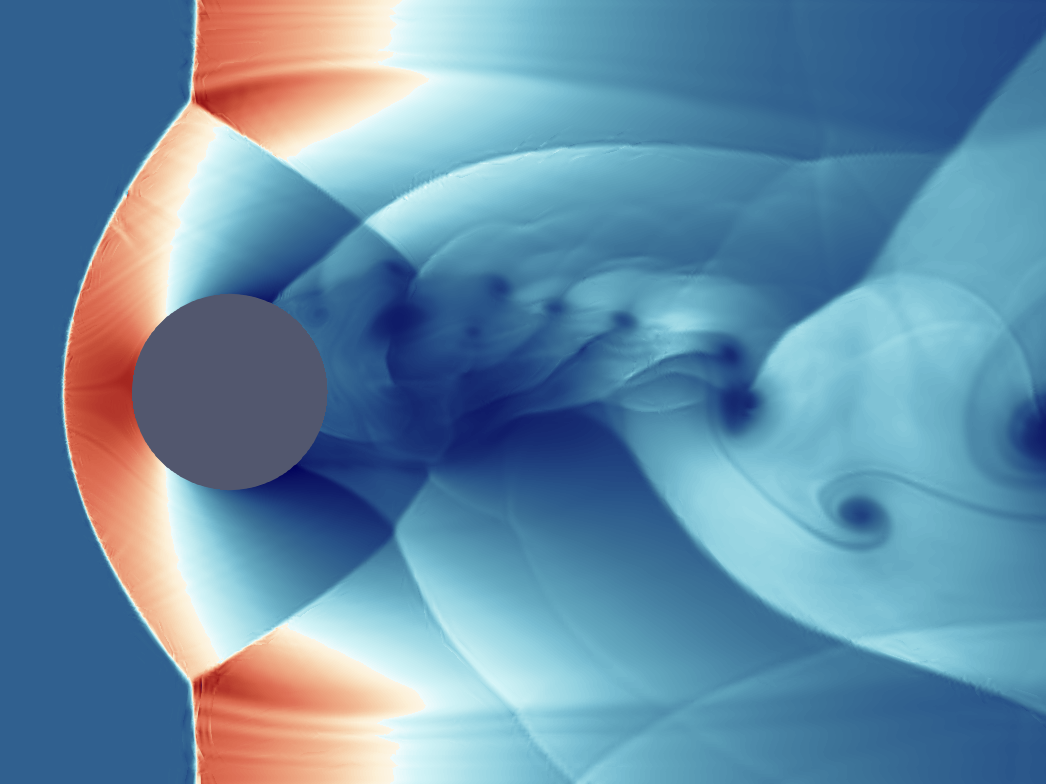}
    \end{minipage}\qquad
    \begin{minipage}{0.4\textwidth}
        \includegraphics[width=\textwidth]{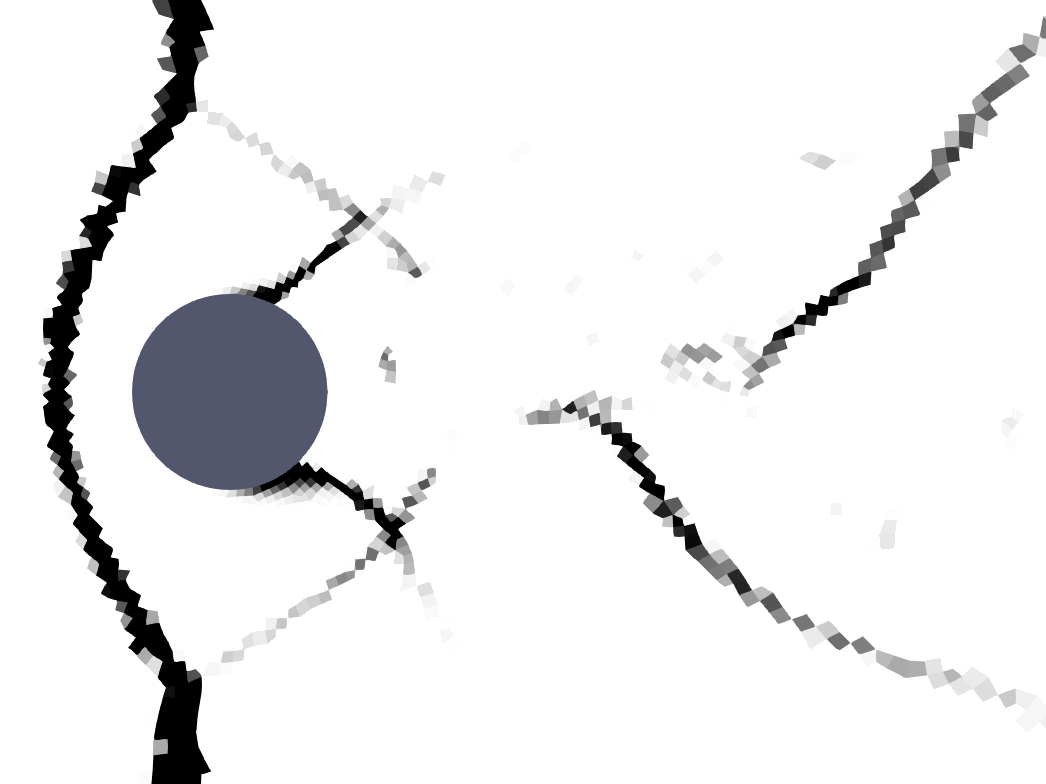}
    \end{minipage}
    }\\
    \subfloat[$\nabla\rho\cdot \svec{n}_v$ with $s_0 = 10.05$ and $\Delta s = 9.95$.]{
    \begin{minipage}{0.4\textwidth}
        \includegraphics[width=\textwidth]{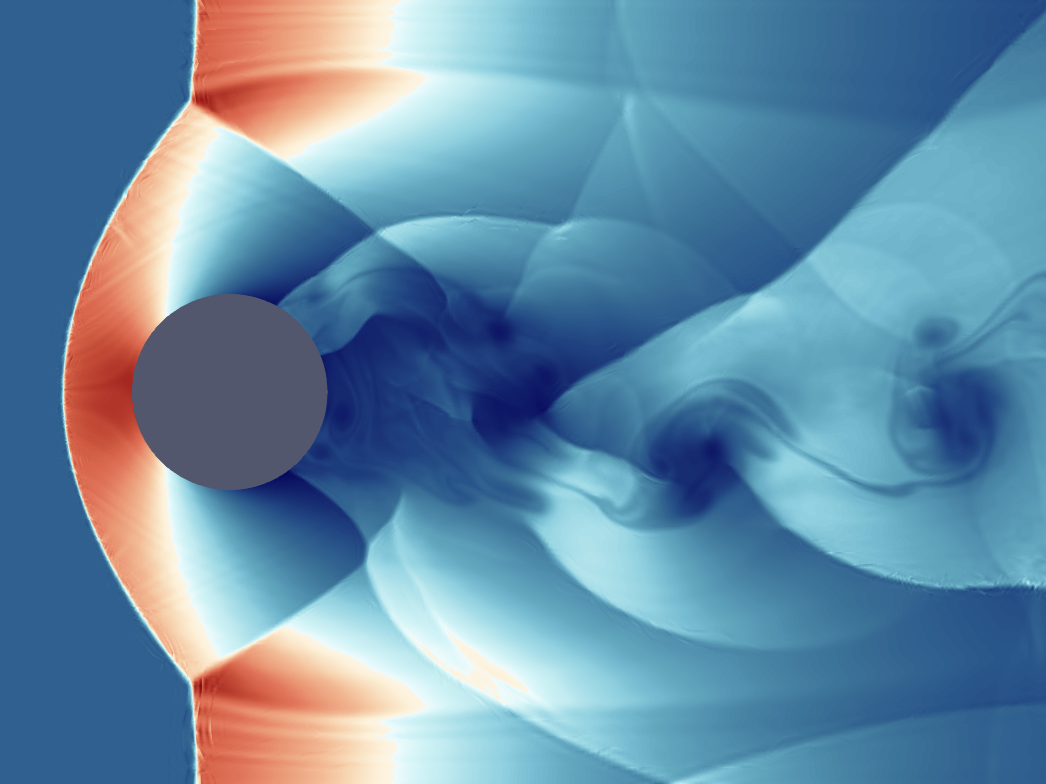}
    \end{minipage}\qquad
    \begin{minipage}{0.4\textwidth}
        \includegraphics[width=\textwidth]{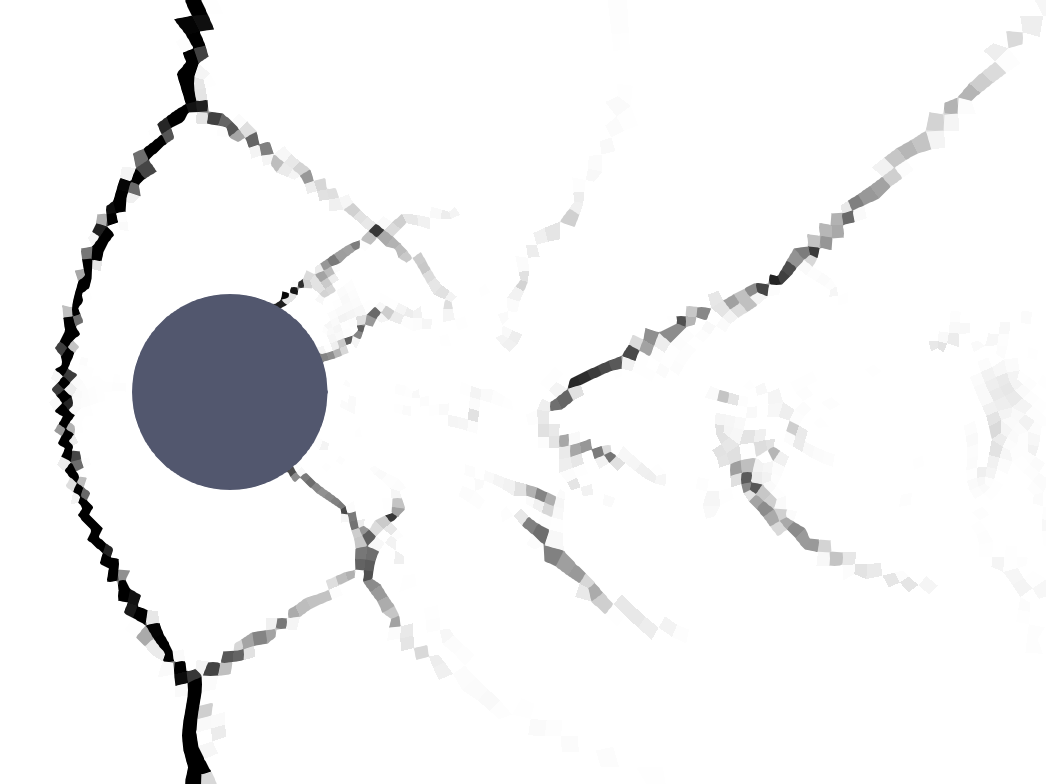}
    \end{minipage}
    }
    \caption{Inviscid case after 300,000 iterations with the integral sensor of \cref{sub:sens:integral}.}
    \label{fig:other:inviscid_integral}
\end{figure*}

\begin{figure*}[htpb]
    \subfloat[$(\nabla\cdot \svec{v})^2$ with $s_0 = 125$ and $\Delta s = 75$.]{
    \begin{minipage}{0.4\textwidth}
        \includegraphics[width=\textwidth]{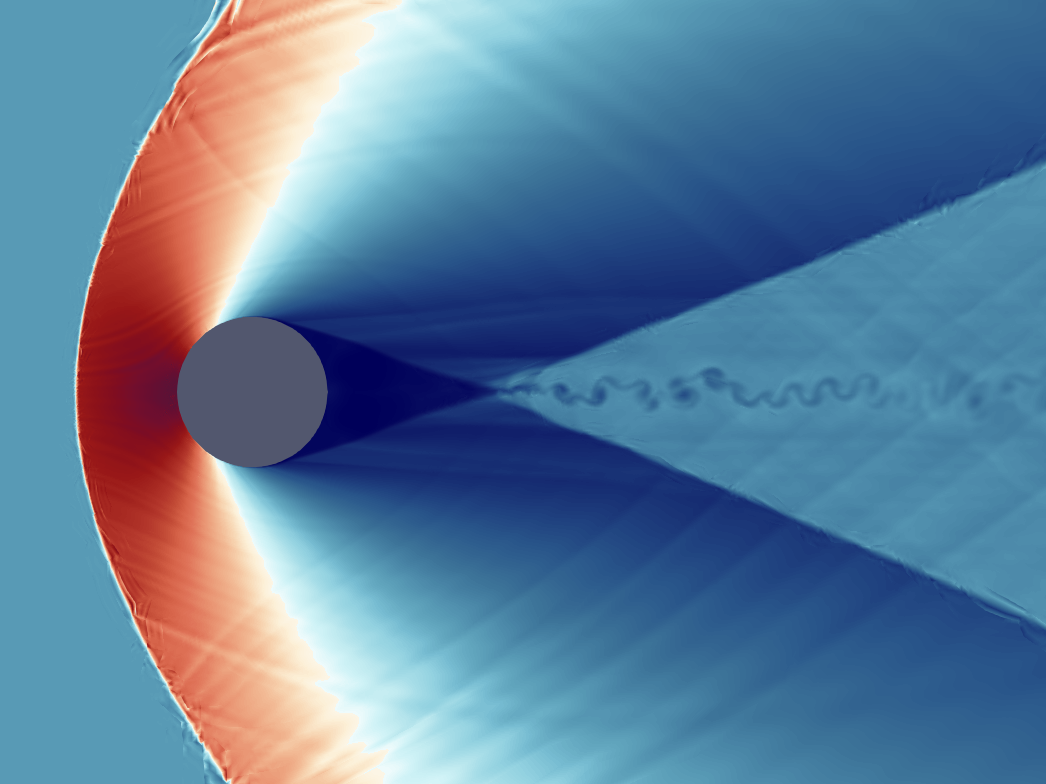}
    \end{minipage}\qquad
    \begin{minipage}{0.4\textwidth}
        \includegraphics[width=\textwidth]{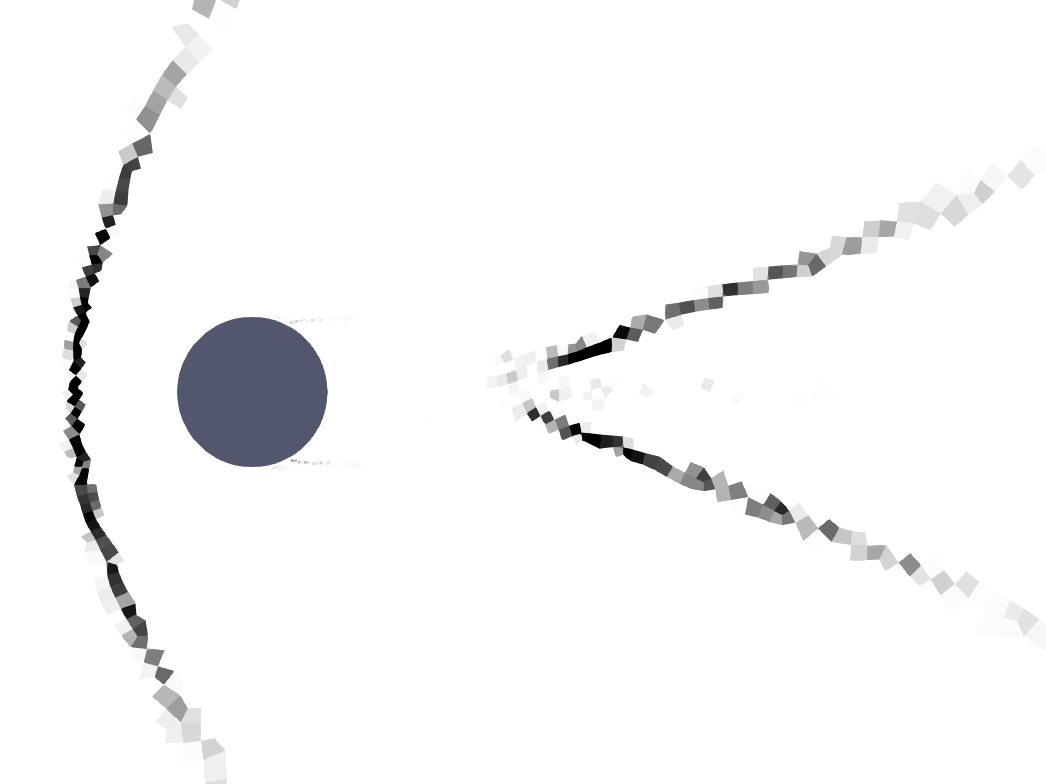}
    \end{minipage}
    }\\
    \subfloat[$\nabla\rho\cdot \svec{n}_v$ with $s_0 = 15.05$ and $\Delta s = 14.95$.]{
    \begin{minipage}{0.4\textwidth}
        \includegraphics[width=\textwidth]{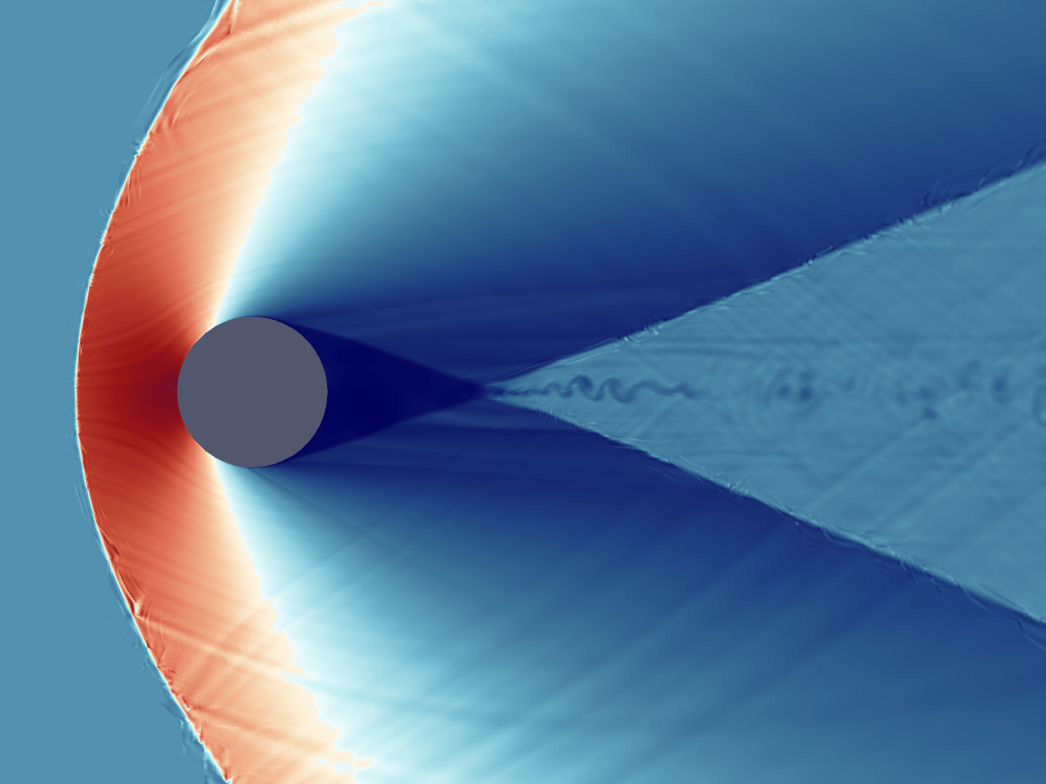}
    \end{minipage}\qquad
    \begin{minipage}{0.4\textwidth}
        \includegraphics[width=\textwidth]{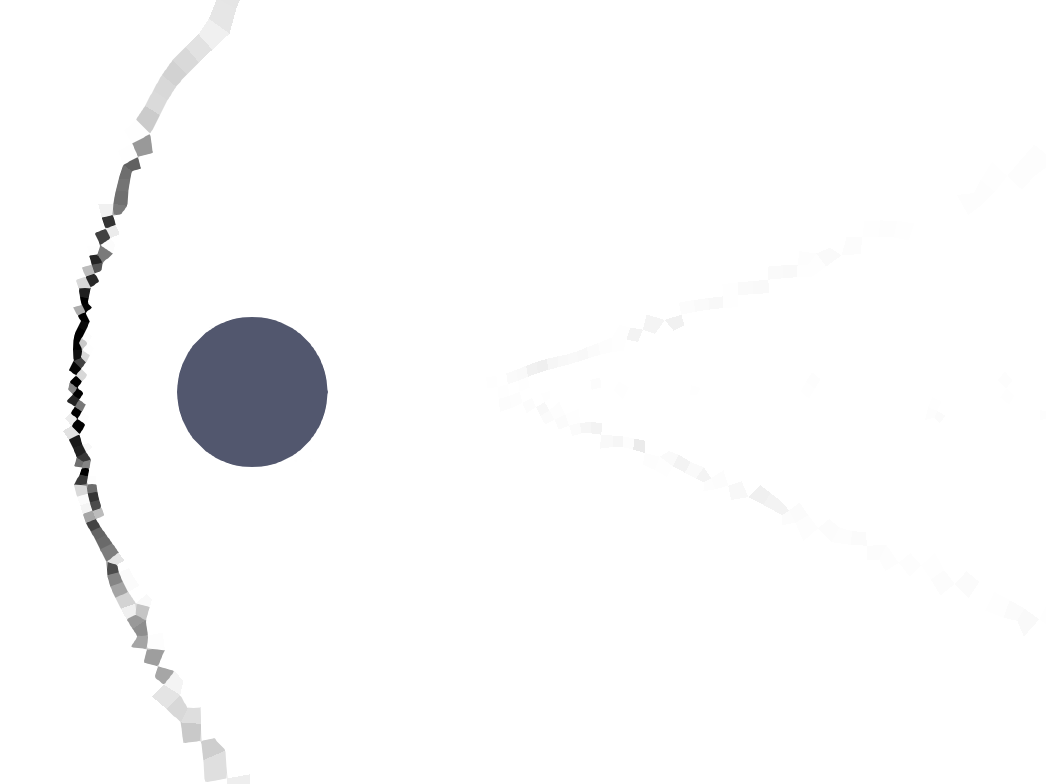}
    \end{minipage}
    }
    \caption{Viscous case after 300,000 iterations with the integral sensor of \cref{sub:sens:integral}.}
    \label{fig:other:viscous_integral}
\end{figure*}

The integral sensor of \cref{sub:sens:integral} with the divergence of the velocity and the projection of the density gradient along the direction of the velocity also shows good results in the inviscid case of \cref{fig:other:inviscid_integral}.
Some noise is generated at strong shocks, and this is more clear in the viscous case (\cref{fig:other:viscous_integral}).
With this configuration, the sensor cannot properly separate the shock waves from the boundary layer and therefore, we had to decrease its sensitivity to avoid detecting both.
Dissipation in the main shock wave is not sufficient, and the oscillations travel downstream polluting the solution and interacting with the shock and vortices of the wake.

\begin{figure*}[htpb]
    \centering
    \begin{minipage}{0.3\textwidth}
        \includegraphics[width=\textwidth]{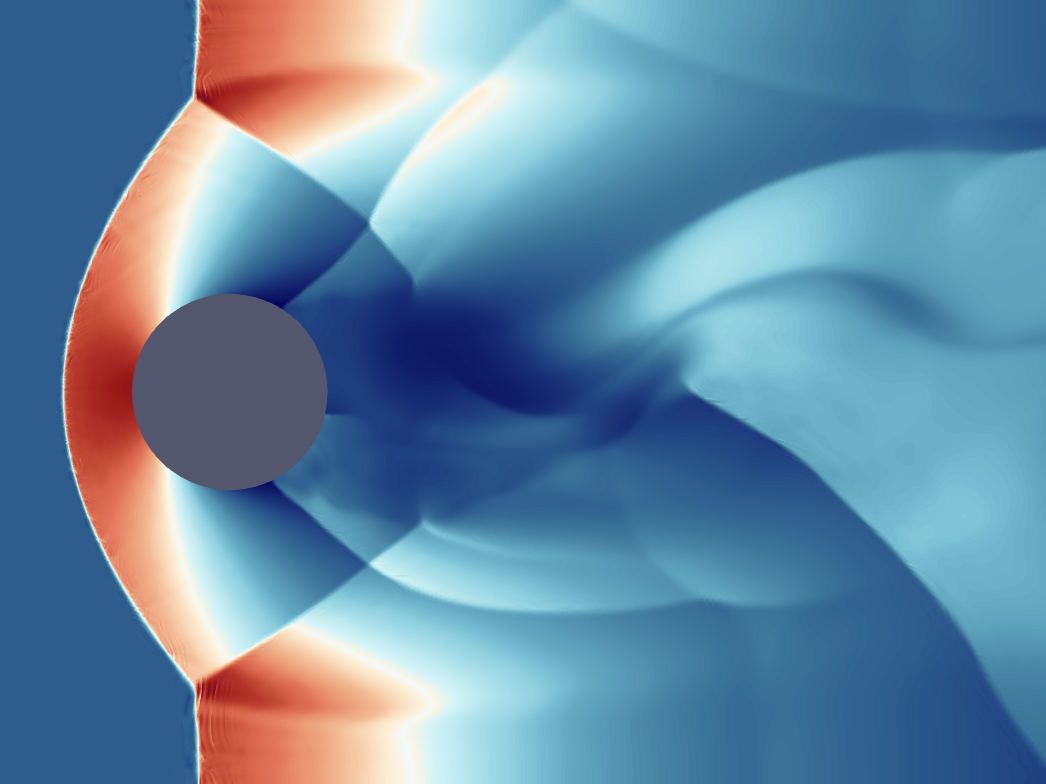}
    \end{minipage}\qquad
    \begin{minipage}{0.3\textwidth}
        \includegraphics[width=\textwidth]{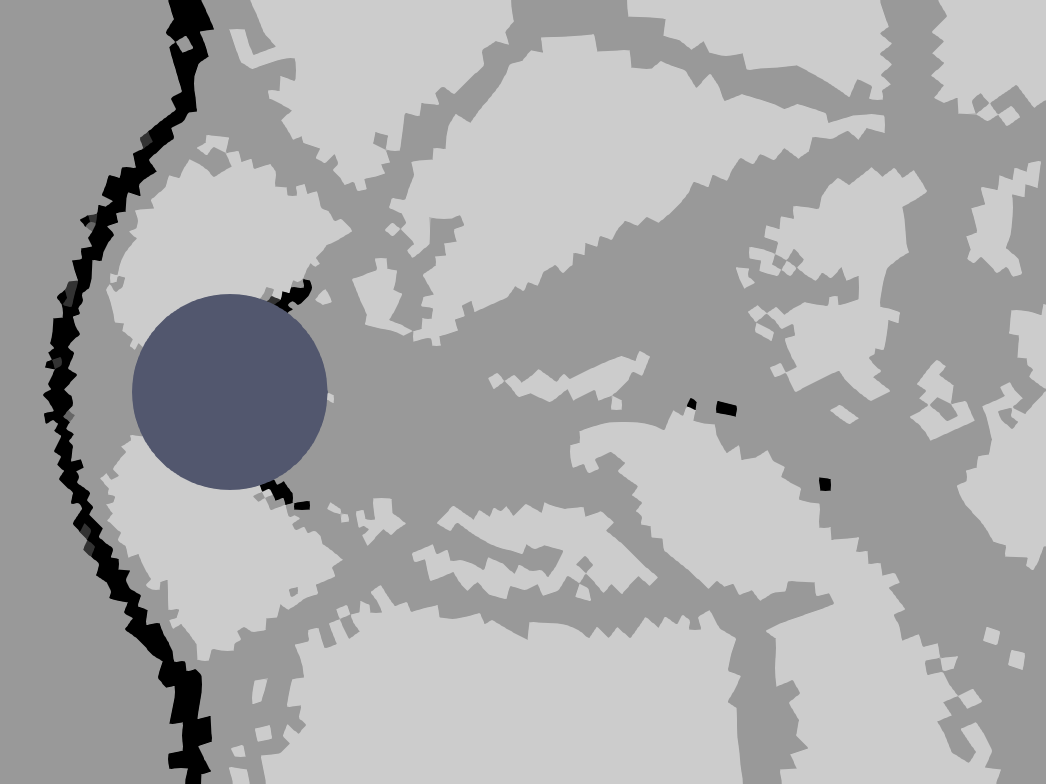}
    \end{minipage}\qquad
    \begin{minipage}{0.3\textwidth}
        \includegraphics[width=\textwidth]{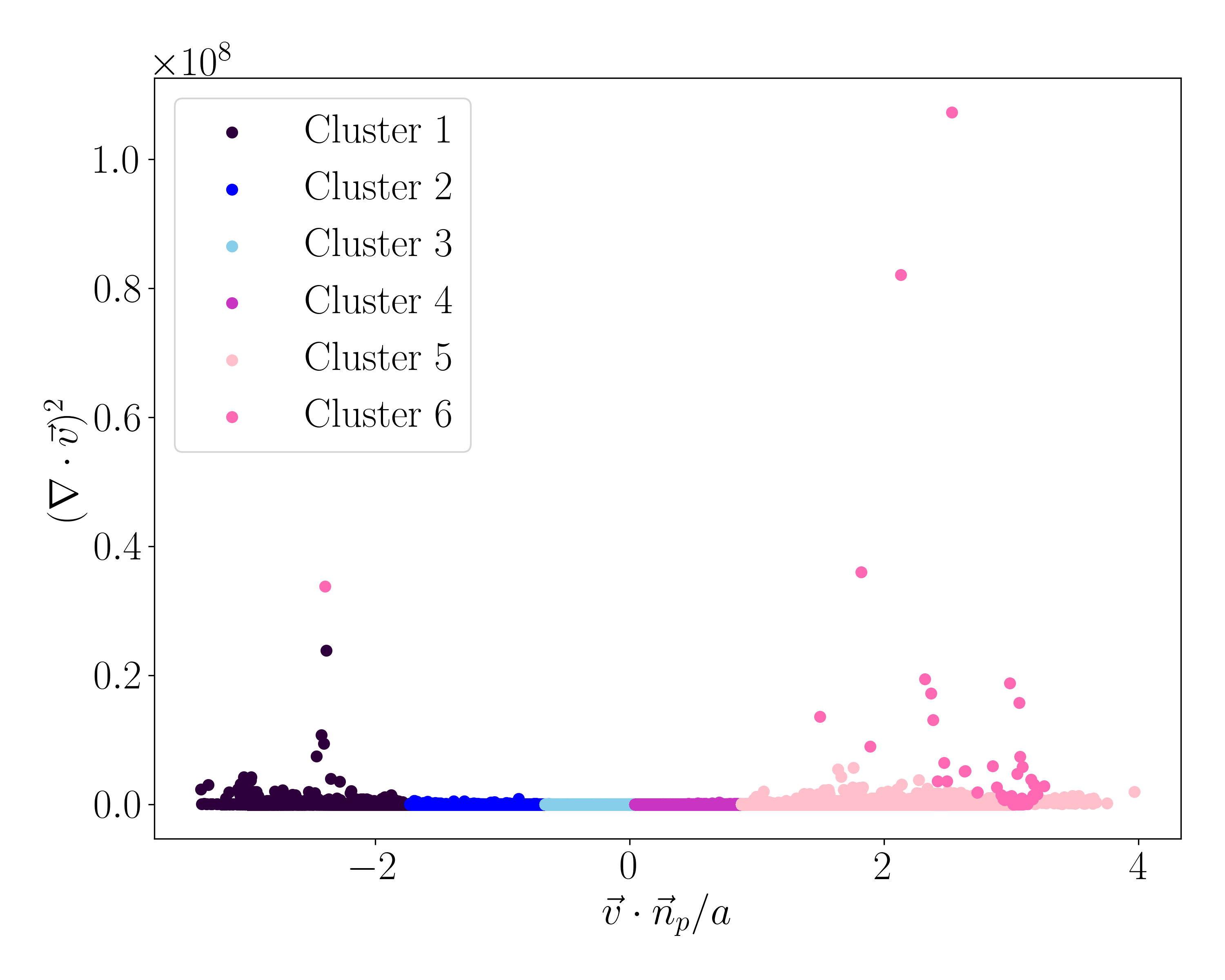}
    \end{minipage}
    \caption{Inviscid case after 300,000 iterations with our adaptive GMM sensor of \cref{sec:gmm} using 6 clusters and the variables $\svec{v}\cdot \svec{n}_p/a, (\nabla\cdot \svec{v})^2$.}
    \label{fig:other:inviscid_gmm}
\end{figure*}

We finally discuss the performance of our GMM-based sensor under different choices of feature \chg{space, defining the following set of possible variables~\cite{saettaIdentificationFlowFielda}}:
\begin{itemize}
    \item $(\nabla\cdot\svec{v})^2$,
    \item $\svec{v}\cdot\svec{n}_p / a, (\nabla\cdot\svec{v})^2$,
    \item $\max(0, M-1), (\nabla\cdot\svec{v})^2$,
\end{itemize}
where~$a$ is the speed of sound,~$\svec{n}_p$ is the unitary vector in the direction of the pressure gradient, and~$M$ is the Mach number.
The results obtained with the inviscid flow (pictured in \cref{fig:other:inviscid_gmm}) show that none of the variables considered in these tests is suitable for this setup, performing significantly better in the viscous case.
The sensor that utilized the Mach number of the velocity in the direction of the pressure gradient was the only one that successfully completed the simulation, while the others crashed during the initial transient.
And even in this case, it exhibited excessive dissipation.

\begin{figure*}[htpb]
    \subfloat[$(\nabla\cdot \svec{v})^2$.]{
    \begin{minipage}{0.3\textwidth}
        \includegraphics[width=\textwidth]{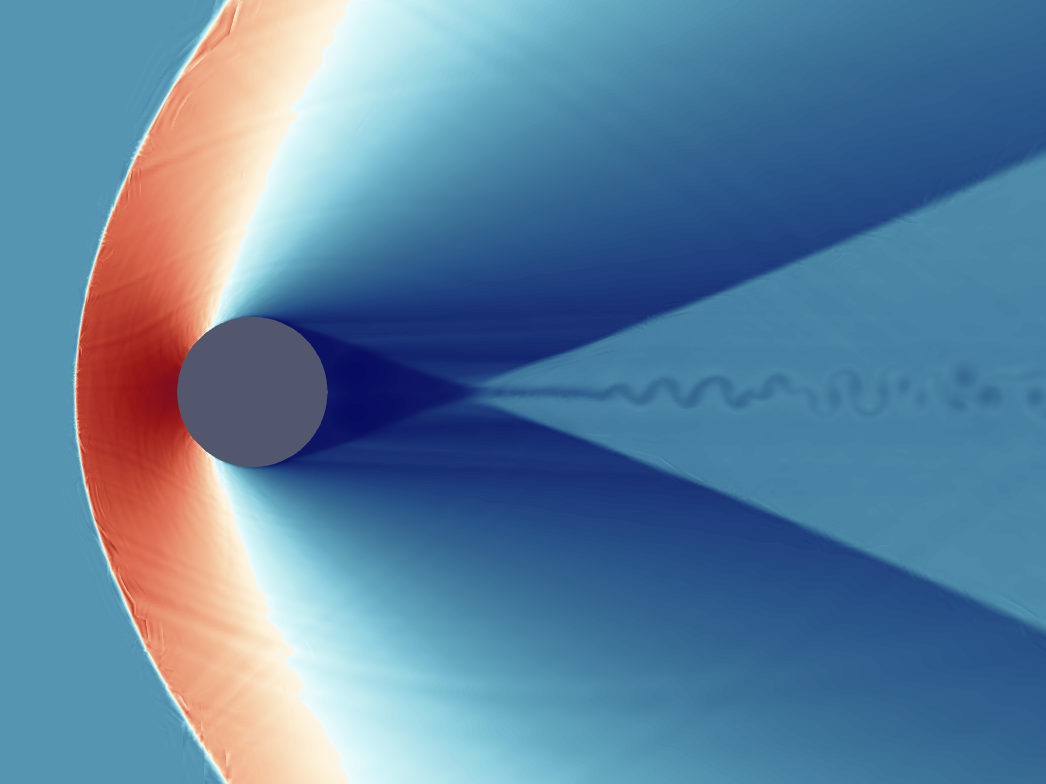}
    \end{minipage}\qquad
    \begin{minipage}{0.3\textwidth}
        \includegraphics[width=\textwidth]{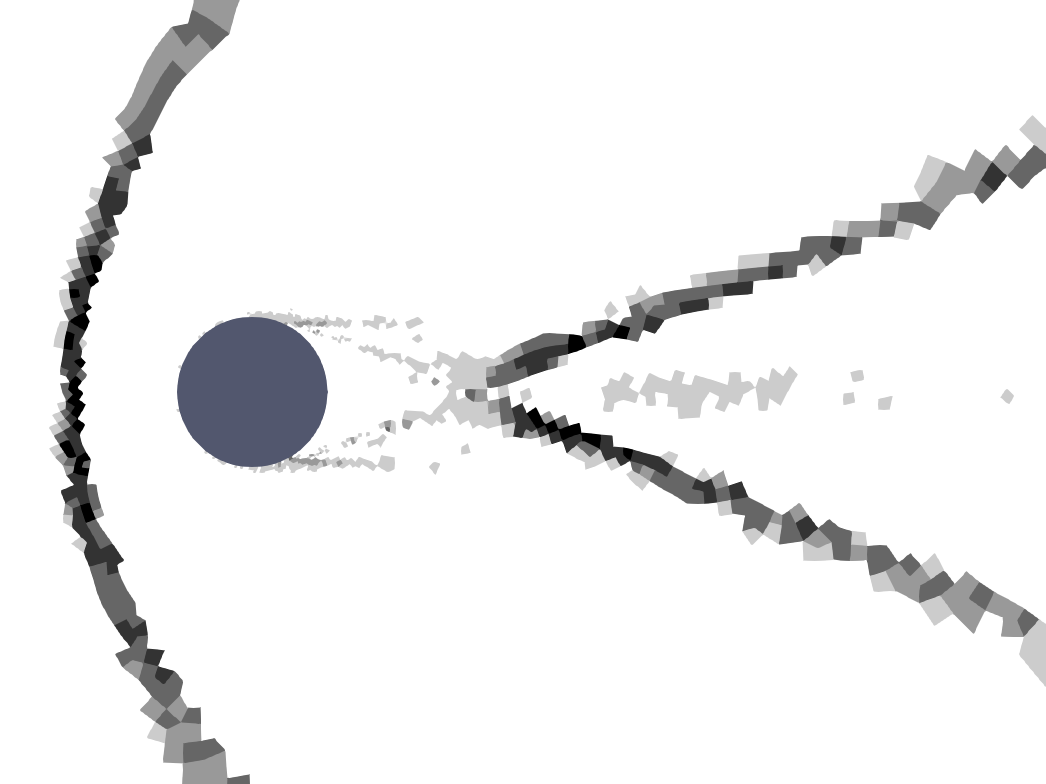}
    \end{minipage}\qquad
    \begin{minipage}{0.3\textwidth}
        \includegraphics[width=\textwidth]{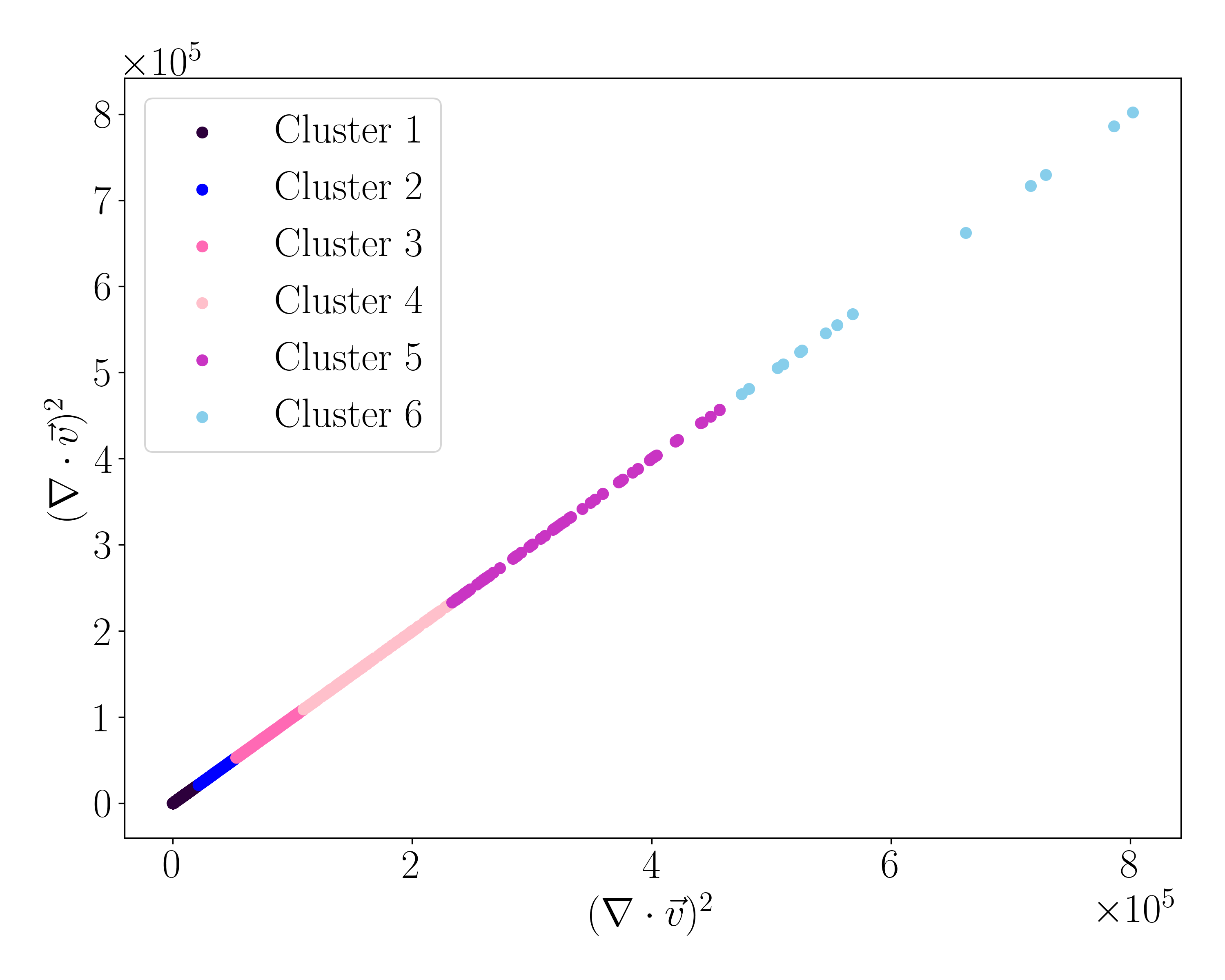}
    \end{minipage}
    }\\
    \subfloat[$\svec{v}\cdot \svec{n}_p/a, (\nabla\cdot \svec{v})^2$.]{
    \begin{minipage}{0.3\textwidth}
        \includegraphics[width=\textwidth]{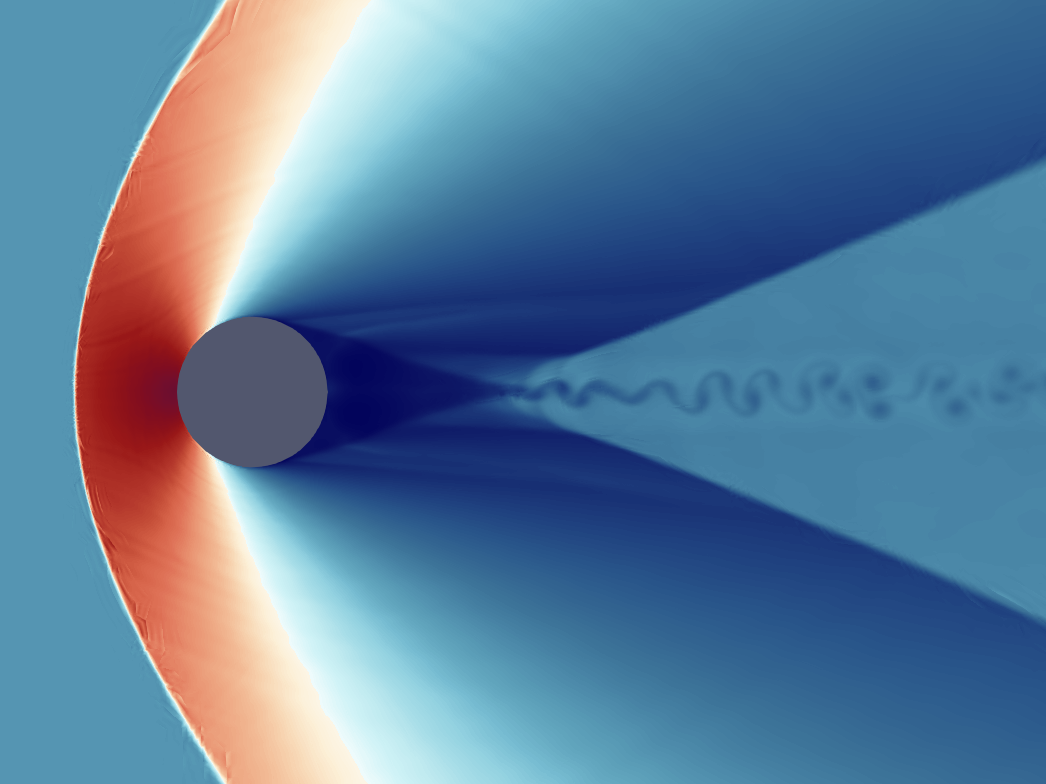}
    \end{minipage}\qquad
    \begin{minipage}{0.3\textwidth}
        \includegraphics[width=\textwidth]{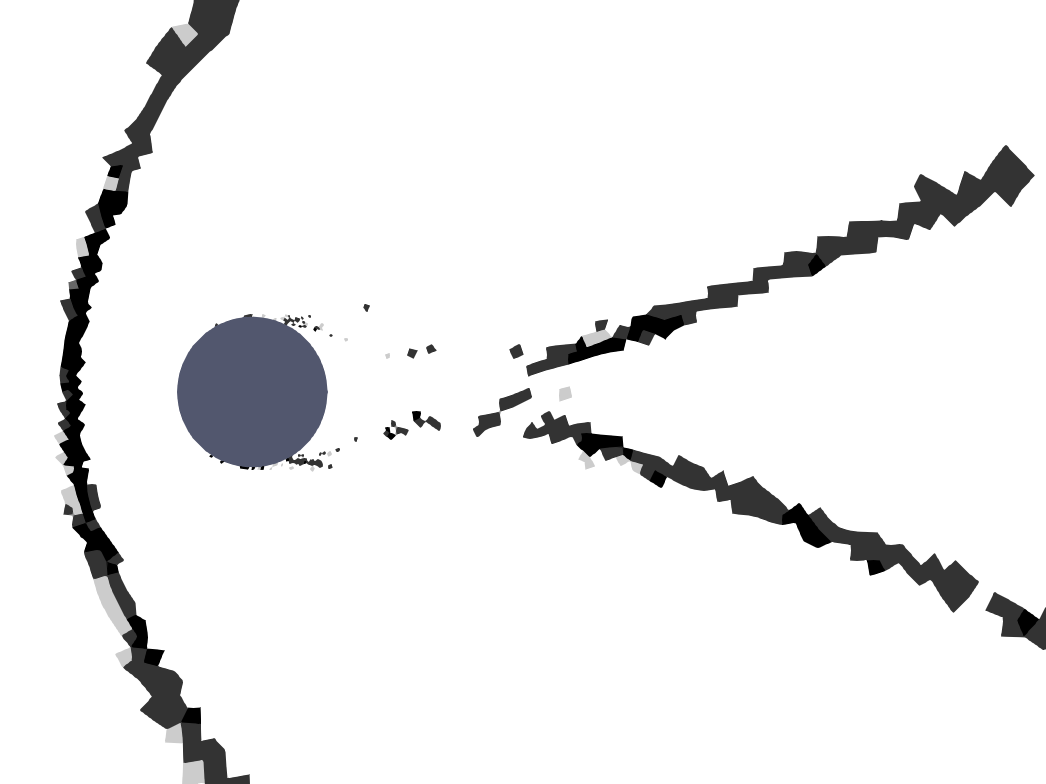}
    \end{minipage}\qquad
    \begin{minipage}{0.3\textwidth}
        \includegraphics[width=\textwidth]{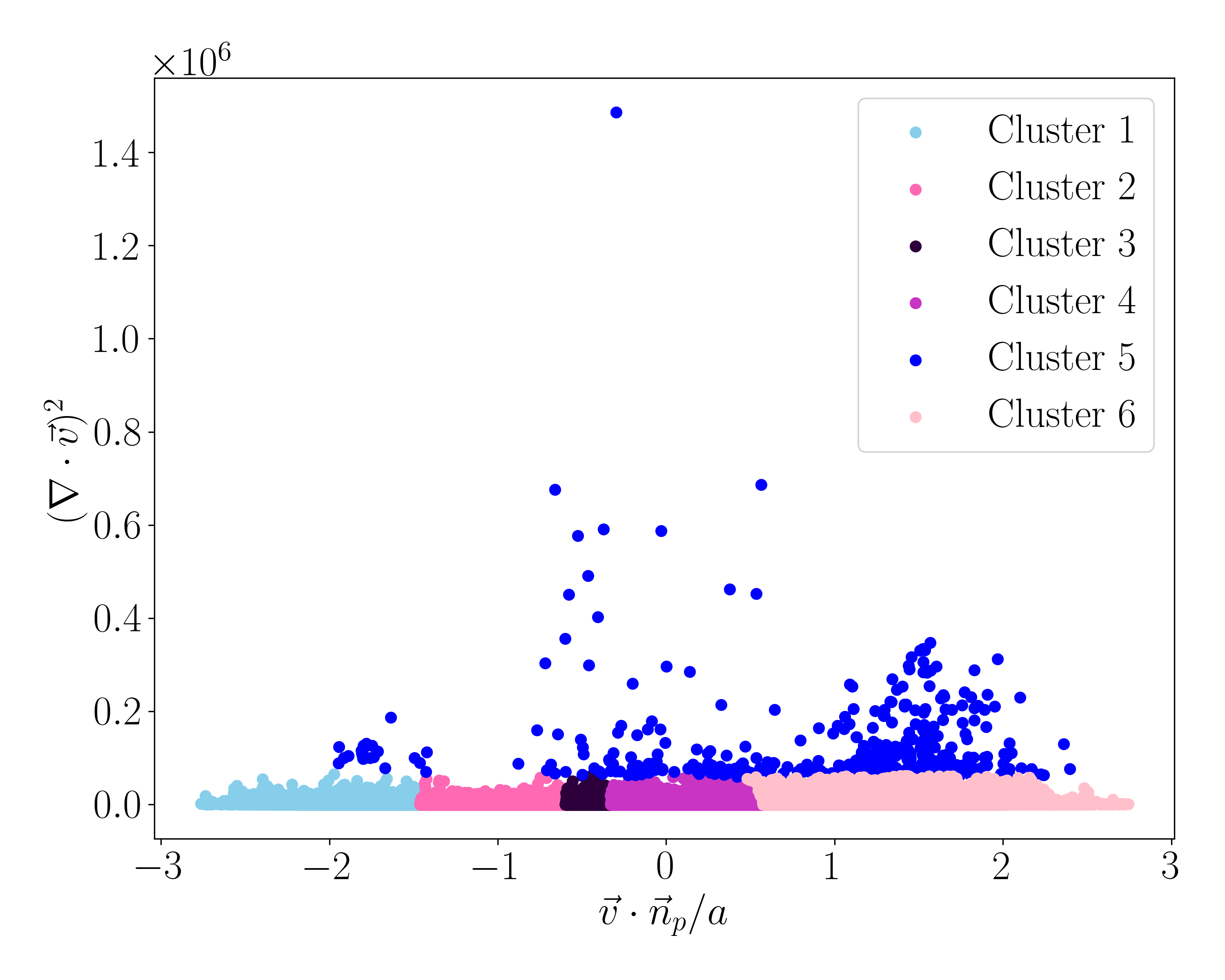}
    \end{minipage}
    }\\\medskip
    \subfloat[$\max(0, M-1), (\nabla\cdot \svec{v})^2$.]{
    \begin{minipage}{0.3\textwidth}
        \includegraphics[width=\textwidth]{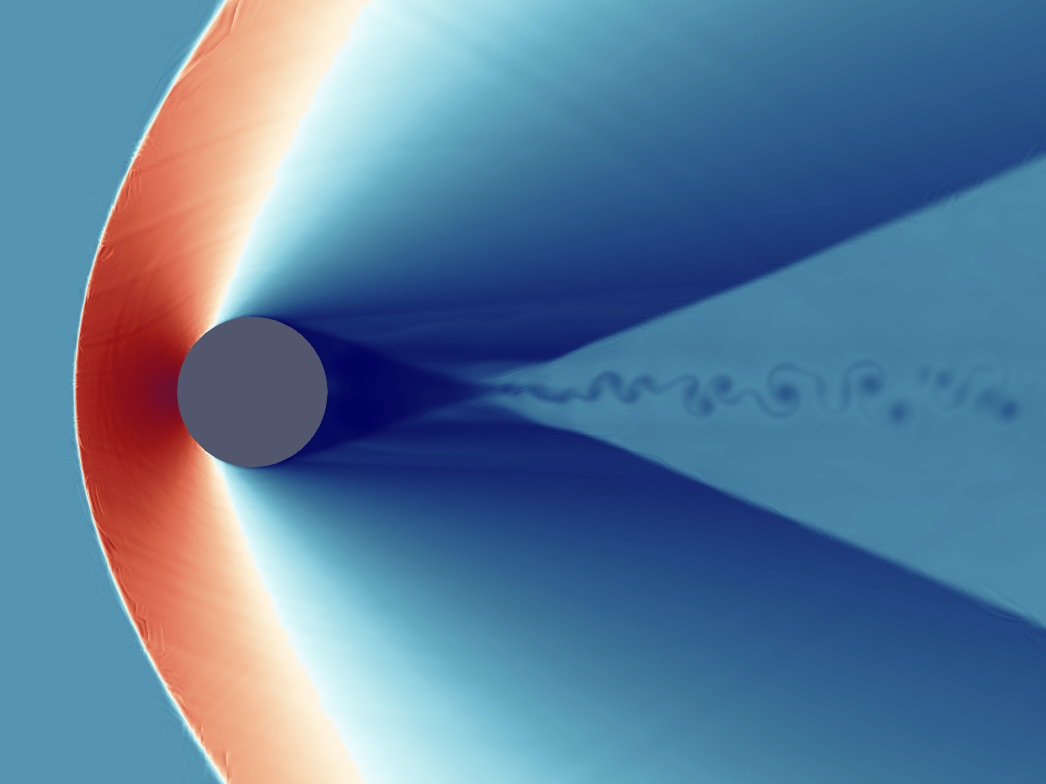}
    \end{minipage}\qquad
    \begin{minipage}{0.3\textwidth}
        \includegraphics[width=\textwidth]{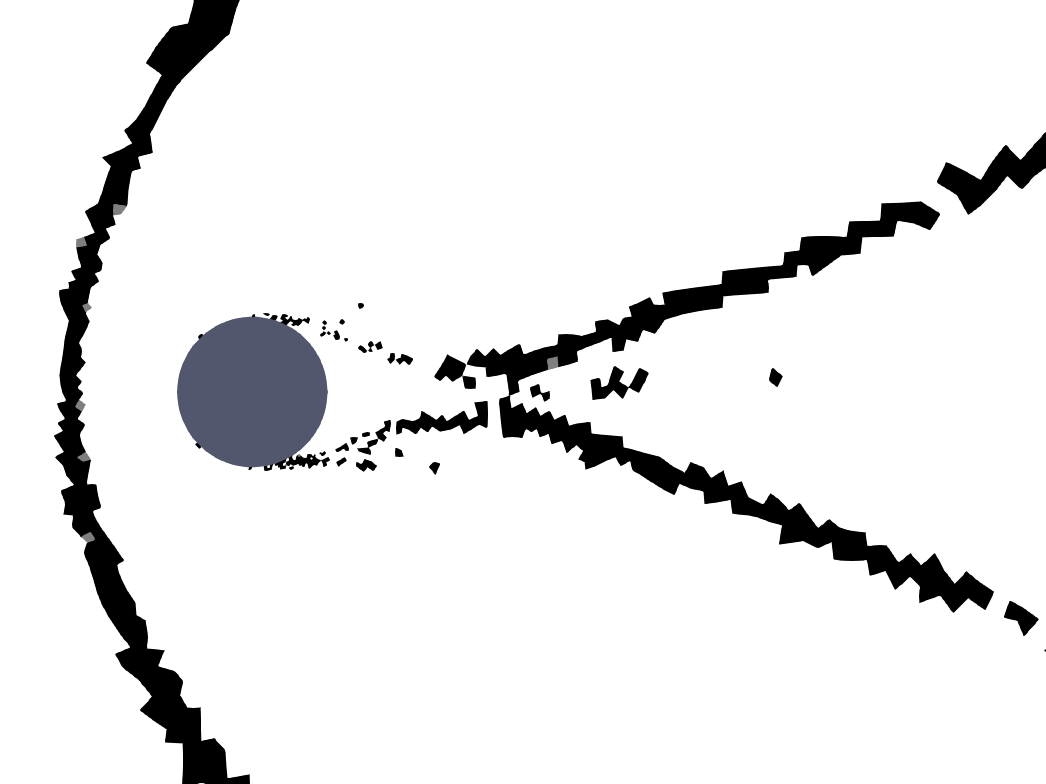}
    \end{minipage}\qquad
    \begin{minipage}{0.3\textwidth}
        \includegraphics[width=\textwidth]{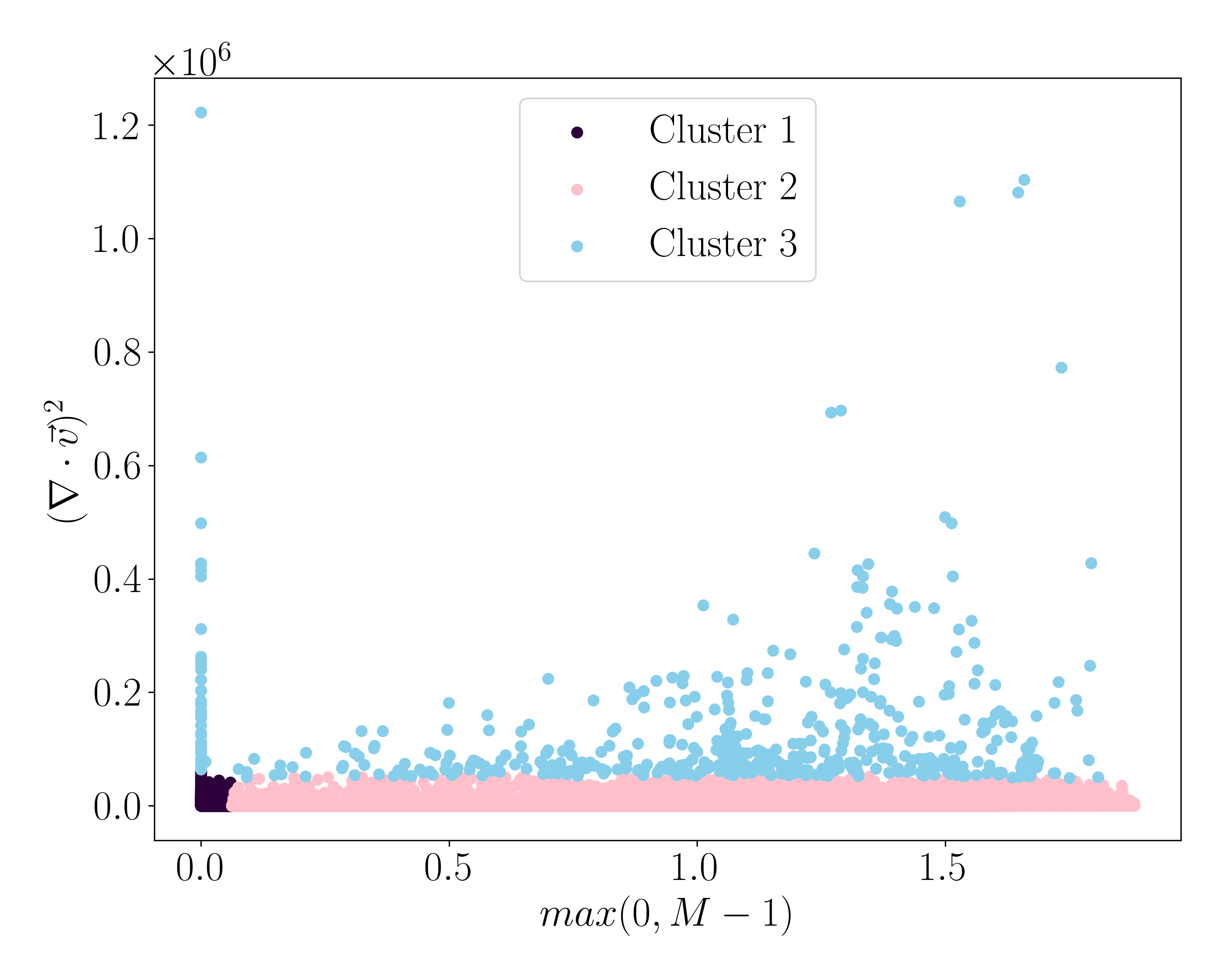}
    \end{minipage}
    }
    \caption{Viscous case after 300,000 iterations with our adaptive GMM sensor of \cref{sec:gmm} using 6 clusters. The simulation with the sensor based on~$(\nabla\cdot \svec{v})^2$ was restarted from the 10,000th time step of the case using $\lVert\nabla p\rVert^2, (\nabla\cdot \svec{v})^2$.}
    \label{fig:other:viscous_gmm}
\end{figure*}

The inherent viscosity of the flow at Re=$10^5$ seems to be beneficial for these sensors.
With weaker oscillations and smoother shocks, they are able to properly detect the shock regions, and they show an interesting behavior. Observing \cref{fig:other:viscous_gmm}, it is evident that the sensor featured in sub-figure a exhibits a higher sensitivity to discontinuities compared to the sensor in sub-figure b, which, in turn, demonstrates a greater sensitivity than the sensor in sub-figure c.
In fact, the adaptive step of our approach removed some of the clusters for the last two sensors, meaning that the algorithm was not able to differentiate all the groups.

\clearpage
\bibliographystyle{elsarticle-num}
\bibliography{library}

\begin{thebibliography}{10}
\expandafter\ifx\csname url\endcsname\relax
  \def\url#1{\texttt{#1}}\fi
\expandafter\ifx\csname urlprefix\endcsname\relax\def\urlprefix{URL }\fi
\expandafter\ifx\csname href\endcsname\relax
  \def\href#1#2{#2} \def\path#1{#1}\fi

\bibitem{anderson1990modern}
J.~D. Anderson, Modern compressible flow: with historical perspective, Vol.~12,
  McGraw-Hill New York, 1990.

\bibitem{pirozzoli2011numerical}
S.~Pirozzoli, Numerical methods for high-speed flows, Annual Review of Fluid
  Mechanics 43 (2011) 163--194.

\bibitem{paciorri2009shock}
R.~Paciorri, A.~Bonfiglioli, A shock-fitting technique for {2D} unstructured
  grids, Computers \& Fluids 38~(3) (2009) 715--726.

\bibitem{Zahr2019}
M.~Zahr, A.~Shi, P.-O. Persson, Implicit shock tracking using an
  optimization-based high-order discontinuous {Galerkin} method, Journal of
  Computational Physics 410 (2020) 109385.

\bibitem{shi2022implicit}
A.~Shi, P.-O. Persson, M.~Zahr, Implicit shock tracking for unsteady flows by
  the method of lines, Journal of Computational Physics 454 (2022) 110906.

\bibitem{eymard2000finite}
R.~Eymard, T.~Gallou{\"e}t, R.~Herbin, Finite volume methods, Handbook of
  Numerical Analysis 7 (2000) 713--1018.

\bibitem{yee1989class}
H.~C. Yee, A class of high-resolution explicit and implicit shock-capturing
  methods, Tech. rep., NACA (1989).

\bibitem{harten1986some}
A.~Harten, S.~Osher, B.~Engquist, S.~R. Chakravarthy, Some results on uniformly
  high-order accurate essentially nonoscillatory schemes, Applied Numerical
  Mathematics 2~(3-5) (1986) 347--377.

\bibitem{shu1988efficient}
C.-W. Shu, S.~Osher, Efficient implementation of essentially non-oscillatory
  shock-capturing schemes, Journal of Computational Physics 77~(2) (1988)
  439--471.

\bibitem{liu1994weighted}
X.-D. Liu, S.~Osher, T.~Chan, Weighted essentially non-oscillatory schemes,
  Journal of Computational Physics 115~(1) (1994) 200--212.

\bibitem{jiang1996efficient}
G.-S. Jiang, C.-W. Shu, Efficient implementation of weighted {ENO} schemes,
  Journal of Computational Physics 126~(1) (1996) 202--228.

\bibitem{leveque2002finite}
R.~J. LeVeque, Finite volume methods for hyperbolic problems, Vol.~31,
  Cambridge University Press, 2002.

\bibitem{shu2003high}
C.-W. Shu, High-order finite difference and finite volume {WENO} schemes and
  discontinuous {Galerkin} methods for {CFD}, International Journal of
  Computational Fluid Dynamics 17~(2) (2003) 107--118.

\bibitem{stevens2020enhancement}
B.~Stevens, T.~Colonius, Enhancement of shock-capturing methods via machine
  learning, Theoretical and Computational Fluid Dynamics 34~(4) (2020)
  483--496.

\bibitem{Huynh2007}
H.~T. Huynh, A flux reconstruction approach to high-order schemes including
  discontinuous {Galerkin} methods, in: 18th AIAA Computational Fluid Dynamics
  Conference, American Institute of Aeronautics and Astronautics, 2007.

\bibitem{vincent2011new}
P.~E. Vincent, P.~Castonguay, A.~Jameson, A new class of high-order energy
  stable flux reconstruction schemes, Journal of Scientific Computing 47 (2011)
  50--72.

\bibitem{cockburn2012discontinuous}
B.~Cockburn, G.~E. Karniadakis, C.-W. Shu, Discontinuous {Galerkin} methods:
  theory, computation and applications, Vol.~11, Springer Science \& Business
  Media, 2012.

\bibitem{cockburn1990runge}
B.~Cockburn, S.~Hou, C.-W. Shu, The {Runge--Kutta} local projection
  discontinuous {Galerkin} finite element method for conservation laws. {IV}.
  {The} multidimensional case, Mathematics of Computation 54~(190) (1990)
  545--581.

\bibitem{burbeau2001problem}
A.~Burbeau, P.~Sagaut, C.-H. Bruneau, A problem-independent limiter for
  high-order {Runge--Kutta} discontinuous {Galerkin} methods, Journal of
  Computational Physics 169~(1) (2001) 111--150.

\bibitem{qiu2005hermite}
J.~Qiu, C.-W. Shu, Hermite {WENO} schemes and their application as limiters for
  {Runge--Kutta} discontinuous {Galerkin} method {II}: {Two} dimensional case,
  Computers \& Fluids 34~(6) (2005) 642--663.

\bibitem{qiu2005comparison}
J.~Qiu, C.-W. Shu, A comparison of troubled-cell indicators for {Runge--Kutta}
  discontinuous {Galerkin} methods using weighted essentially nonoscillatory
  limiters, SIAM Journal on Scientific Computing 27~(3) (2005) 995--1013.

\bibitem{balsara2007sub}
D.~S. Balsara, C.~Altmann, C.-D. Munz, M.~Dumbser, A sub-cell based indicator
  for troubled zones in {RKDG} schemes and a novel class of hybrid {RKDG+}
  {HWENO} schemes, Journal of Computational Physics 226~(1) (2007) 586--620.

\bibitem{yang2009parameter}
M.~Yang, Z.-J. Wang, A parameter-free generalized moment limiter for high-order
  methods on unstructured grids, in: 47th AIAA Aerospace Sciences Meeting
  Including The New Horizons Forum and Aerospace Exposition, 2009, p. 605.

\bibitem{zhong2013simple}
X.~Zhong, C.-W. Shu, A simple weighted essentially nonoscillatory limiter for
  {Runge--Kutta} discontinuous {Galerkin} methods, Journal of Computational
  Physics 232~(1) (2013) 397--415.

\bibitem{dumbser2014posteriori}
M.~Dumbser, O.~Zanotti, R.~Loub{\`e}re, S.~Diot, A posteriori subcell limiting
  of the discontinuous {Galerkin} finite element method for hyperbolic
  conservation laws, Journal of Computational Physics 278 (2014) 47--75.

\bibitem{ntoukas2020freeenergy}
G.~Ntoukas, J.~Manzanero, G.~Rubio, E.~Valero, E.~Ferrer, A free-energy stable
  p-adaptive nodal discontinuous {Galerkin} for the {Cahn--Hilliard} equation,
  Journal of Computational Physics 442 (2020) 110409.

\bibitem{ntoukas2021entropy}
G.~Ntoukas, J.~Manzanero, G.~Rubio, E.~Valero, E.~Ferrer, An entropy-stable
  p-adaptive nodal discontinuous {Galerkin} for the coupled
  {Navier--Stokes}/{Cahn--Hilliard} system, Journal of Computational Physics
  458 (2021) 111093.

\bibitem{mossier2022padaptive}
P.~Mossier, A.~Beck, C.~D. Munz, A p-adaptive discontinuous {Galerkin} method
  with hp-shock capturing, Journal of Scientific Computing 91~(1) (2022) 4.

\bibitem{HENNEMANN2021109935}
S.~Hennemann, A.~M. Rueda-Ramírez, F.~J. Hindenlang, G.~J. Gassner, A provably
  entropy stable subcell shock capturing approach for high order split form
  {DG} for the compressible {Euler} equations, Journal of Computational Physics
  426 (2021) 109935.

\bibitem{RUEDARAMIREZ2022105627}
A.~M. Rueda-Ramírez, W.~Pazner, G.~J. Gassner, Subcell limiting strategies for
  discontinuous {Galerkin} spectral element methods, Computers \& Fluids 247
  (2022) 105627.

\bibitem{MATEOGABIN2023112298}
A.~Mateo-Gabín, A.~M. Rueda-Ramírez, E.~Valero, G.~Rubio, A flux-differencing
  formulation with {Gauss} nodes, Journal of Computational Physics 489 (2023)
  112298.

\bibitem{lin2023high}
Y.~Lin, J.~Chan, High order entropy stable discontinuous {Galerkin} spectral
  element methods through subcell limiting, arXiv preprint arXiv:2306.12663
  (2023).

\bibitem{vonneumann1950method}
J.~von Neumann, R.~D. Richtmyer, A method for the numerical calculation of
  hydrodynamic shocks, Journal of Applied Physics 21~(3) (1950) 232--237.

\bibitem{tadmor1989convergence}
E.~Tadmor, Convergence of spectral methods for nonlinear conservation laws,
  SIAM Journal on Numerical Analysis 26~(1) (1989) 30--44.

\bibitem{guo2001spectral}
B.-y. Guo, H.-p. Ma, E.~Tadmor, Spectral vanishing viscosity method for
  nonlinear conservation laws, SIAM Journal on Numerical Analysis 39~(4) (2001)
  1254--1268.

\bibitem{hartmann2002adaptive}
R.~Hartmann, P.~Houston, Adaptive discontinuous {Galerkin} finite element
  methods for the compressible {Euler} equations, Journal of Computational
  Physics 183~(2) (2002) 508--532.

\bibitem{persson2006sub}
P.-O. Persson, J.~Peraire, Sub-cell shock capturing for discontinuous
  {Galerkin} methods, in: 44th AIAA Aerospace Sciences Meeting and Exhibit,
  2006, p. 112.

\bibitem{feistauer2010discontinuous}
M.~Feistauer, V.~Ku{\v{c}}era, J.~Prokopov{\'a}, Discontinuous {Galerkin}
  solution of compressible flow in time-dependent domains, Mathematics and
  Computers in Simulation 80~(8) (2010) 1612--1623.

\bibitem{barter2010shock}
G.~E. Barter, D.~L. Darmofal, Shock capturing with {PDE-based} artificial
  viscosity for {DGFEM}: Part {I}. formulation, Journal of Computational
  Physics 229~(5) (2010) 1810--1827.

\bibitem{lv2016entropy}
Y.~Lv, Y.~C. See, M.~Ihme, An entropy-residual shock detector for solving
  conservation laws using high-order discontinuous {Galerkin} methods, Journal
  of Computational Physics 322 (2016) 448--472.

\bibitem{fraysse2016upwind}
F.~Fraysse, C.~Redondo, G.~Rubio, E.~Valero, Upwind methods for the
  {Baer--Nunziato} equations and higher-order reconstruction using artificial
  viscosity, Journal of Computational Physics 326 (2016) 805--827.

\bibitem{redondo2017artificial}
C.~Redondo, F.~Fraysse, G.~Rubio, E.~Valero, Artificial viscosity discontinuous
  {Galerkin} spectral element method for the {Baer--Nunziato} equations, in:
  Spectral and High Order Methods for Partial Differential Equations ICOSAHOM
  2016: Selected Papers from the ICOSAHOM conference, June 27-July 1, 2016, Rio
  de Janeiro, Brazil, Springer International Publishing, 2017, pp. 613--625.

\bibitem{mateogabin2022entropy}
A.~Mateo-Gabín, J.~Manzanero, E.~Valero, An entropy stable spectral vanishing
  viscosity for discontinuous {Galerkin} schemes: Application to shock
  capturing and {LES} models, Journal of Computational Physics 471 (2022)
  111618.

\bibitem{jameson1981numerical}
A.~Jameson, W.~Schmidt, E.~Turkel, Numerical solution of the euler equations by
  finite volume methods using {Runge} {Kutta} time stepping schemes, in: 14th
  Fluid and Plasma Dynamics Conference, 1981, p. 1259.

\bibitem{klockner2011viscous}
A.~Kl{\"o}ckner, T.~Warburton, J.~S. Hesthaven, Viscous shock capturing in a
  time-explicit discontinuous {Galerkin} method, Mathematical Modelling of
  Natural Phenomena 6~(3) (2011) 57--83.

\bibitem{huerta2012simple}
A.~Huerta, E.~Casoni, J.~Peraire, A simple shock-capturing technique for
  high-order discontinuous {Galerkin} methods, International Journal for
  Numerical Methods in Fluids 69~(10) (2012) 1614--1632.

\bibitem{rusanov1973processing}
V.~Rusanov, Processing and analysis of computation results for multidimensional
  problems of aerohydrodynamics, in: Proceedings of the Third International
  Conference on Numerical Methods in Fluid Mechanics, Springer, 1973, pp.
  154--162.

\bibitem{vorozhtsov1987shock}
E.~Vorozhtsov, On shock localization by digital image processing techniques,
  Computers \& Fluids 15~(1) (1987) 13--45.

\bibitem{liou1995image}
S.-P. Liou, A.~Singh, S.~Mehlig, D.~Edwards, R.~Davis, An image analysis based
  approach to shock identification in {CFD}, in: 33rd Aerospace Sciences
  Meeting and Exhibit, 1995, p. 117.

\bibitem{wu2013review}
Z.~Wu, Y.~Xu, W.~Wang, R.~Hu, Review of shock wave detection method in {CFD}
  post-processing, Chinese Journal of Aeronautics 26~(3) (2013) 501--513.

\bibitem{sheshadri2014shock}
A.~Sheshadri, A.~Jameson, Shock detection and capturing methods for high order
  {discontinuous-Galerkin} finite element methods, in: 32nd AIAA Applied
  Aerodynamics Conference, 2014, p. 2688.

\bibitem{Tezduyar2004}
T.~E. Tezduyar, Finite Element Methods for Fluid Dynamics with Moving
  Boundaries and Interfaces, John Wiley \& Sons, Ltd, 2004, Ch.~17.

\bibitem{Tezduyar2006a}
T.~E. Tezduyar, M.~Senga, Stabilization and shock-capturing parameters in
  {SUPG} formulation of compressible flows, Computer Methods in Applied
  Mechanics and Engineering 195~(13) (2006) 1621--1632, a Tribute to Thomas
  J.R. Hughes on the Occasion of his 60th Birthday.

\bibitem{Tezduyar2006b}
T.~E. Tezduyar, M.~Senga, D.~Vicker, Computation of inviscid supersonic flows
  around cylinders and spheres with the {SUPG} formulation and {YZ$\beta$}
  shock-capturing, Computational Mechanics 38 (2006) 469--481.

\bibitem{Tezduyar2007a}
T.~E. Tezduyar, M.~Senga, {SUPG} finite element computation of inviscid
  supersonic flows with {YZ$\beta$} shock-capturing, Computers \& Fluids 36~(1)
  (2007) 147--159, challenges and Advances in Flow Simulation and Modeling.

\bibitem{Tezduyar2007b}
Y.~Bazilevs, V.~M. Calo, T.~E. Tezduyar, T.~J.~R. Hughes, {YZ$\beta$}
  discontinuity capturing for advection-dominated processes with application to
  arterial drug delivery, International Journal for Numerical Methods in Fluids
  54~(6--8) (2007) 593--608.

\bibitem{brunton_kutz_2019}
S.~L. Brunton, J.~N. Kutz, Data-Driven Science and Engineering: Machine
  Learning, Dynamical Systems, and Control, Cambridge University Press, 2019.

\bibitem{GARNIER2021104973}
P.~Garnier, J.~Viquerat, J.~Rabault, A.~Larcher, A.~Kuhnle, E.~Hachem, A review
  on deep reinforcement learning for fluid mechanics, Computers \& Fluids 225
  (2021) 104973.

\bibitem{vinuesa2022enhancing}
R.~Vinuesa, S.~L. Brunton, Enhancing computational fluid dynamics with machine
  learning, Nature Computational Science 2~(6) (2022) 358--366.

\bibitem{brenner2019perspective}
M.~Brenner, J.~Eldredge, J.~Freund, Perspective on machine learning for
  advancing fluid mechanics, Physical Review Fluids 4~(10) (2019) 100501.

\bibitem{brunton2020machine}
S.~L. Brunton, B.~R. Noack, P.~Koumoutsakos, Machine learning for fluid
  mechanics, Annual Review of Fluid Mechanics 52 (2020) 477--508.

\bibitem{fukami2020assessment}
K.~Fukami, K.~Fukagata, K.~Taira, Assessment of supervised machine learning
  methods for fluid flows, Theoretical and Computational Fluid Dynamics 34~(4)
  (2020) 497--519.

\bibitem{clainche2022improving}
S.~{Le Clainche}, E.~Ferrer, S.~Gibson, E.~Cross, A.~Parente, R.~Vinuesa,
  Improving aircraft performance using machine learning: A review, Aerospace
  Science and Technology 138 (2023) 108354.

\bibitem{veiga2018general}
M.~H. Veiga, R.~Abgrall, Towards a general stabilisation method for
  conservation laws using a multilayer perceptron neural network: {1D} scalar
  and system of equations, in: ECCM - ECFD2018 6th European Conference on
  Computational Mechanics (Solids, Structures and Coupled Problems) 7th
  European Conference on Computational Fluid Dynamics, Glasgow, United Kingdom,
  2018.

\bibitem{ray2018artificial}
D.~Ray, J.~Hesthaven, An artificial neural network as a troubled-cell
  indicator, Journal of Computational Physics 367 (2018) 166--191.

\bibitem{ray2019detecting}
D.~Ray, J.~Hesthaven, Detecting troubled-cells on two-dimensional unstructured
  grids using a neural network, Journal of Computational Physics 397 (2019)
  108845.

\bibitem{morgan2020machine}
N.~Morgan, S.~Tokareva, X.~Liu, A.~Morgan, A machine learning approach for
  detecting shocks with high-order hydrodynamic methods, in: AIAA Scitech 2020
  Forum, 2020, p. 2024.

\bibitem{BECK2020109824}
A.~D. Beck, J.~Zeifang, A.~Schwarz, D.~G. Flad, A neural network based shock
  detection and localization approach for discontinuous {Galerkin} methods,
  Journal of Computational Physics 423 (2020) 109824.

\bibitem{Xinyue2021}
X.~Yu, C.-W. Shu, Multi-layer perceptron estimator for the total variation
  bounded constant in limiters for discontinuous {Galerkin} methods, La
  Matematica 1 (2022) 53--84.

\bibitem{KOSSACZKA2021100201}
T.~Kossaczká, M.~Ehrhardt, M.~Günther, Enhanced fifth order {WENO}
  shock-capturing schemes with deep learning, Results in Applied Mathematics 12
  (2021) 100201.

\bibitem{Xue2022}
Z.~Xue, Y.~Xia, C.~Li, X.~Yuan, A simplified multilayer perceptron detector for
  the hybrid {WENO} scheme, Computers \& Fluids 244 (2022) 105584.

\bibitem{ZEIFANG2021110475}
J.~Zeifang, A.~Beck, A data-driven high order sub-cell artificial viscosity for
  the discontinuous {Galerkin} spectral element method, Journal of
  Computational Physics 441 (2021) 110475.

\bibitem{otmaniRobustDetectionViscous2022}
K.-E. Otmani, G.~Ntoukas, O.~A. Mari{\~n}o, E.~Ferrer, Toward a robust
  detection of viscous and turbulent flow regions using unsupervised machine
  learning, Physics of Fluids 35~(2) (2023) 027112.

\bibitem{saettaIdentificationFlowFielda}
E.~Saetta, R.~Tognaccini, Identification of flowfield regions by machine
  learning, AIAA Journal 61~(4) (2023) 1503--1518.

\bibitem{tlales2022machine}
K.~Tlales, K.~E. Otmani, G.~Ntoukas, G.~Rubio, E.~Ferrer, Machine learning
  adaptation for laminar and turbulent flows: applications to high order
  discontinuous {Galerkin} solvers, arXiv preprint arXiv:2209.02401 (2022).

\bibitem{Zhu2021}
H.~Zhu, H.~Wang, Z.~Gao, A new troubled-cell indicator for discontinuous
  {Galerkin} methods using k-means clustering, SIAM Journal on Scientific
  Computing 43~(4) (2021) A3009--A3031.

\bibitem{Black1999}
K.~Black, A conservative spectral element method for the approximation of
  compressible fluid flow, Kybernetika 35 (1999).

\bibitem{kopriva2009implementing}
D.~A. Kopriva, Implementing spectral methods for partial differential
  equations: Algorithms for scientists and engineers, Springer Science \&
  Business Media, 2009.

\bibitem{Ferrer2023}
E.~Ferrer, G.~Rubio, G.~Ntoukas, W.~Laskowski, O.~Mariño, S.~Colombo,
  A.~Mateo-Gabín, H.~Marbona, F.~M. de~Lara, D.~Huergo, J.~Manzanero,
  A.~Rueda-Ramírez, D.~Kopriva, E.~Valero, {HORSES3D}: A high-order
  discontinuous {Galerkin} solver for flow simulations and multi-physics
  applications, Computer Physics Communications 287 (2023) 108700.

\bibitem{Gassner2018}
G.~J. Gassner, A.~R. Winters, F.~J. Hindenlang, D.~A. Kopriva, The {BR1} scheme
  is stable for the compressible {Navier--Stokes} equations, Journal of
  Scientific Computing 77 (2018) 154--200.

\bibitem{Guermond2014}
J.~L. Guermond, B.~Popov, Viscous regularization of the {Euler} equations and
  entropy principles, SIAM Journal on Applied Mathematics 74 (2014) 284--305.

\bibitem{Persson2006}
P.-O. Persson, J.~Peraire, Sub-cell shock capturing for discontinuous
  {Galerkin} methods, Collection of Technical Papers - 44th AIAA Aerospace
  Sciences Meeting 2 (2006) 1408--1420.

\bibitem{Joshi2023}
S.~Joshi, J.~Kou, A.~{Hurtado de Mendoza}, K.~Puri, C.~Hirsch, G.~Rubio,
  E.~Ferrer, Length-scales for efficient {CFL} conditions in high-order methods
  with distorted meshes: Application to local-timestepping for p-multigrid,
  Computers \& Fluids 265 (2023) 106011.

\bibitem{Fisher2013}
T.~C. Fisher, M.~H. Carpenter, High-order entropy stable finite difference
  schemes for nonlinear conservation laws: Finite domains, Journal of
  Computational Physics 252 (2013) 518--557.

\bibitem{Zhang2011}
X.~Zhang, C.-W. Shu, Maximum-principle-satisfying and positivity-preserving
  high-order schemes for conservation laws: survey and new developments,
  Journal of Computational Physics 467 (2011) 3091--3120.

\bibitem{Bassi1997}
F.~Bassi, S.~Rebay, A high-order accurate discontinuous finite element method
  for the numerical solution of the compressible {Navier--Stokes} equations,
  Journal of Computational Physics 131 (1997) 267--279.

\bibitem{Chandrashekar2013}
P.~Chandrashekar, Kinetic energy preserving and entropy stable finite volume
  schemes for compressible {Euler} and {Navier--Stokes} equations,
  Communications in Computational Physics 14 (2013) 1252--1286.

\bibitem{Ismail2009}
F.~Ismail, P.~L. Roe, Affordable, entropy-consistent {Euler} flux functions
  {II}: Entropy production at shocks, Journal of Computational Physics 228
  (2009) 5410--5436.

\bibitem{ryujin-2021-1}
M.~Maier, M.~Kronbichler, Efficient parallel {3D} computation of the
  compressible {Euler} equations with an invariant-domain preserving
  second-order finite-element scheme, ACM Transactions on Parallel Computing
  8~(3) (2021) 16:1--30.

\bibitem{Carlson2011}
J.-R. Carlson, Inflow/outflow boundary conditions with application to {FUN3D},
  Tech. rep., NASA (2011).

\bibitem{Mengaldo2014}
G.~Mengaldo, D.~D. Grazia, F.~Witherden, A.~Farrington, P.~Vincent, S.~Sherwin,
  J.~Peiro, A guide to the implementation of boundary conditions in compact
  high-order methods for compressible aerodynamics, in: 7th AIAA Theoretical
  Fluid Mechanics Conference, American Institute of Aeronautics and
  Astronautics, 2014.

\bibitem{Gowen1952}
F.~E. Gowen, E.~W. Perkins, Drag of circular cylinders for a wide range of
  {Reynolds} numbers and {Mach} numbers, Tech. rep., NACA (6 1952).

\bibitem{Bashkin2000}
V.~A. Bashkin, I.~V. Egorov, M.~V. Egorova, D.~V. Ivanov, Supersonic
  laminar-turbulent gas flow past a circular cylinder, Fluid Dynamics 35 (2000)
  652--662.

\bibitem{Bashkin2002}
V.~A. Bashkin, A.~V. Vaganov, I.~V. Egorov, D.~V. Ivanov, G.~A. Ignatova,
  Comparison of calculated and experimental data on supersonic flow past a
  circular cylinder, Fluid Dynamics 37 (2002) 473--483.

\bibitem{sklearn}
F.~Pedregosa, G.~Varoquaux, A.~Gramfort, V.~Michel, B.~Thirion, O.~Grisel,
  M.~Blondel, P.~Prettenhofer, R.~Weiss, V.~Dubourg, J.~Vanderplas, A.~Passos,
  D.~Cournapeau, M.~Brucher, M.~Perrot, E.~Duchesnay, Scikit-learn: Machine
  learning in {Python}, Journal of Machine Learning Research 12 (2011)
  2825--2830.

\bibitem{Gassner2016}
G.~J. Gassner, A.~R. Winters, D.~A. Kopriva, Split form nodal discontinuous
  {Galerkin} schemes with summation-by-parts property for the compressible
  euler equations, Journal of Computational Physics 327 (2016) 39--66.

\end{thebibliography}

\end{document}